\documentclass[journal]{IEEEtran}
\usepackage{graphicx}
\usepackage{amsmath}
\usepackage{amssymb}
\usepackage{subfigure}
\usepackage{stfloats}
\usepackage{bm}
\usepackage{bbm}
\usepackage{algorithm}
\usepackage{algorithmicx}
\usepackage{algpseudocode}
\usepackage{tabularx}
\usepackage{multirow,hhline}
\usepackage{array}
\usepackage{multirow}
\usepackage{color}
\usepackage[table]{xcolor}
\usepackage{pifont}
\usepackage{url}
\usepackage[linkcolor=black,citecolor=black,urlcolor=green,colorlinks=true]{hyperref}
\usepackage{verbatim}
\usepackage{setspace}
\usepackage{mathrsfs}
\usepackage[]{pifont}
\usepackage{cite}
\usepackage{hyperref}
\hypersetup{urlcolor=magenta}

\begin{document}
\title{\huge ERPoT: Effective and Reliable Pose Tracking for Mobile Robots Using Lightweight Polygon Maps}
\author{\IEEEauthorblockN{Haiming Gao,~\IEEEmembership{Member,~IEEE}, Qibo Qiu, Hongyan Liu, Dingkun Liang, Chaoqun Wang,~\IEEEmembership{Member,~IEEE}, \\Xuebo Zhang,~\IEEEmembership{Senior Member,~IEEE}}
	\thanks{This work was supported in part by the National	Natural Science Foundation of China under Grant 62303428, and in part by Beijing-Tianjin-Hebei Fundamental Research Cooperation Project under Grant 24JCZXJC00390. \emph{(Corresponding author: Dingkun Liang)}}%
	\thanks{Haiming Gao is with the ZJU-Hangzhou Global Scientific and Technological Innovation Center, Zhejiang University, Hangzhou, P.~R.~China, 311215 (Email: ghm@zju.edu.cn).}
	\thanks{Qibo Qiu is with State Key Lab of CAD{\&}CG, Zhejiang University, and also with China Mobile (Zhejiang) Research {\&} Innovation Institute.}
	\thanks{Hongyan Liu is with Beijing Fuyouhua Intelligent Technology Co., Ltd.}
	\thanks{Dingkun Liang is with College of Information Engineering, Zhejiang University of Technology, Hangzhou, P.~R.~China, 310023 (Email: liangdk@zjut.edu.cn).}
	\thanks{Chaoqun Wang is with School of Control Science and Engineering, Shandong University, Jinan, P.~R.~China, 250061.}
	\thanks{Xuebo Zhang is with the Institute of Robotics and Automatic Information System (IRAIS), Tianjin Key Laboratory of Intelligent Robotics (TJKLIR), Nankai University, Tianjin, P.~R.~China, 300350.}%
}

\maketitle

\begin{abstract}
This paper presents an effective and reliable pose tracking solution, termed ERPoT, for mobile robots operating in large-scale outdoor and challenging indoor environments, underpinned by an innovative prior polygon map. Especially, to overcome the challenge that arises as the map size grows with the expansion of the environment, the novel form of a prior map composed of multiple polygons is proposed. Benefiting from the use of polygons to concisely and accurately depict environmental occupancy, the prior polygon map achieves long-term reliable pose tracking while ensuring a compact form. More importantly, pose tracking is carried out under pure LiDAR mode, and the dense 3D point cloud is transformed into a sparse 2D scan through ground removal and obstacle selection. On this basis, a novel cost function for pose estimation through point-polygon matching is introduced, encompassing two distinct constraint forms: point-to-vertex and point-to-edge. In this study, our primary focus lies on two crucial aspects: lightweight and compact prior map construction, as well as effective and reliable robot pose tracking. Both aspects serve as the foundational pillars for future navigation across diverse mobile platforms equipped with different LiDAR sensors in varied environments. Comparative experiments based on the publicly available datasets and our self-recorded datasets are conducted, and evaluation results show the superior performance of ERPoT on reliability, prior map size, pose estimation error, and runtime over the other six approaches. The corresponding code can be accessed at {\url{https://github.com/ghm0819/ERPoT}}, and the supplementary video is at {\url{https://youtu.be/6XdcXyUrLKw}}.
\end{abstract}

\begin{IEEEkeywords}
Pose Tracking, Mobile Robots, Prior Polygon Map, Point-Polygon Matching
\end{IEEEkeywords}
\IEEEpeerreviewmaketitle
\section{Introduction}

\IEEEPARstart{M}{obile} robotics is the trending area where related research has made great progress in the past decade, with accurate localization and robust navigation emerging as cornerstones for the development of autonomous capabilities \cite{Placed, JWen}. As the application of mobile robots extends into more complex environments, the demand for high-precision and reliable localization systems intensifies \cite{Singamaneni}.

In response, many state-of-the-art approaches \cite{Campos, Shan2, JDeng} have been proposed to obtain environmental maps, which serve as the prior information for robust localization of mobile robots in various applications such as service \cite{Graf}, navigation \cite{CWang}, and exploration \cite{QBi}. When both the prior map and the initial pose are provided, the localization problem can be regarded as \emph{pose tracking} \cite{HYin}, which continuously estimates the robot pose. The pursuit of effective and reliable approaches for pose tracking is central to advancing the field of mobile robotics.

\begin{figure}[!t]
	\centering
	\includegraphics[width=6.5cm]{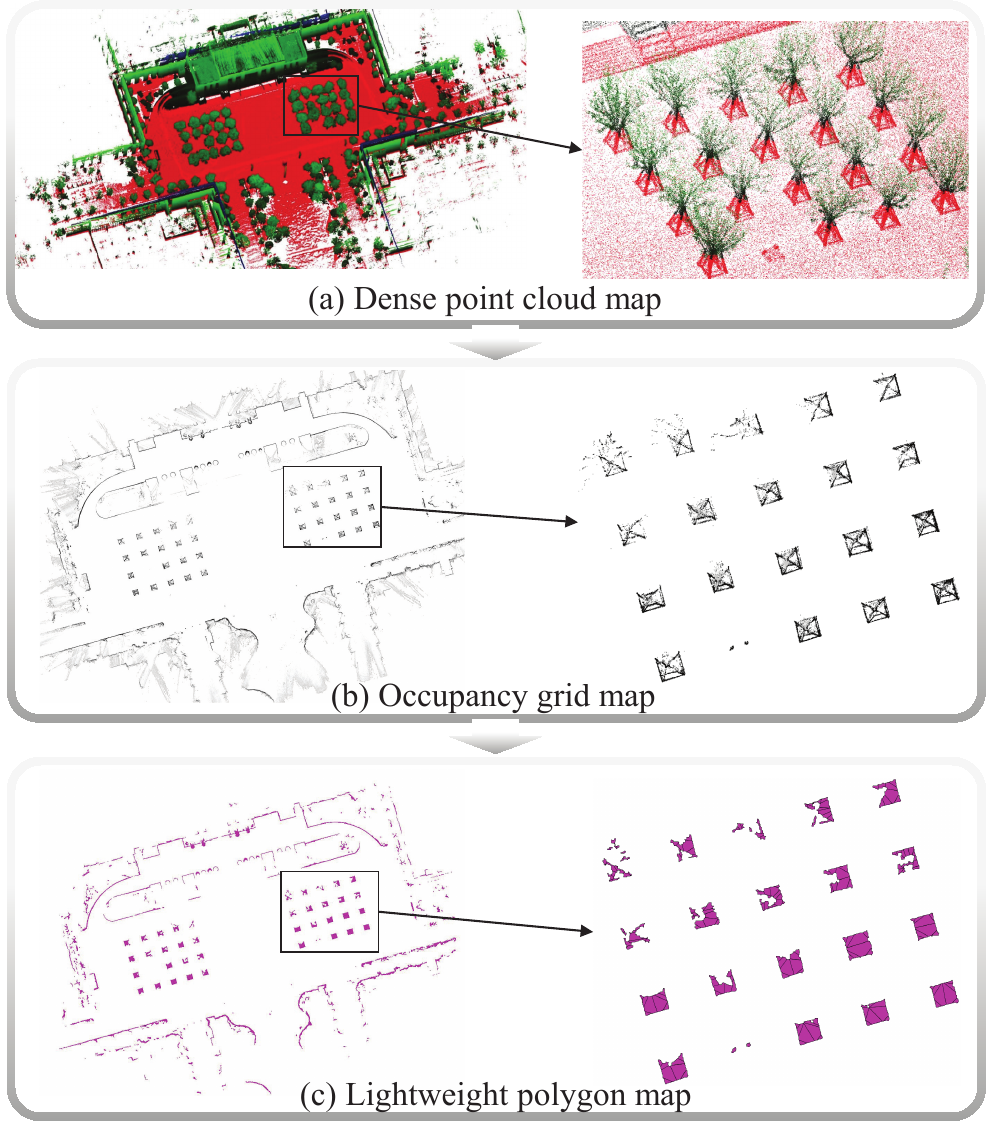}
	\caption{Lightweight and compact polygon map construction.}
	\label{Fig_1}
\end{figure}

More importantly, the effectiveness and reliability of pose tracking fundamentally depend on the quality of the prior map used for organizing environmental elements, and the superiority of the algorithms employed for processing sensor data and achieving accurate data association. For pose tracking methodologies that leverage LiDAR sensors, they can generally be categorized into key approaches, including point cloud registration techniques \cite{Koide2, Dellenbach, Vizzo}, feature-based matching \cite{JZhang, Shan, HDong}, machine learning-based algorithms \cite{JDeng, Isaacson, RHuang}, and various filter-based methods \cite{XChen, Akai, Rico}. These diverse strategies reflect the multifaceted nature of pose estimation within the field of autonomous systems. However, despite the success of previous pose tracking methods, they often face certain challenges, such as the large size of prior maps \cite{Dube1}, particularly in large-scale environments, which can hinder computational efficiency and map storage. In addition, complex pose tracking processes with specific semantic clues and structure characteristics may not always guarantee effectiveness and reliability\cite{YHuang}. In light of these challenges, there is a pressing need for innovative solutions that can offer both \emph{compact map form} and \emph{reliable localization technique}.

Motivated by the requirements mentioned above, we propose an effective and reliable pose tracking approach for mobile robots based on the novel prior map, called ERPoT. For mobile robots operating in terrestrial environments, the roll and pitch angles are typically stable and exhibit minimal variation \cite{Kummerle}. Consequently, the proposed ERPoT, which operates on three degrees of freedom, is generally adequate to effectively address the requirements of nearly all task scenarios. In addition, the proposed pose tracking framework is a general solution that applies to various environments and is not limited to environments containing specific semantic elements. Specifically, LiDAR data serves as the sole input, from which a sparse 2D scan is derived through ground removal and obstacle selection. Subsequently, an offline map construction process is performed to acquire a lightweight and compact polygon map of the environment, as shown in Fig. \ref{Fig_1}. Multi-polygon-based representation (Fig. \ref{Fig_1}(c)) offers superior advantages over point cloud (Fig. \ref{Fig_1}(a)) and occupancy grid (Fig. \ref{Fig_1}(b)), particularly in applications demanding precision and efficient data utilization \cite{DYu}. Finally, we can make use of the 2D scan and the prior polygon map to realize effective and reliable pose tracking for mobile robots, which takes the advantage of the newly proposed point-polygon matching. This research focuses on constructing lightweight and compact prior maps and establishing effective and reliable pose tracking. The main contributions include three aspects:

(1) \textbf{Innovative form of the prior map}: The prior map composed of multiple polygons is proposed for pose tracking of mobile robots, which effectively compresses the environmental information without relying on any semantic elements. More importantly, the prior polygon map realizes the lightweight and compact representation of the large-scale environment, while ensuring effective and reliable pose tracking performance.

(2) \textbf{Point-polygon matching-based pose tracking}: Based on the prior map, an effective and reliable pose tracking approach is proposed, called ERPoT. Firstly, the dense point cloud is compressed into a sparse 2D scan through ground removal and obstacle selection. On this basis, the newly proposed cost function on \emph{point-polygon matching} is constructed, which includes \emph{point-to-vertex} and \emph{point-to-edge}.

(3) \textbf{Generalization across platforms and environments}: Extensive comparative experiments on public datasets (including building squares, road scenes, vegetation, \emph{etc.}) and self-recorded datasets (including long-term changes, and terrain changes), which are acquired through different platforms equipped with different LiDAR sensors, demonstrating the effectiveness and reliability of our proposed approach.

The remaining parts of this paper are structured as follows. Section II describes related works on prior map-based pose tracking. Section III presents the effective and reliable pose tracking approach called ERPoT in detail. Section IV provides the experimental results to show the effectiveness of ERPoT. Finally, the paper is concluded in Section V.

\section{Related work}
Reliable pose tracking is the backbone of autonomous mobile robots, enabling them to move intelligently, interact safely, and perform complex tasks. In this section, we briefly discuss previous related works in the fields of LiDAR odometry, prior map construction, and pose tracking.

\subsection{LiDAR odometry}
LiDAR odometry, a critical technique for mobile robots, leverages LiDAR sensors to estimate the trajectory and construct a map of the environment in real-time. Current LiDAR odometry pipelines commonly employ an iterative closest point (ICP) algorithm and its variants to realize pose estimation incrementally. Especially, the authors in \cite{Koide2} present a voxelized generalized iterative closest point (VGICP) approach, bridging the gap between the generalized iterative closest point (GICP) and the normal distributions transform (NDT), which achieves fast and accurate 3D point cloud registration. By introducing the definition of continuous-time trajectory, Dellenbach \emph{et al.} \cite{Dellenbach} propose a novel real-time LiDAR-only odometry approach called CT-ICP, combined with a novel loop detection procedure to realize a complete SLAM system. In recent years, research on ICP-based LiDAR odometry has been an active field. Vizzo \emph{et al.} \cite{Vizzo} present a novel pose estimation system called KISS-ICP, which is a lightweight and efficient ICP algorithm designed for real-time applications, emphasizing simplicity and performance without compromising accuracy. In addition, MAP-ICP is proposed in \cite{Ferrari}, utilizing an efficient and versatile KD-tree data structure and dynamically maintaining a robust environment model through the incorporation of estimated pose uncertainty.

However, GICP and other ICP-variants, which heavily depend on nearest neighbor search, can struggle to achieve real-time performance in large-scale environments, especially when operating on computers with limited computational capabilities \cite{WXu}. On the other hand, feature-based registration methods first extract relatively fewer salient features from point clouds and then estimate the transformation only with the feature points. As the milestone work, LiDAR odometry and mapping (LOAM) is proposed in \cite{JZhang}, computing the robot pose with planar and edge feature points. Inspired by the well-known LOAM, Shan \emph{et al.} \cite{Shan} propose the novel LeGO-LOAM, which adds ground constraints to improve the accuracy of localization and mapping. Furthermore, the authors in \cite{Shan2} have introduced a framework termed LIO-SAM, which excels in delivering highly accurate, real-time trajectory estimation and map-building capabilities for mobile robots. Recently, F-LOAM \cite{HWang}, presented as a comprehensive solution, aims to provide a computationally efficient and precise framework for LiDAR-based SLAM. In addition, to adapt to resource-limited platforms, a lightweight LiDAR SLAM system, Light-LOAM \cite{SYi}, is proposed with comparative performance to the state-of-the-art methods, while also striking a balance with its computational requirements.

Different from the above classical LiDAR odometry systems, deep learning-based LiDAR odometry does not rely on hand-tuned feature extraction and data association, and has attracted much attention in the last few years. Especially, Neural Radiance Field (NeRF) presents an innovative approach to map representation, showing great potential for utilization in robotics applications. On this basis, Isaacson \emph{et al.} \cite{Isaacson} propose LONER, the first real-time neural implicit LiDAR SLAM adapting to outdoor environments with accurate online state estimation. To address the problem of NeRF-based LiDAR SLAM used in outdoor large-scale environments, the work in \cite{JDeng} proposes a novel NeRF-based LiDAR odometry and mapping approach, NeRF-LOAM, which obtains both dense 3D representation and accurate poses. On the other hand, several related works \cite{RHuang, ZQin} firstly extract local robust descriptors of the point cloud and then make use of classical refinement techniques to realize pose estimation.

\subsection{Prior map construction}
The method and form of prior map construction are fundamental to the performance of prior map-based pose tracking for mobile robots. PoseMap proposed in \cite{Egger} integrates the trajectory directly into the prior map, each robot pose is associated with a submap consisting of surfels, transcending the limitations of purely metric map representations and enhancing the overall efficiency. This topological structure of prior maps can improve pose tracking efficiency in large-scale environments. To fulfill the increasing computational requirements that come with the size of the prior map, Feng \emph{et al.} \cite{YFeng} propose a localization system based on Block Maps to reduce the computational load. In \cite{FXie}, Xie \emph{et al.} present a solution for robust indoor localization in environments like offices and corridors, utilizing 3D LiDAR point clouds within a compact, hierarchical, topometric semantic map.

In contrast, leveraging salient objects and distinctive feature information within the environment allows for a more compact representation, thereby enhancing the robustness of pose estimation. Especially, the work in \cite{Dube1} proposes Segmap, a segment-based approach for map representation in localization and mapping based on the previous works \cite{Dube2, Dube3}, enhancing traditional approaches with its distinctive, learning-based descriptors that offer superior descriptive power and competitive localization performance. On the other hand, pole-like objects are well suited to serve as landmarks for vehicle localization in urban environments due to their widespread presence and consistent reliability over time, Schaefer \emph{et al.} \cite{Schaefer} present a long-term 2D vehicle localization approach in urban environments with the extracted pole landmarks from mobile LiDAR data. Furthermore, the study in \cite{HDong} capitalizes on extracted poles to train a deep neural network, facilitating online range image-based pole segmentation. This approach avoids processing the 3D point cloud, thereby enabling rapid pole extraction for long-term localization. Recently, based on the semantics of pole-like structures, the authors in \cite{YHuang} introduce a multi-layer semantic pole-map by aggregating the detected pole-like structures, and the semantic particle-filtering localization scheme is also proposed for vehicle localization.

Prior map construction methods mentioned above, when confronted with large-scale outdoor environments, employ special map structures like topology to enhance the efficiency of data association, yet struggle to effectively manage map size. On the other hand, leveraging semantically salient objects allows for a compact environmental description and ensures reliable pose tracking. However, this approach is only viable in outdoor settings that meet specific semantic requirements.

\subsection{Pose tracking}
Accurate prior map-based pose tracking allows mobile robots to determine their exact location within an environment, which is essential for navigating and interacting with their surroundings effectively. Especially, for autonomous navigation, self-driving cars commonly rely on precise localization within a prior map to guide their way \cite{Hafez}. In \cite{Wolcott}, Wolcott \emph{et al.} model the world as a mixture of several Gaussians and present a generic probabilistic-based localization system that can handle harsh weather conditions and poorly textured roadways. Based on the mesh-based prior map, the authors in \cite{XChen} introduce an innovative observation model, seamlessly integrating it within a Monte Carlo localization framework to enhance accuracy and robustness. The Monte Carlo-based localization method has been widely used in large-scale outdoor tasks due to its advantages in dealing with uncertainty, nonlinear problems, and computational efficiency. For instance,  Akai \cite{Akai} merges the strengths of Monte Carlo localization with those of scan matching, culminating in the proposal of an efficient map-based localization solution that capitalizes on the best of both methodologies. Considering non-flat terrain characteristics of the operational environments, Rico \emph{et al.} \cite{Rico} present a mobile robot localization algorithm grounded in AMCL, utilizing elevation maps and probabilistic 3D occupancy maps to determine the robot pose accurately.

On the other hand, scan matching also plays a pivotal role in mobile robot pose tracking by aligning sequential point cloud data with the prior map to compute the robot pose. In \cite{Koide1}, Koide \emph{et al.} propose a real-time sensor localization system, integrating the NDT scan matching \cite{Koide2} with an angular velocity-based pose prediction using an unscented Kalman filter (UKF). For long-term applications of mobile robots, the work in \cite{BPeng} introduces a resilient long-term LiDAR-based localization system, which incorporates a temporary mapping module to circumvent potential localization failures due to incorrect global matching. In contrast, in a data-driven manner, Hroob \emph{et al.} \cite{Hroob} propose a novel stability scan filter for robust localization of field mobile robots through long-term stability landmarks in a continuously changing environment. In addition, the authors in \cite{Montano} present a novel tightly-coupled graph localization approach that exploits prior topo-metric knowledge of the environment and seamlessly integrates various modules into a graph optimization framework.

For the aforementioned prior map-based pose tracking methods, a challenge arises as the map size grows with the expansion of the environment. Particularly in large-scale outdoor settings, the storage of prior maps can demand substantial memory resources, and the large maps can present significant computational challenges for pose tracking. This challenge is the driving force behind our novel polygon map approach, which ensures minimal map sizes in extensive environments without compromising the reliability of pose tracking. 

\section{Effective and reliable pose tracking}

\begin{figure*}[!htb]
	\centering
	\includegraphics[width=17.5cm]{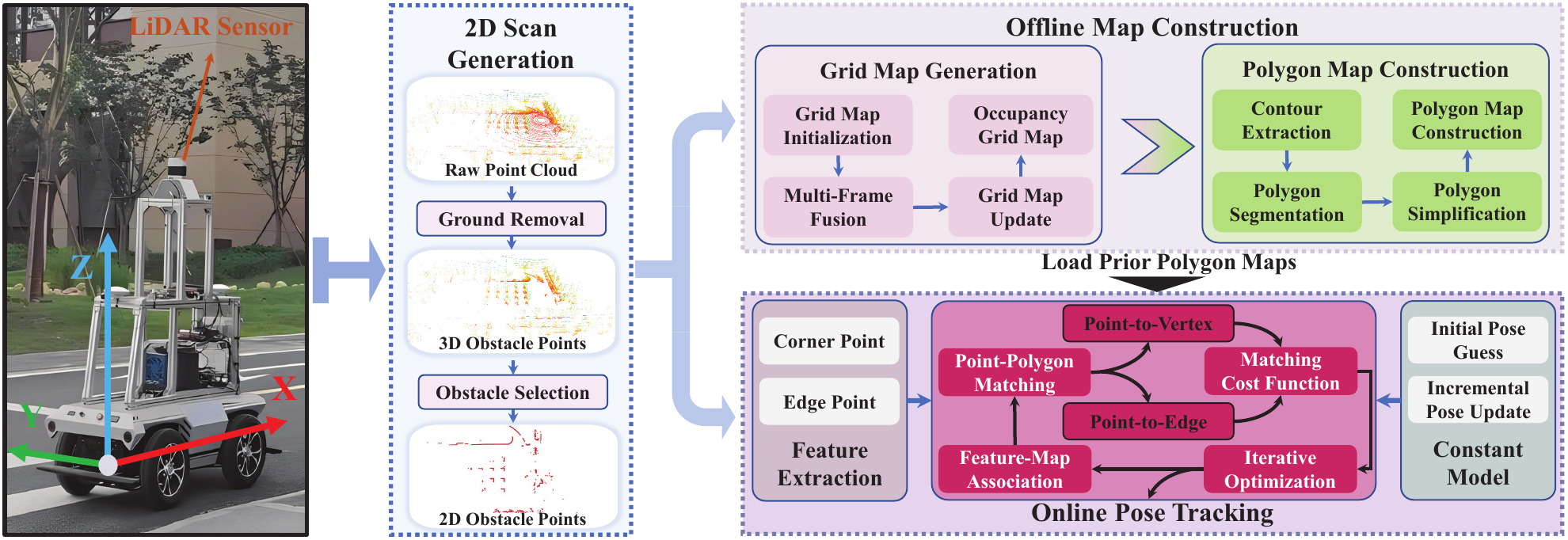}
	\caption{The process of \textbf{E}ffective and \textbf{R}eliable \textbf{Po}se \textbf{T}racking approach (\textbf{ERPoT}) for mobile robots, which mainly includes two parts: offline map construction and online pose tracking. The left subfigure shows the experimental platform for collecting the self-recorded datasets, and the middle subfigure represents the point cloud preprocessing process, named 2D scan generation (\textbf{Section III-B}), which is used in the following two main processes. In particular, pink module and purple module represent offline map construction (\textbf{Section III-C}) and online pose tracking (\textbf{Section III-D}), respectively.}
	\label{Fig_2}
\end{figure*}

The detailed procedures (Fig. \ref{Fig_2}) of the proposed ERPoT are introduced in this section. At first, the LiDAR data is the only input information, and the corresponding sparse 2D scan is acquired through the point cloud preprocessing process. Afterward, the process of offline map construction is carried out, and the lightweight and compact polygon map of the environment can be obtained. Based on the above process, we can make use of the real-time 2D scan and the polygon map to realize effective and reliable pose tracking.

\subsection{Problem formulation}
The problem of pose tracking can be formulated as two main parts: offline map construction and online pose tracking. 

\subsubsection{Offline map construction} We propose a novel form of map representation called polygon map, which is more lightweight and more compact compared with traditional dense point cloud maps. To be specific, the input sensor data only includes the LiDAR data, which is represented by ${{\cal L}_{t}} = \{\mathbf{l}_1, \mathbf{l}_2,..., \mathbf{l}_{t}\}$ with $t$ being the timestamp, while the $i$-th frame $\mathbf{l}_i$ is denoted by the set $\{\mathbf{p}_1, \mathbf{p}_2, ..., \mathbf{p}_n\}$ with $n$ being the number of points, and $\mathbf{p}_i = \left[x_i, y_i, z_i\right]$.

On the other hand, the corresponding pose sequence ${{\cal S}_{t}} = \{\mathbf{s}_1, \mathbf{s}_2,..., \mathbf{s}_{t}\}$ is also required in map construction, and $\mathbf{s}_{i} = \left[x_i, y_i, \theta_i\right]$ with $\theta_i$ being the yaw angle. Note that ${{\cal S}_{t}}$ can be obtained through several LiDAR SLAM approaches or the provided ground truth from public datasets. More importantly, the polygon map is denoted by ${\cal M}_p$, which is composed of multiple polygons $\{\mathbf{P}_1, \mathbf{P}_2,..., \mathbf{P}_{m}\}$ with $m$ being the number of polygons. Therefore the joint posterior probability density function by time step $t$ is defined as
{
\begin{align}
	p\left( {\cal M}_{p}|{\cal L}_{t}, {{\cal S}_{t}}\right).
\end{align}}In practical applications, we cannot directly obtain the polygon map ${\cal M}_{p}$ from LiDAR data sequence ${{\cal L}_{t}}$ and pose sequence ${{\cal S}_{t}}$. Accordingly, let the occupancy grid map denoted as ${\cal M}_g$ as the intermediate form, and equation (1) can be rewritten as
{
\begin{align}
p\left( {{\cal M}_p|{\cal L}_t, {\cal S}_t} \right) = \int_{{\cal M}_g} {p\left( {\cal M}_p|{\cal M}_g \right) \cdot p\left( {{\cal M}_g|{\cal L}_t, {\cal S}_t} \right)},
\end{align}}where the occupancy grid map ${{\cal M}_g}$ is composed of independent grids with fixed size, the process of grid map generation can refer to Karto-SLAM \cite{Konolige}, while $p\left( {\cal M}_p|{\cal M}_g \right)$ represents the process of polygon generation. Before generating the 2D occupancy grid map, we need to convert 3D LiDAR points into 2D scan information via ground removal and obstacle selection, as shown in the middle subfigure in Fig. \ref{Fig_2}.

\subsubsection{Online pose tracking} Given the prior polygon map of the environment and the initial pose $\tilde{\mathbf{s}}_1$, then the process of pose tracking can be carried out. Specifically, the robot pose at timestamp $t$ is denoted by $\mathbf{s}_t$, and pose tracking result is denoted by the pose sequence ${\cal S}_{t} = \{\mathbf{s}_1, \mathbf{s}_2,...,\mathbf{s}_t\}$. In addition, let ${\cal L}_{t} = \{\mathbf{l}_1, \mathbf{l}_2,..., \mathbf{l}_{t}\}$ represent the corresponding LiDAR observation. Therefore the joint posterior probability density function by time step $t$ is defined as
\begin{align}
	p\left({\cal S}_{t}|\tilde{\mathbf{s}}_1, {{\cal L}_{t}}, {\cal M}_{p} \right) = &~p\left({\mathbf{s}_t}|{\cal S}_{t-1}, \mathbf{l}_t, {\cal M}_{p} \right) \cdot  \nonumber \\
	&~p\left({\cal S}_{t-1}|\tilde{\mathbf{s}}_1, {{\cal L}_{t - 1}},{\cal M}_{p} \right),
\end{align}
where $p\left({\cal S}_{t-1}|\tilde{\mathbf{s}}_1, {{\cal L}_{t - 1}},{\cal M}_{p} \right)$ denotes the probability distribution over pose sequence at timestamp $t-1$, while $p\left({\mathbf{s}_t}|{\cal S}_{t-1}, \mathbf{l}_t, {\cal M}_{p} \right)$ represents the probability distribution of the robot pose at timestamp $t$.

In the implementation, we make use of the constant velocity motion model $p\left(\mathbf{u}_{t-1}| \mathbf{s}_{t-1}, \mathbf{s}_{t-2} \right)$ to provide the initial guess $\tilde{\mathbf{s}}_{t}$ of the robot pose $p\left(\tilde{\mathbf{s}}_{t}| \mathbf{s}_{t-1}, \mathbf{u}_{t-1} \right)$, and $\mathbf{u}_{t-1}$ represents the motion change of the robot between timestamp $t-1$ and timestamp $t-2$. Therefore, the term $p\left({\mathbf{s}_t}|{\cal S}_{t-1}, \mathbf{l}_t, {\cal M}_{p} \right)$$(t > 2)$ in equation (3) can be rewritten as
\begin{align}
	p\left({\mathbf{s}_t}|{\cal S}_{t-1}, \mathbf{l}_t, {\cal M}_{p} \right) \triangleq p\left({\mathbf{s}_t}|{\mathbf{s}_{t-1}}, {\mathbf{s}_{t-2}}, \mathbf{l}_t, {\cal M}_{p} \right)  \nonumber \\
	= \int_{\tilde{\mathbf{s}}_t} p\left({\mathbf{s}_t}|\tilde{\mathbf{s}}_t, \mathbf{l}_t, {\cal M}_{p} \right) \cdot  p\left(\tilde{\mathbf{s}}_t| \mathbf{s}_{t-1}, \mathbf{s}_{t-2} \right).
\end{align}

\subsection{2D scan generation}
\begin{figure}[!htb]
	\centering
	\subfigure[]{\includegraphics[width=7.5cm]{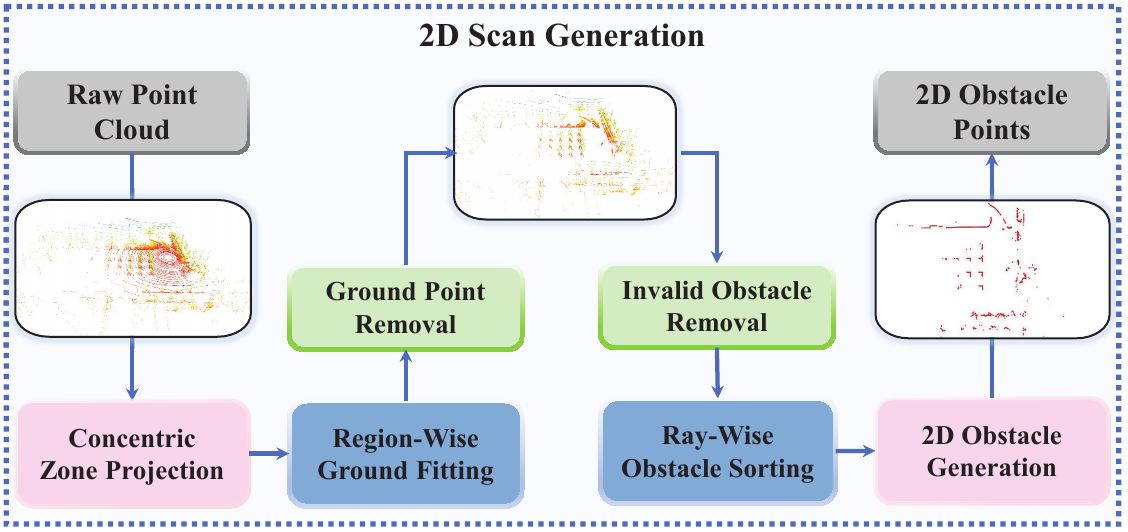}}
	\centering
	\subfigure[]{\includegraphics[width=8cm]{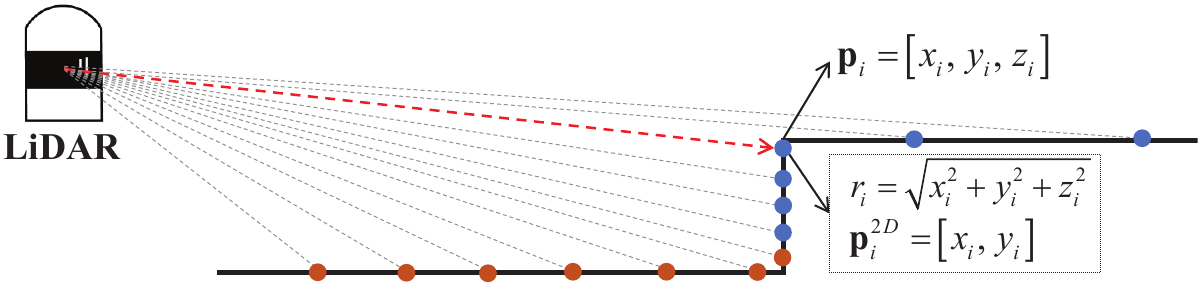}}
	\caption{(a) The process of 2D scan generation; (b) Ray-wise obstacle point selection. The orange point and the blue point represent the ground point and obstacle point, respectively. In addition, the obstacle point indicated by the red arrow is the final selection with the minimum range value.}
	\label{Fig_3}
\end{figure}

Considering that the workspace of mobile robots mainly lies on the ground, accurate and real-time three degrees of freedom pose estimation is enough to complete various complex tasks. Therefore, we convert the dense 3D obstacle points into sparse but representative 2D obstacle points, which is enough for mobile robot pose tracking. This section describes the detailed process of 2D scan generation, as shown in Fig. \ref{Fig_3}.

At first, the point cloud is projected into a fan-shaped grid map using the concentric zone model. Ground plane fitting \cite{SLee} is subsequently performed for each region to separate ground and obstacle points. Unlike previous work \cite{HGao} that emphasized obstacle point recall, we prioritize accurate pose tracking using precise environmental information. Unreliable obstacle points in regions with poor ground fitting and high-altitude points are removed to enhance accuracy.

Afterward, the valid 3D obstacle points are transformed into 2D obstacle points through ray-wise sorting and 2D generation. Specifically, the number $N$ of 2D obstacle points is determined by the LiDAR's horizontal angle resolution $\delta_a$,
\begin{align}
N = \lceil  {2 \pi / {\delta_a}} \rceil.
\end{align}
Accordingly, 2D obstacle points are derived from ray-wise sorting, with $N$ rays, each having multiple 3D points. For the $i$-th valid point $\mathbf{p}_i = [x_i, y_i, z_i]$, the range $r_i$ is calculated as shown in Fig. \ref{Fig_3}(b), and the horizontal angle $\theta_i$ and ray index $j$ are determined as follows
\begin{align}
	\theta_i = \mathbf{atan2}(y_i, x_i), j = \left\lfloor {\theta_i / \delta_a} \right\rfloor.
\end{align}

Based on the above, the ray $\mathbf{R}_{j}$ is denoted by $\{\mathbf{p}_1, \mathbf{p}_2, ..., \mathbf{p}_k\}$ with $k$ being the number of points. After nearest obstacle selection, the final 2D point $\mathbf{p}_i^{2D} = [x_i, y_i]$ is based on the nearest 3D point ($\mathbf{p}_i$ in Fig. \ref{Fig_3}(b)).

\subsection{Offline map construction}

Based on the obtained 2D scan information, the process of offline map construction is described in detail in this section.

\subsubsection{Grid map generation}
Let $\mathbf{C}_i$ denote the 2D scan, represented by $\mathbf{C}_i = \{\mathbf{p}_1^{2D}, \mathbf{p}_2^{2D}, ..., \mathbf{p}_{N}^{2D}\}$, corresponding to the sparse perception. While reliable static obstacles aid pose estimation, LiDAR noise and dynamic obstacles pose challenges. Additionally, efficiently fusing multiple observations, compactly representing the environment, and constructing the prior map are crucial tasks.

\begin{figure}[!htb]
	\centering
	\includegraphics[width=6.5cm]{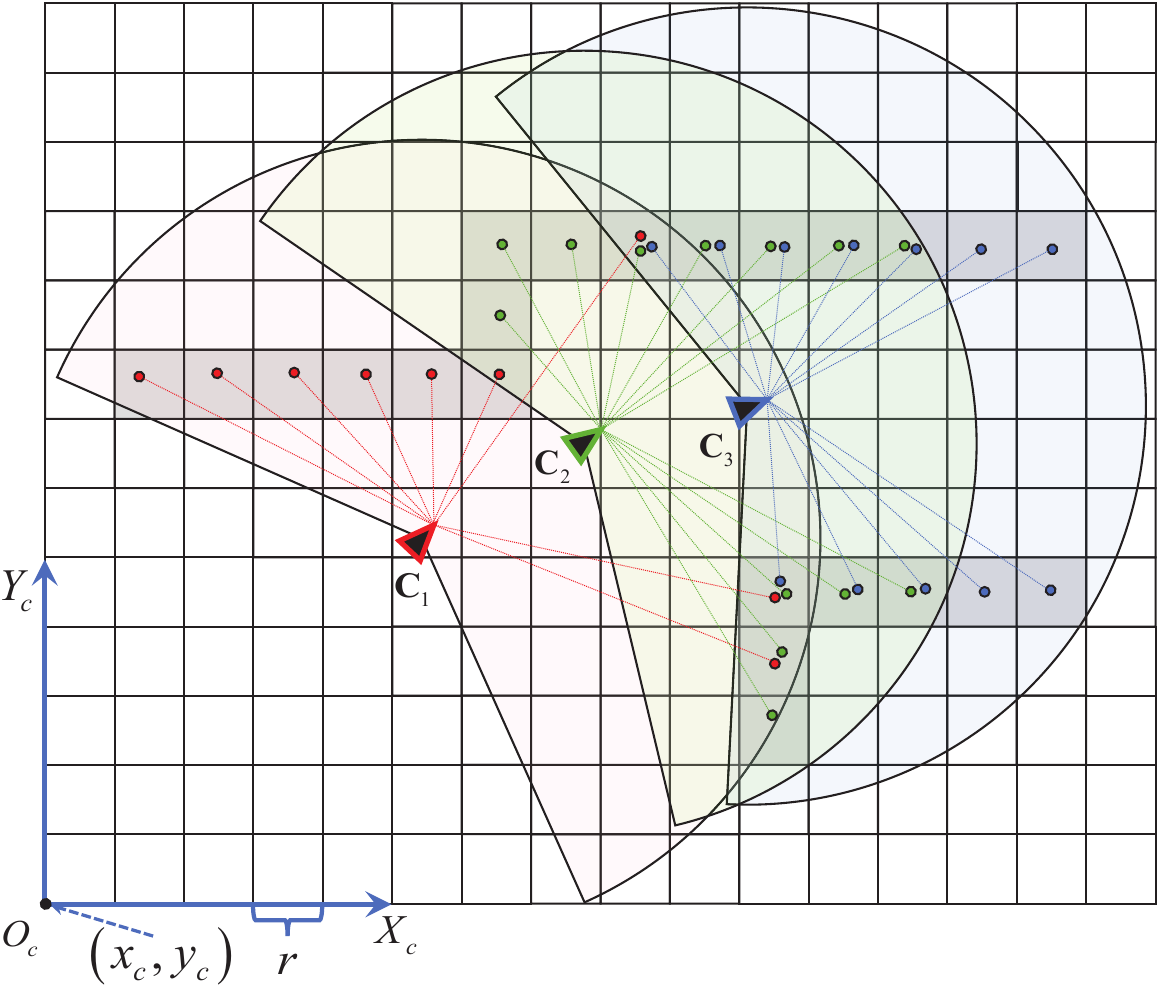}
	\caption{The process of occupancy grid map generation. Continuous 2D scan observations $\mathbf{C}_1$, $\mathbf{C}_2$, and $\mathbf{C}_3$ are denoted by the red, green, and blue triangles, respectively, and the fan-shaped area represents the observation range. In addition, scan points obtained from continuous observations are represented by the color circles.}
	\label{Fig_4}
\end{figure}

Accordingly, the occupancy grid map generation is introduced in Fig. \ref{Fig_4}, which is combined with the corresponding pose sequence ${{\cal S}_{t}}$. Therefore, equation (2) can be rewritten as
{
\begin{align}
	p\left( {{\cal M}_p|{\cal C}_t, {\cal S}_t} \right) = \int_{{\cal M}_g} {p\left( {\cal M}_p|{\cal M}_g \right) \cdot p\left( {{\cal M}_g|{\cal C}_t, {\cal S}_t} \right)},
\end{align}}where ${\cal C}_t = \{\mathbf{C}_1, \mathbf{C}_2, ..., \mathbf{C}_t\}$ denotes the 2D scan sequence. The construction problem of grid map ${\cal M}_g$ is converted into the state estimation problem of each grid as follows
{
\begin{align}
	p\left({\cal M}_{g}|{\cal C}_t, {\cal S}_t \right) = \prod\limits_{i = 1}^{N_{g}} {p\left( {\cal C}_t, {\cal S}_t \right)},
\end{align}}where $g_i$ is the $i$-th grid, and $N_g$ is the total number of grids. As shown in Fig. \ref{Fig_4}, the ray trace algorithm updates the grid map by connecting scan points to the sensor origin. Specifically, each grid's hits $N_{hit}$ and passes $N_{pass}$ are counted, and the occupancy probability of grid $g_i$ is calculated as
{
\begin{align}
		p\left(g_i\right)=\frac{N_{hit}}{N_{hit} + N_{pass}}.
\end{align}
}

In the generation process, the size and the state of the grid map gradually update as new areas are observed. Finally, the occupancy probabilities of all grids are calculated, and each grid determines whether the current state is occupied based on the given threshold.

\subsubsection{Polygon map construction}
After the above generation process of the grid map, the binary image representing the static obstacles in the environment can be obtained. On this basis,  the contour extraction process of the grid map is carried out to achieve a lightweight and compact representation.

\begin{figure}[!t]
	\centering
	\subfigure[Binary image]{\includegraphics[width=3.6cm]{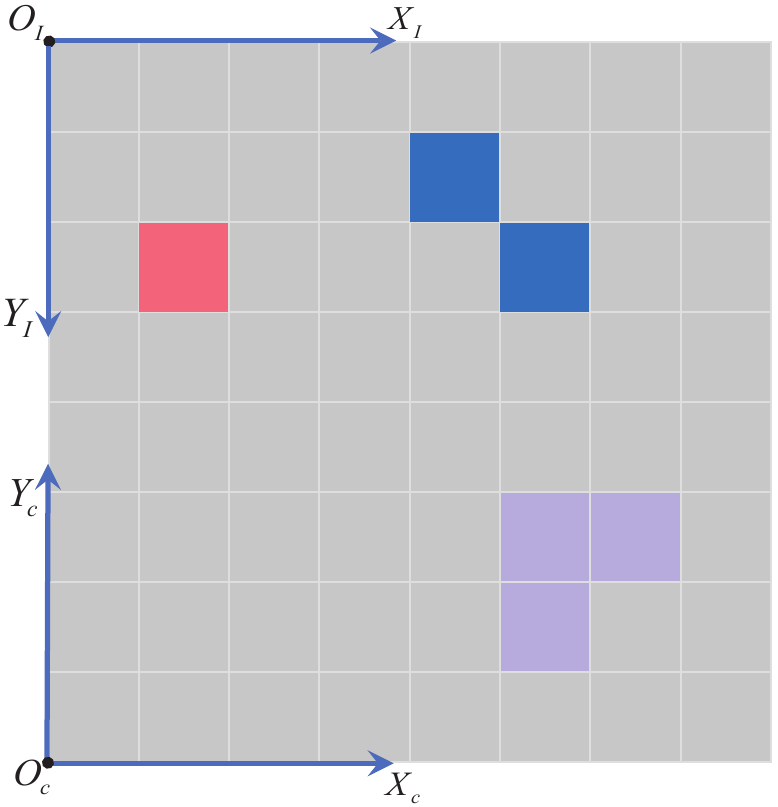}}
	\centering
	\subfigure[Dilation operation]{\includegraphics[width=3.6cm]{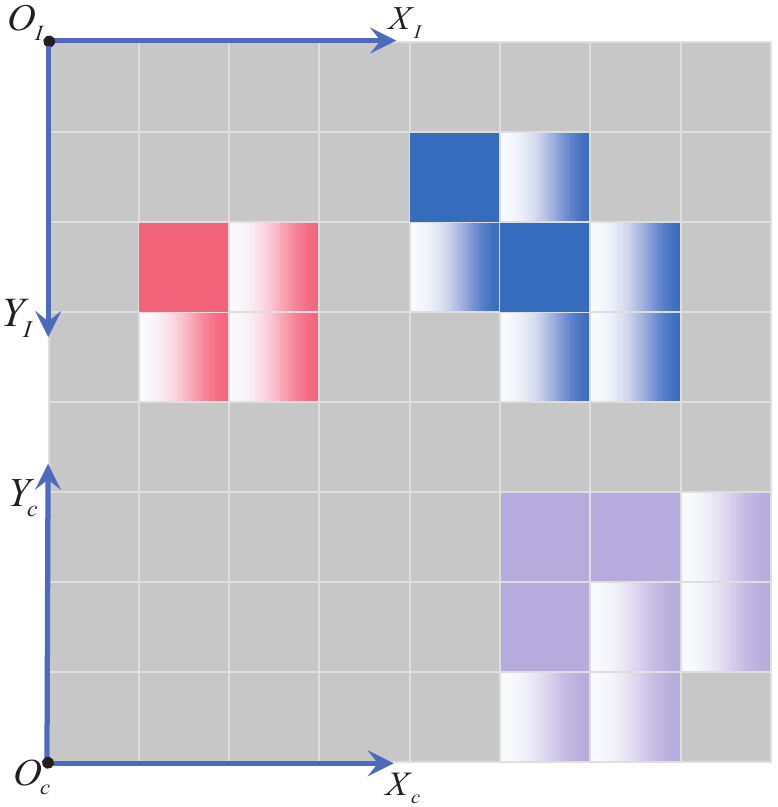}}
	\centering
	\subfigure[Contour extraction]{\includegraphics[width=3.6cm]{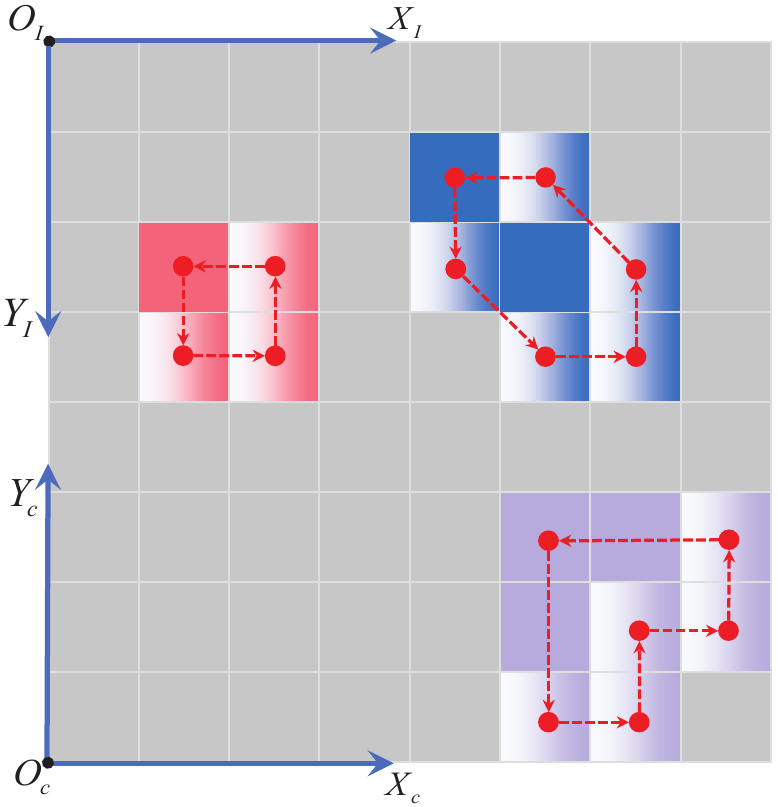}}
	\centering
	\subfigure[Contour translation]{\includegraphics[width=3.6cm]{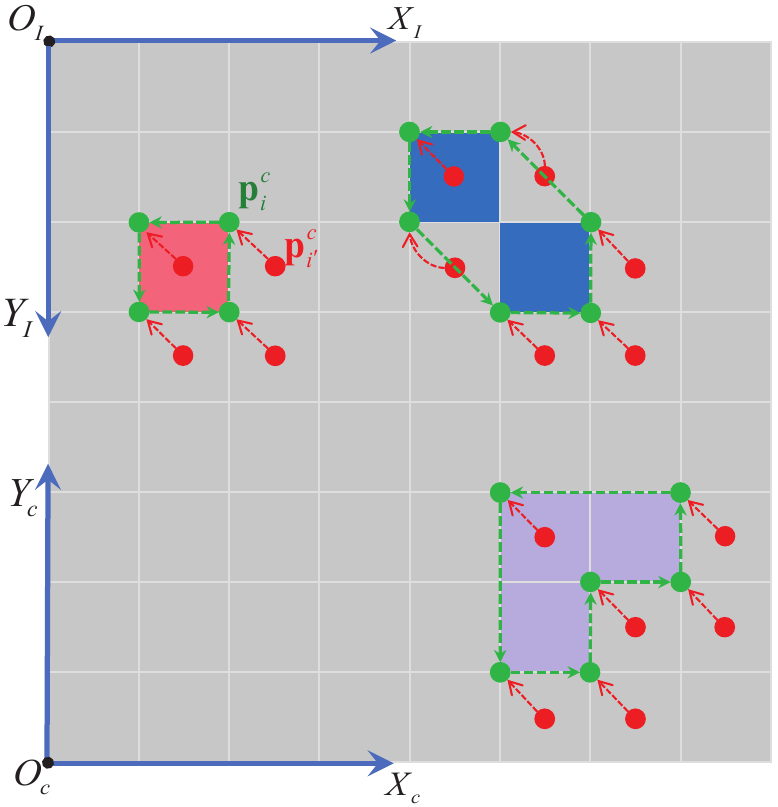}}
	\caption{The process of contour extraction for the grid map. The grids with color indicate the occupancy status, and red filled circles and green filled circles represent the contour vertices.}
	\label{Fig_5}
\end{figure}

In Fig. \ref{Fig_5}(a), the obtained grid map is transferred from the grid map coordinate system to the image coordinate system, denoted by $X_c \text{-} O_c \text{-} Y_c$ and $X_I \text{-} O_I \text{-} Y_I$, respectively, and three different obstacles are taken to describe the contour extraction process in detail. The process of vectorized representation can be divided into three steps:

(1) As shown in Fig. \ref{Fig_5}(b), firstly, the original binary map is processed by dilation operation, and kernel size is $2$, then the newly obtained binary map is used for contour extraction.

(2) Subsequently, the contour extraction process using the newly obtained binary map is performed, with the results depicted in Fig. \ref{Fig_5}(c) as red and directional contours.

(3) Moreover, the red contour in Fig. \ref{Fig_5}(c) does not match the actual obstacle information. The accurate final contours are obtained through contour translation, as shown in Fig. \ref{Fig_5}(d).

Furthermore, using points $\mathbf{p}_{i'}^{c}$ and $\mathbf{p}_{i}^{c}$ as an example, we explain how to obtain the contour vertex coordinates. Specifically, the image coordinates of $\mathbf{p}_{i'}^{c}$ are $(\text{row}_i,\text{col}_i)$, representing the row and column indexes. In the grid map coordinate system, $\mathbf{p}_{i'}^{c} = \left[x_{i'}^c, y_{i'}^c\right]$ is calculated as follows
\begin{align}
	&x_{i'}^c = r \cdot ( \text{row}_i + 0.5 ), \\
	&y_{i'}^c = r \cdot ( {N_{\text{col}} - \text{col}_i - 0.5}),
\end{align}where $N_{\text{col}}$ denotes the number of columns. After the contour translation process, $\mathbf{p}_{i}^{c} = \left[x_{i}^c, y_{i}^c\right]$ is obtained as
\begin{align}
	&x_{i}^c = x_{i'}^c - 0.5 \cdot r = r \cdot {\text{row}_i}, \\
	&y_{i}^c = y_{i'}^c + 0.5 \cdot r = r \cdot ( {N_{\text{col}} - \text{col}_i}).
\end{align}Taking into account the offset information of the grid map, the corresponding point coordinate $\mathbf{p}_{i}^{w} = \left[x_{i}^w, y_{i}^w\right]$ in the world coordinate system is calculated as follows
\begin{align}
	x_{i}^w = x_{i}^c + x_c, y_{i}^w = y_{i}^c + y_c.
\end{align}

\begin{algorithm}[!b]\scriptsize
	\caption{\small Polygon segmentation}
	\begin{algorithmic}[1]
		\Require Original contour ${\mathbf{O}_k } = \{\mathbf{p}_{1}^{w}, \mathbf{p}_{2}^{w}, ,..., \mathbf{p}_{N_k}^{w}\}$, vertex number threshold $N_{v}$, and the segmentation distance threshold $D_{s}$.
		\Ensure Polygon set ${\cal P} = \{{\mathbf{P}}_1, \mathbf{P}_2, ...,{\mathbf{P}}_l\}$ with $l$ being the number of polygons.
		\State The number of vertices ${\emph{N}}_k$ = size($\mathbf{O}_k$).
		\If{${\emph{N}}_k < N_{v}$}
		\State ${\cal P} \leftarrow {\cal P} \cup \{\mathbf{O}_k\}$
		\State Return
		\EndIf
		\State Initialize the open\_contour queue ${\cal O}_{open}$ with $\mathbf{O}_k$
		\While{${\cal O}_{open}$ is not empty}
		\State Obtain current open\_contour ${\mathbf{O}_c} \leftarrow {\cal O}_{open}$.front(), ${\cal O}_{open}$.pop\_front().
		\State $N_{c}$ = size($\mathbf{O}_c$), and let $\delta$ denote the index interval, $\delta \leftarrow \left\lfloor N_{c} / 2 \right\rfloor$.
		\State Let ${\mathbf{O}_l}$ and ${\mathbf{O}_r}$ represent two contours after segmentation.
		\State Let ${D_{min}}$ and ${I_{min}}$ denote the min distance and the corresponding index,
		\State and ${D_{min}}$ is initialized with maximum value.
		\State $flag_S = False$, indicating whether segmentation has been performed.
		\While{$\delta \geqslant 2$}
		\For{$i = 1 \to (N_{c} - \delta)$}
		\State $I_1 \leftarrow i$, $I_2 \leftarrow (i + \delta)$
		\State Obtain two corresponding points $\mathbf{p}_{I_1}^w$ and $\mathbf{p}_{I_2}^w$
		\State $flag_{I} = \mathbf{JudgeInteraction}(\mathbf{p}_{I_1}, \mathbf{p}_{I_2}, \mathbf{O}_c)$
		\If{$flag_{I} == True$}
		\State Continue
		\EndIf
		\If{$\mathbf{DistancePoint}(\mathbf{p}_{I_1}^w, \mathbf{p}_{I_2}^w) < D_{min}$}
		\State $D_{min} \leftarrow \mathbf{DistancePoint}(\mathbf{p}_{I_1}^w, \mathbf{p}_{I_2}^w)$, $I_{min} \leftarrow i$
		\EndIf
		\EndFor
		\If{$D_{min} < D_{s}$}
		\State $flag_S = True$
		\State $[{\mathbf{O}_l}, {\mathbf{O}_r}] \leftarrow \mathbf{PolySegmentation}(\mathbf{p}_{I_{min}}^w, \mathbf{p}_{I_{min} + \delta}^w, \mathbf{O}_c)$
		\If{size($\mathbf{O}_l$) $ < N_{v}$}
		\State ${\cal P} \leftarrow {\cal P} \cup \{\mathbf{O}_l\}$
		\Else
		\State ${\cal O}_{open}$.emplace(${\mathbf{O}_l})$
		\EndIf
		\If{size($\mathbf{O}_r$) $< N_{v}$}
		\State ${\cal P} \leftarrow {\cal P} \cup \{\mathbf{O}_r\}$
		\Else
		\State ${\cal O}_{open}$.emplace(${\mathbf{O}_r}$)
		\EndIf
		\EndIf
		\State $\delta \leftarrow \left\lfloor \delta / 2 \right\rfloor$
		\EndWhile
		\If{$flag_S == False$}
		\State ${\cal P} \leftarrow {\cal P} \cup \{\mathbf{O}_c\}$
		\EndIf
		\EndWhile
	\end{algorithmic}
\end{algorithm}

\begin{figure}[!htb]
	\centering
	\subfigure[Original contour]{\includegraphics[width=4.3cm]{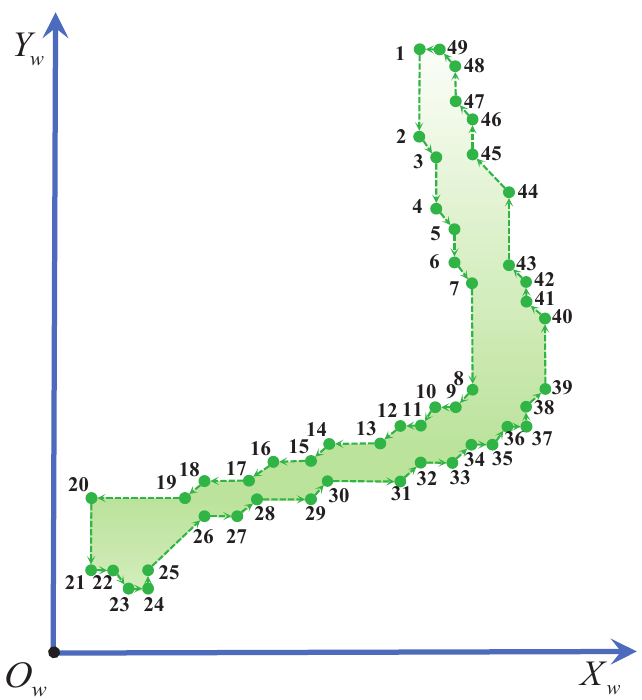}}
	\centering
	\subfigure[First polygon segmentation]{\includegraphics[width=4.3cm]{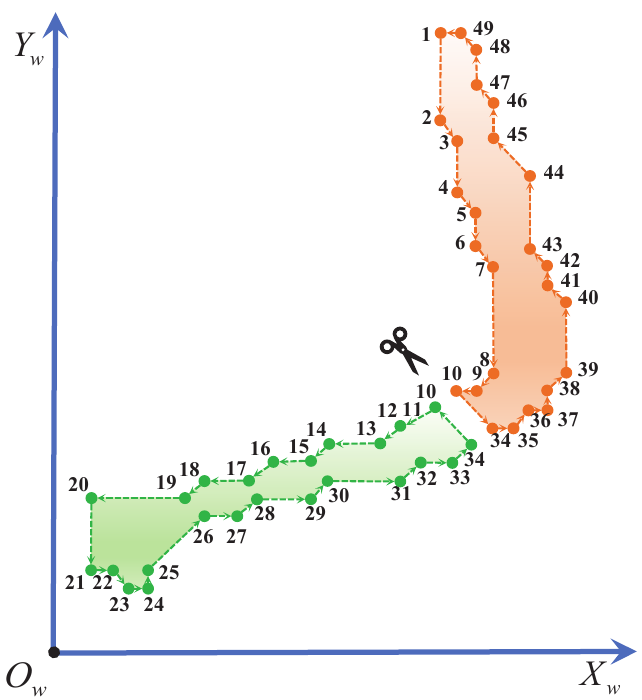}}
	\centering
	\subfigure[Second polygon segmentation]{\includegraphics[width=4.3cm]{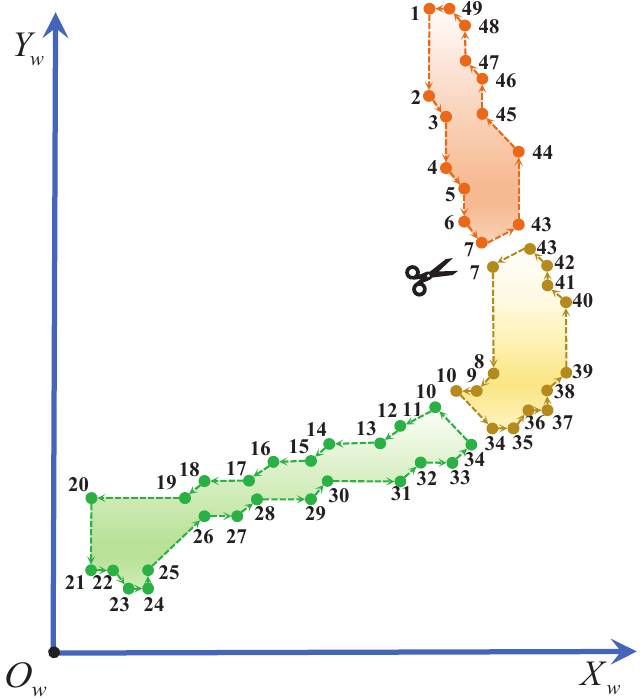}}
	\centering
	\subfigure[Third polygon segmentation]{\includegraphics[width=4.3cm]{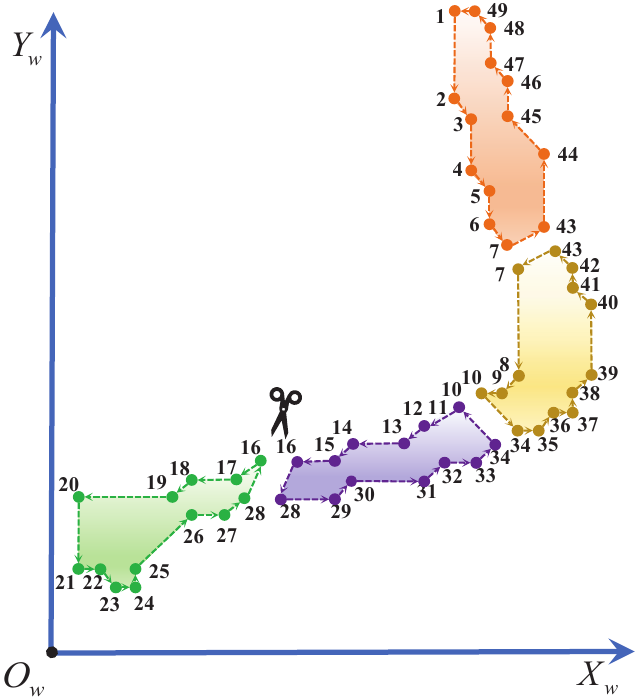}}
	\caption{The process of polygon segmentation. Specifically, the filled circles represent the vertices of the polygons, dashed lines represent the contour lines, color filled areas represent the polygon regions, and numbers refer to the index order. After three times of polygon segmentation (\emph{i.e.}, a$\rightarrow$b$\rightarrow$c$\rightarrow$d), the original contour is split into four non-intersecting polygons.}
	\label{Fig_6}
\end{figure}

Based on the above, a contour set ${\cal O} = \{\mathbf{O}_1, \mathbf{O}_2\, ..., \mathbf{O}_{N_{\cal O}}\}$ is obtained for the environment, and $N_{\cal O}$ is the number of contours, while $\mathbf{O}_k$ is the $k$-th contour and represented by the vertex set $\{\mathbf{p}_{1}^{w}, \mathbf{p}_{2}^{w}, ,..., \mathbf{p}_{N_k}^{w}\}$ with $N_k$ being the number of vertices. However, elongated and irregular contours can cause interference in subsequent data association and matching, further segmentation of the contours containing a large number of vertices into multiple small polygons is necessary. Accordingly, the proposed polygon segmentation process is shown in \textbf{Algorithm 1}, taking the original contour as input, the output is a number of polygons with fewer vertices than a given threshold. To be specific, the core process is shown in {\bf lines 15-39}, and the function $\mathbf{JudgeInteraction}(\cdot)$ is used to judge whether the $\mathbf{Line}(\mathbf{p}_{I_1}, \mathbf{p}_{I_2})$ intersects with any edges on the contour. In addition, the contour is segmented into two polygons through the function $\mathbf{PolySegmentation}(\cdot)$, as shown in Fig. \ref{Fig_6}(b).

In particular, as depicted in Fig. \ref{Fig_6}(a), the example of polygon segmentation involves a contour with dozens of vertices arranged in counter-clockwise order. In the initial segmentation step, the vertices denoted as $\mathbf{p}_{10}^w$ and $\mathbf{p}_{34}^w$ serve as endpoints of the segmentation line, effectively dividing the contour into two roughly equal parts. This result is illustrated in Fig. \ref{Fig_6}(b). Ultimately, following two iterations of polygon segmentation, the original contour is divided into four distinct polygons, as illustrated in Fig. \ref{Fig_6}(d). After polygon segmentation and simplification, we make use of the polygon map ${\cal M}_p = \{\mathbf{P}_1, \mathbf{P}_2\, ..., \mathbf{P}_{m}\}$ to represent the environmental occupancy, and the $i$-th $\mathbf{P}_i$ is also denoted by the vertex set $\{\mathbf{p}_{1}^{w}, \mathbf{p}_{2}^{w}, ,..., \mathbf{p}_{N_i}^{w}\}$ with $N_i$ being the number of vertices. Note that throughout the entire process of polygon segmentation and subsequent pose tracking, the counter-clockwise characteristic of the polygon vertices is preserved, as shown in Fig. \ref{Fig_6}.

\subsubsection{Prior map representation}
Given the polygon map ${\cal M}_p$, before commencing with pose tracking, the process of prior map representation is carried out, and this step is crucial for establishing a solid foundation for accurate tracking. Specifically, for the $i$-th polygon $\mathbf{P}_i = \{\mathbf{p}_1^w, \mathbf{p}_2^w, ..., \mathbf{p}_{N_i}^w\}$, the corresponding centroid point $\mathbf{p}_i^e = [x_i^e, y_i^e]$ is calculated as
\begin{align}
	x_i^e = \frac{m_{10}}{m_{00}}, \  y_i^e = \frac{m_{01}}{m_{00}},
\end{align}where $m_{10}$ and $m_{01}$ represent the first-order moments about the x-coordinate and the y-coordinate, respectively, while $m_{00}$ is the zeroth-order moment indicating the total area of $\mathbf{P}_i$. Therefore, we can calculate the corresponding centroid points for all polygons and form a centroid point cloud $\mathbb{C}_e = \{\mathbf{p}_1^e, \mathbf{p}_2^e, ..., \mathbf{p}_m^e\}$, accelerating the pose tracking process. On the other hand, all vertices contained in ${\cal M}_p$ form a vertex point cloud $\mathbb{C}_v = \{\mathbf{p}_1^v, \mathbf{p}_2^v, ..., \mathbf{p}_{N_v}^v\}$ with $N_v$ being the number, and duplicate vertices are removed.

\subsection{Online pose tracking}
Once the prior map of the environment has been obtained, it can be utilized as prior information to facilitate effective and reliable pose tracking. Specifically, the newly proposed pose tracking approach encompasses the following detailed processes: feature extraction, point-polygon matching, and real-time pose tracking.

\subsubsection{Feature extraction}
Different from the previous offline map construction, the obtained 2D scan information is not immediately applied to pose tracking. Instead, it is processed to generate a more representative set of features with fewer elements. On the other hand, LiDAR odometry based on the LOAM (Lidar Odometry and Mapping) approach has achieved significant success in the field of mobile robotics and autonomous driving. Capitalizing on this established groundwork, we have adopted the LiDAR feature extraction process. 

However, rather than extracting LiDAR features from dense point clouds, we opt to extract features from the generated 2D scan data, following processes of ground removal and obstacle selection. Specifically, for the $i$-th 2D scan $\mathbf{C}_i = \{\mathbf{p}_1^{2D}, \mathbf{p}_2^{2D}, ..., \mathbf{p}_{N}^{2D}\}$, the smoothness value $c_j$ of the $j$-th scan point is calculated with its neighbour scan points $\cal S$,
\begin{align}
c_j = \left\| {\sum\limits_{k \in {\cal S},k \ne j} {\left( {r_k^{2D} - r_j^{2D}} \right)} } \right\|^2,\\
r_k^{2D} = \sqrt{(x_k)^2 + (y_k)^2}.
\end{align}

\begin{figure}[!htb]
	\centering
	\subfigure[Original obstacle points]{\includegraphics[width=4.2cm]{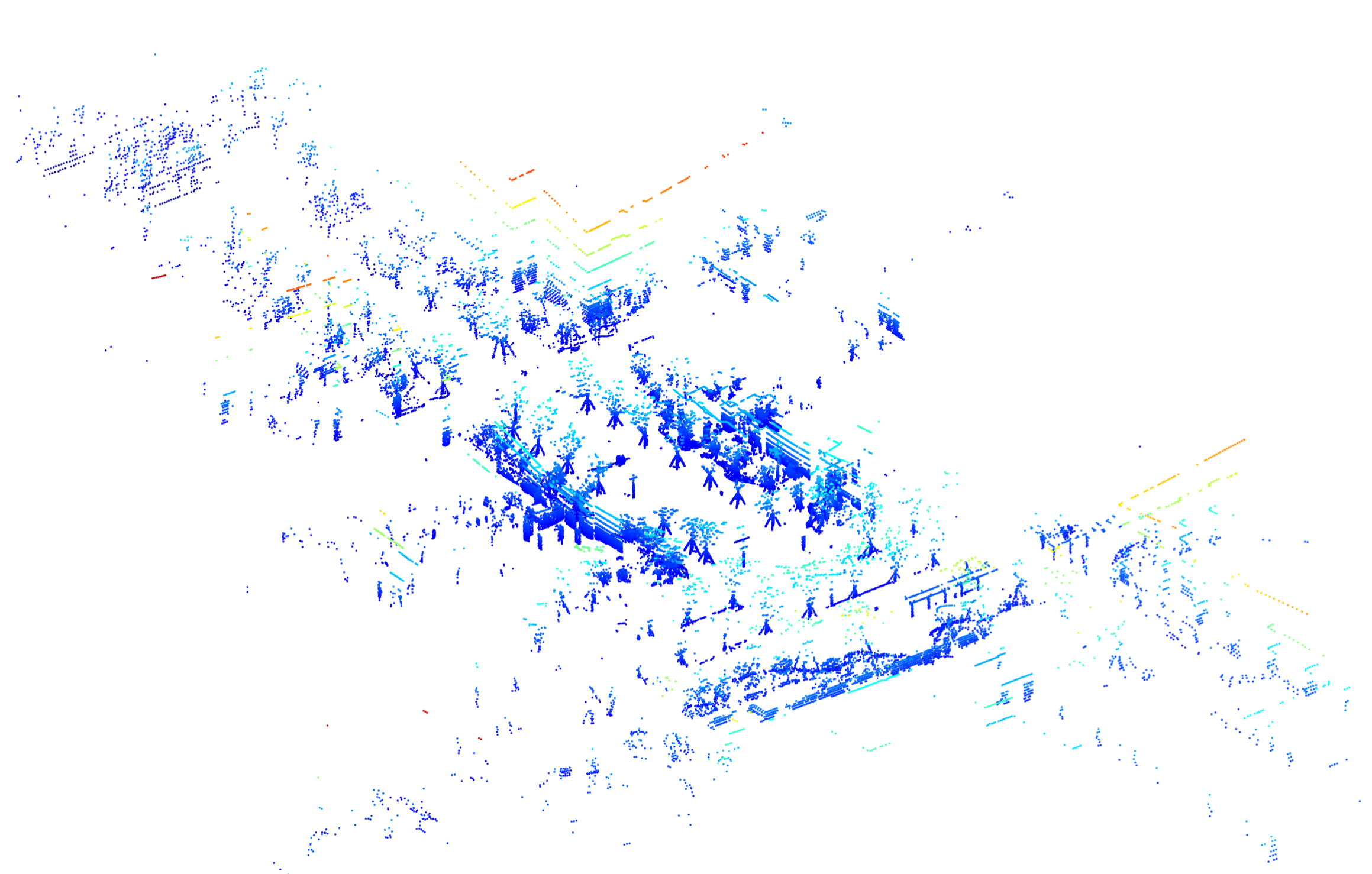}}
	\centering
	\subfigure[Ray-wise obstacle points]{\includegraphics[width=4.2cm]{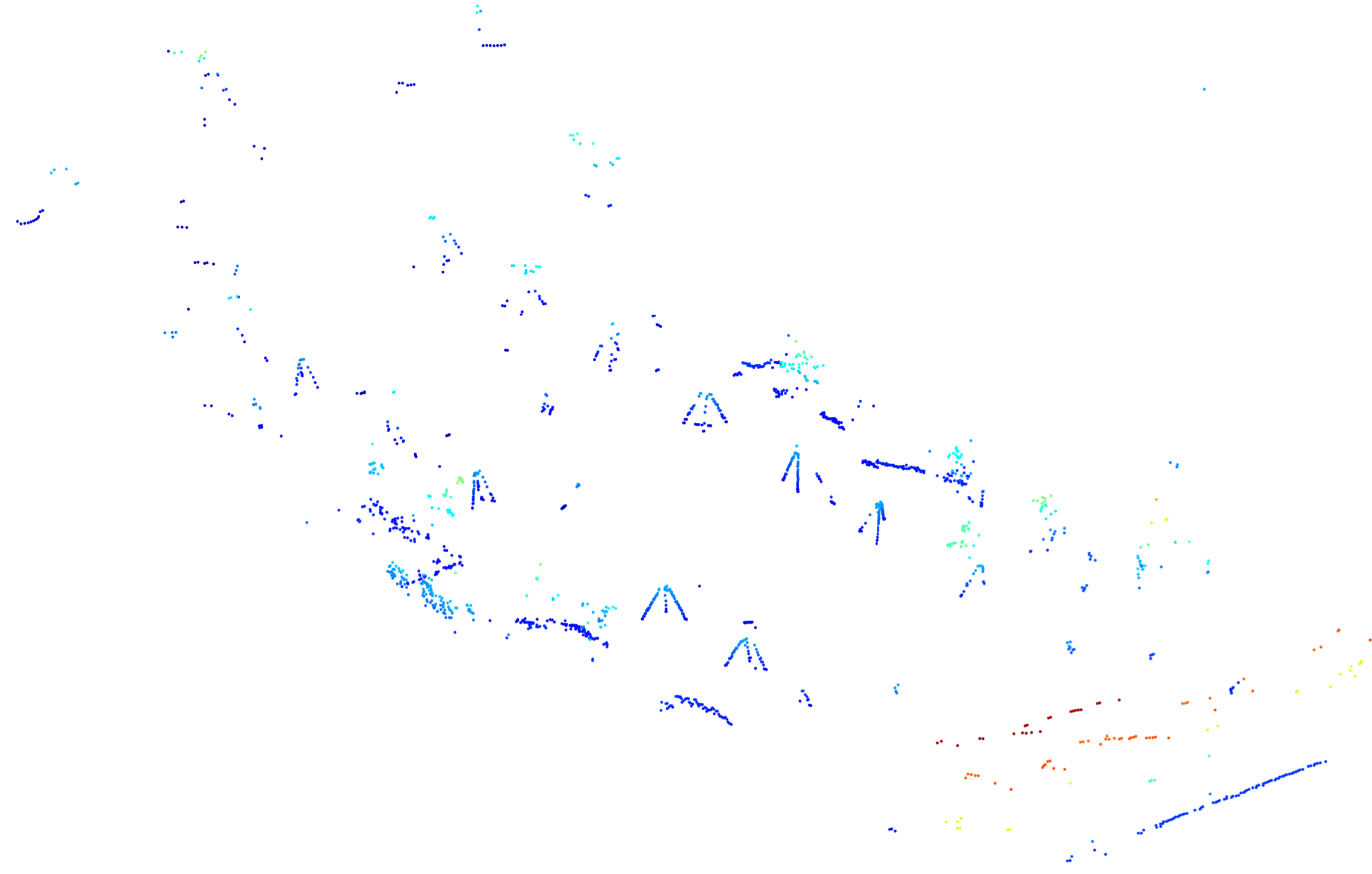}}
	\centering
	\subfigure[Representative feature points]{\includegraphics[width=8cm]{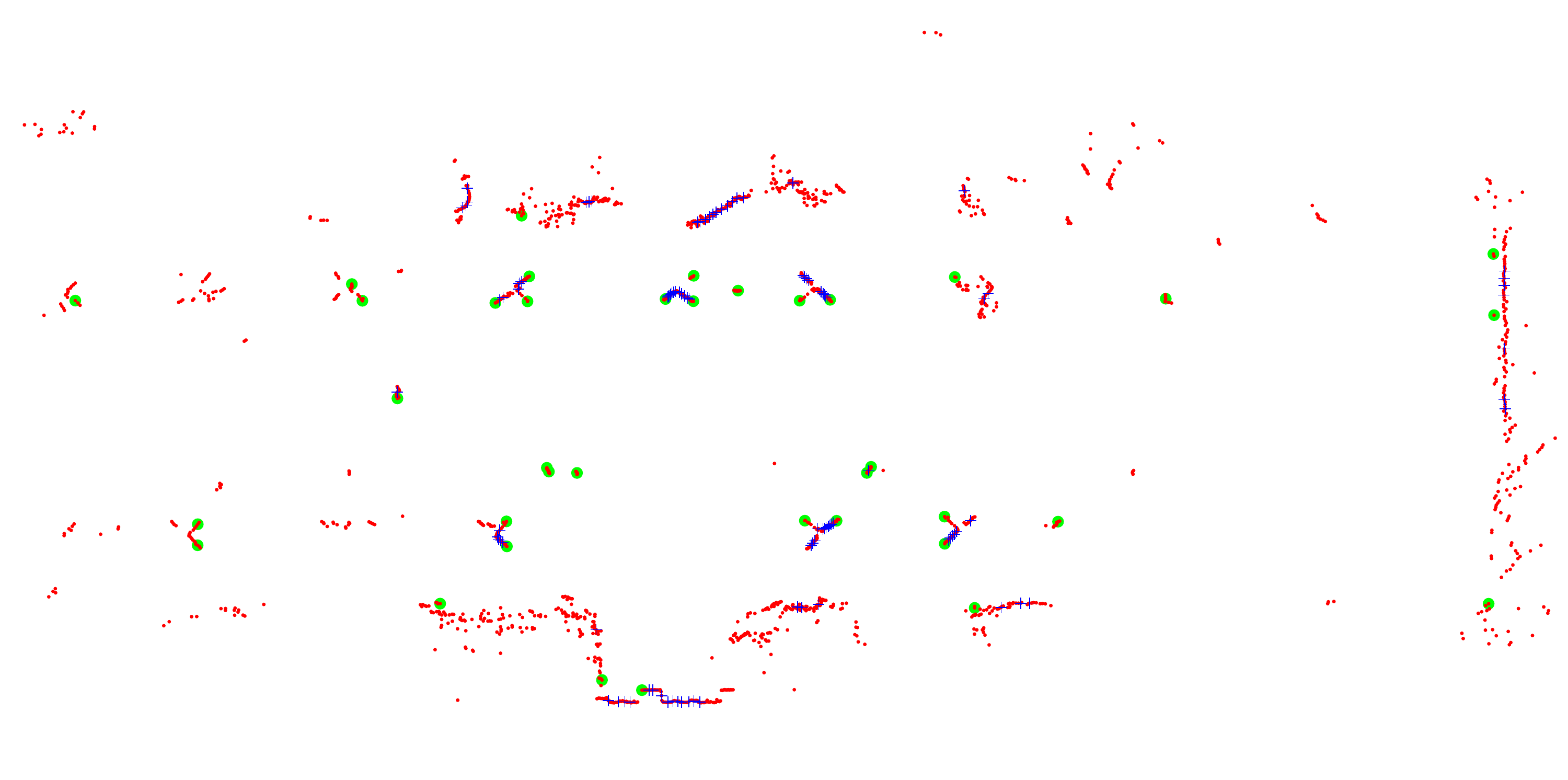}}
	\caption{The process of sparse feature extraction. The original scan points are denoted by the red filled circles in (c), which are derived from the ray-wise obstacle points in (b), while the original obstacle points are illustrated in (a). In addition, the filled green circle and the blue cross represent the corner feature point and the edge feature point, respectively.}
	\label{Fig_7}
\end{figure}

Afterward, according to the given smoothness threshold ($\sigma_c$ and $\sigma_e$), the corner feature set $\mathbb{F}_c$ and edge feature set $\mathbb{F}_e$ can be obtained. Specifically, we exclusively employ the derived single-line scan for feature extraction, while corner features and edge features are identified as scan points exhibiting relatively lower and higher smoothness values, respectively, as shown in Fig. \ref{Fig_7}. Note that the obtained nearest obstacle points (Fig. \ref{Fig_7}(b)) retain their x-coordinates and y-coordinates, thereby compressing them into a 2D scan (Fig. \ref{Fig_7}(c)) for feature extraction, which is different from the conventional methodologies employed in 3D LiDAR odometry.

\subsubsection{Point-polygon matching}

Based on the prior map obtained through polygon map construction, the point-polygon matching process is executed in this subsection, leveraging two point clouds denoted as ($\mathbb{C}_e$ and $\mathbb{C}_v$), as well as two sets of feature points ($\mathbb{F}_c$ and $\mathbb{F}_e$). In addition, the polygon map ${\cal M}_p$ with counter-clockwise polygon vertices is also used for the point-to-edge matching step. Note that before implementing a matching process, it is necessary to ensure that the current number of observed corner features and edge features are greater than the thresholds $\tau_c$ and $\tau_e$, respectively.

Specifically, the polygon map form utilized in this study is actually a lightweight and accurate description of environmental occupancy information. Consequently, during the processes of feature matching or pose estimation, it is essential that the feature points within the current 2D scan align with or closely conform to the polygons as much as possible.

Given the prior pose $\tilde{\mathbf{s}}_t = [\tilde{x}_t, \tilde{y}_t, \tilde{\theta}_t]$ regarding the current observation, it represents an imprecise estimation. Accordingly, we must formulate a cost function $\mathbf{F}(\tilde{\mathbf{s}}_t, \mathbb{F}_c, \mathbb{F}_e, \mathbb{C}_v, \mathbb{C}_e, {\cal M}_p)$ that incorporates the current features, pose hypotheses, and the prior map to accurately determine current pose
\begin{align}
\mathbf{s}_t^* = \arg \operatorname{min}\mathbf{F}(\tilde{\mathbf{s}}_t, \mathbb{F}_c, \mathbb{F}_e, \mathbb{C}_v, \mathbb{C}_e, {\cal M}_p),
\end{align}where $\mathbf{s}_t^*$ is the optimal estimation of the current pose, and the cost function is represented as follows
\begin{align}
&\mathbf{F}(\tilde{\mathbf{s}}_t, \mathbb{F}_c, \mathbb{F}_e, \mathbb{C}_v, \mathbb{C}_c, {\cal M}_p) \\ \nonumber
&\triangleq {\mathbf{F}_{c}(\tilde{\mathbf{s}}_t, \mathbb{F}_c, \mathbb{C}_v)} + \mathbf{F}_{e}(\tilde{\mathbf{s}}_t, \mathbb{F}_e, \mathbb{C}_e, {\cal M}_p),
\end{align}where $\mathbf{F}_{c}(\cdot)$ and $\mathbf{F}_{e}(\cdot)$ represent the matching cost functions for point-to-vertex and point-to-edge, respectively. It is noteworthy that this study introduces an innovative approach, transforming the matching relationship between feature points and polygons into two distinct types of matching: point-to-vertex and point-to-edge, thereby enhancing the precision and adaptability of the feature matching technique.

\emph{\textbf{Point-to-Vertex}}: In the realm of point-to-vertex matching, our methodology commences by constructing a KD-tree based on $\mathbb{C}_v$. Subsequently, assuming that the closest vertex retrieved from the KD-tree represents the accurate correspondence for each corner feature point, provided that the distance from the closest vertex is less than the given threshold $\epsilon_v$.

\begin{figure}[!t]
	\centering
	\includegraphics[width=7.5cm]{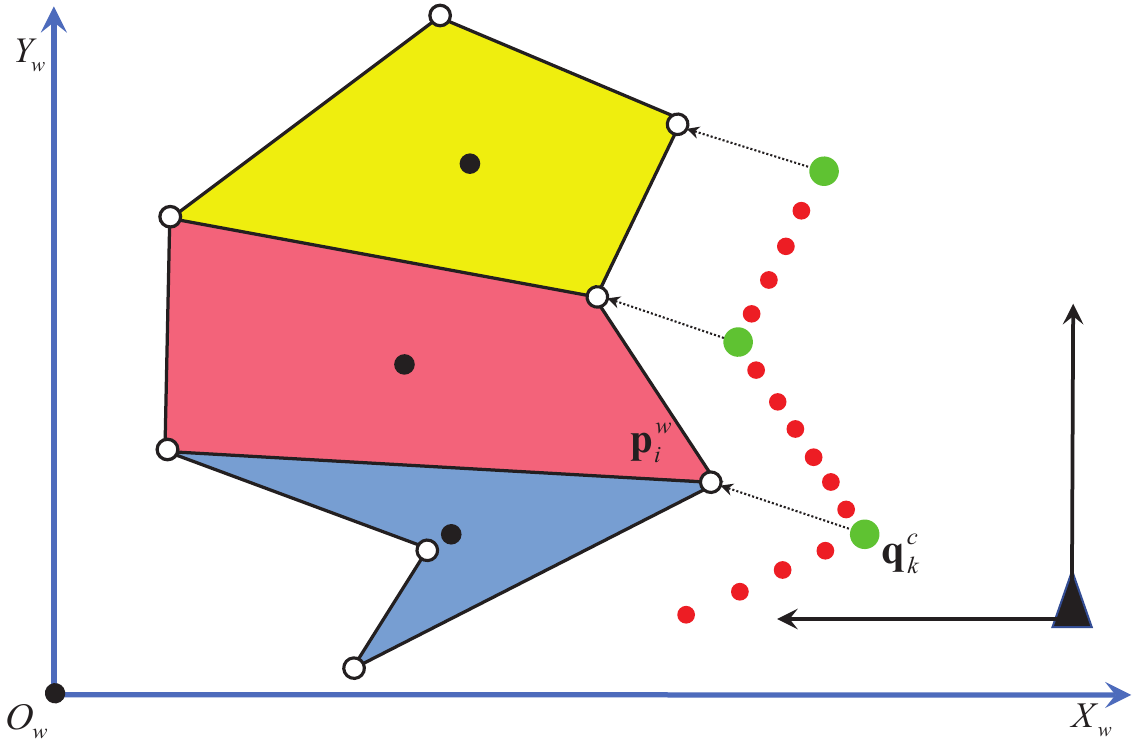}
	\caption{The schematic illustration of data association for point-to-vertex. The polygon represents the occupancy information in ${\cal M}_p$, with its vertices indicated by white circles, and the black filled circle is the centroid point. In addition, the larger green circle represents the extracted corner feature point.}
	\label{Fig_8}
\end{figure}

For the retrieved closest vertex, all polygons containing these vertices are recalled. Afterward, the spatial relationship between the current feature point and all polygons is analyzed. In instances where the point is found to be enclosed by a polygon, it will be excluded from the subsequent phase of cost function formulation. Otherwise, as shown in Fig. \ref{Fig_8}, the corner feature point $\mathbf{q}_{{k}}^c \in \mathbb{F}_c$ corresponds the vertex $\mathbf{p}_{{k \to i}}^w \in \mathbb{C}_v$, and these two points form a matching relationship,
\begin{align}
\mathbf{p}_{{k \to i}}^w = \tilde{\emph{\textbf{R}}}_t \cdot \mathbf{q}_{{k}}^c + \tilde{\emph{\textbf{t}}}_t,
\end{align}where
\begin{align}
\tilde{\emph{\textbf{R}}}_t = \left[ {\begin{array}{*{20}{c}}
		{\cos ( {{\tilde{\theta} _t}})}&{ - \sin ( {{\tilde{\theta} _t}} )} \\ 
		{\sin ( {{\tilde{\theta} _t}} )}&{\cos ( {{\tilde{\theta} _t}} )} 
\end{array}} \right],
\tilde{\emph{\textbf{t}}}_t = \left[ {\begin{array}{*{20}{c}}
		{{\tilde{x}_t}} \\ 
		{{\tilde{y}_t}} 
\end{array}} \right].
\end{align}Accordingly, the mathematical expression of the Euclidean distance of matching pairs is as follows
\begin{align}
{\mathbf{F}_{c}(\tilde{\mathbf{s}}_t, \mathbb{F}_c, \mathbb{C}_v)} = \sum\limits_{k = 1}^{{N_c}} {{{\left\| {{\mathbf{p}}_{k \to i}^w - \tilde{\emph{\textbf{R}}}_t \cdot {\mathbf{q}}_k^c - \tilde{\emph{\textbf{t}}}_t} \right\|}^2}},
\end{align}where ${N_c}$ denotes the number of valid matching pairs formed by corner feature points.

\begin{figure}[!htb]
	\centering
	\includegraphics[width=8cm]{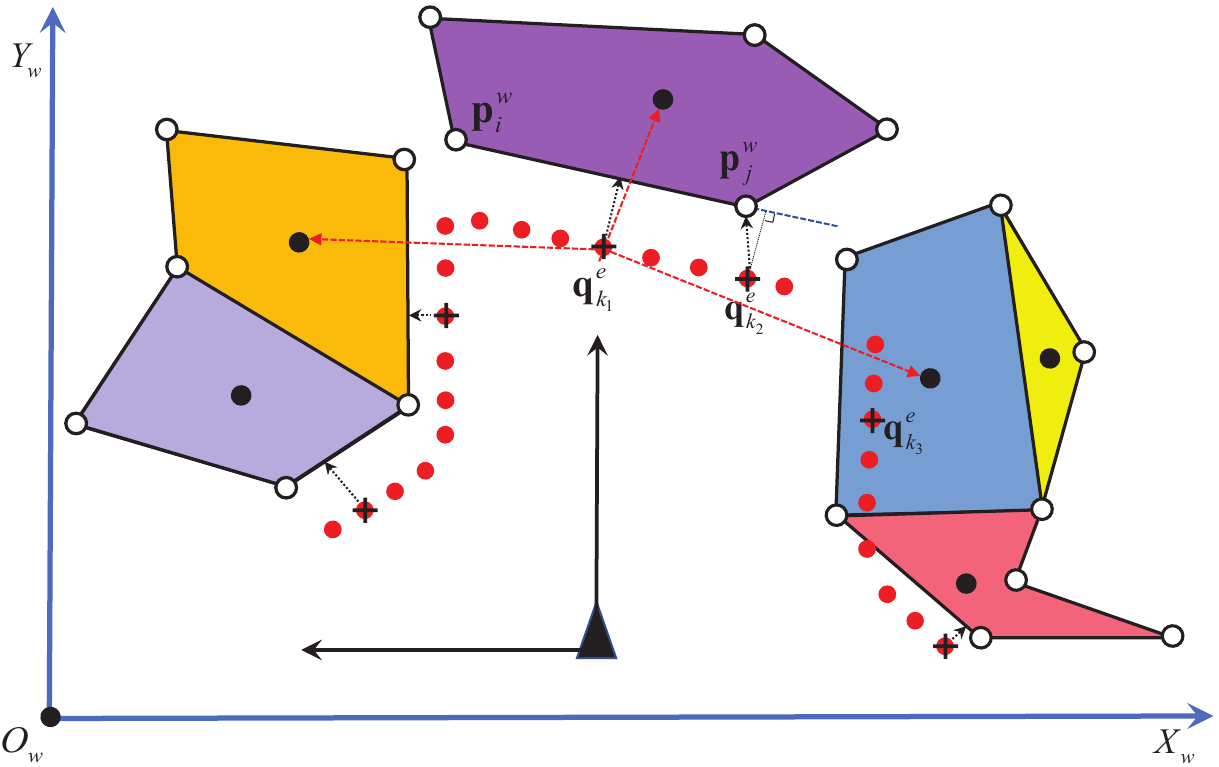}
	\caption{The schematic illustration of data association for point-to-edge. The polygon represents the occupancy information in ${\cal M}_p$, with its vertices indicated by white circles, and the black filled circle is the centroid point. In addition, the black cross represents the extracted edge feature point, and three distinct forms of data association emerge from the edge feature points, marked by ${\mathbf{q}}_{k_1}^e$, ${\mathbf{q}}_{k_2}^e$, and ${\mathbf{q}}_{k_3}^e$, respectively.}
	\label{Fig_9}
\end{figure}

\emph{\textbf{Point-to-Edge}}: For the purpose of point-to-edge matching, a KD-tree is meticulously constructed from the point cloud denoted as $\mathbb{C}_e$, where each point within $\mathbb{C}_e$ corresponds to a specific polygon in ${\cal M}_p$ with counter-clockwise vertices. This KD-tree is subsequently employed to identify the $K$-nearest polygons associated with each edge feature point. In addition, the distance from the closest polygon should be less than the given threshold $\epsilon_p$. In scenarios where a point is determined to be enclosed by the $K$-nearest polygons, as shown by ${\mathbf{q}}_{k_3}^e$ in Fig. \ref{Fig_9}, it is deliberately excluded from the subsequent phase of formulating the cost function.

On the other hand, two different constraint forms are constructed for edge feature points, as shown by ${\mathbf{q}}_{k_1}^e$ and ${\mathbf{q}}_{k_2}^e$ in Fig. \ref{Fig_9}. Firstly, for the edge feature point ${\mathbf{q}}_{k_2}$, find the polygon that is closest to it, as shown by the top polygon in Fig. \ref{Fig_9}. Subsequently, the next step involves identifying the edge on the polygon that is in closest proximity to the edge feature point. Finally, the identification of various constraint forms depends on whether the projection of the edge feature point onto the nearest edge falls precisely on the edge. Therefore, same as the form of point-to-vertex, the matching relationship between ${\mathbf{q}}_{k_2}^e$ and ${\mathbf{p}}_{k \to j}^w$ is established
\begin{align}
	{\mathbf{F}_{e}^1} = \sum\limits_{{k_2} = 1}^{{N_e^1}} {{{\left\| {{\mathbf{p}}_{{k_2} \to j}^w - \tilde{\emph{\textbf{R}}}_t \cdot {\mathbf{q}}_{k_2}^e - \tilde{\emph{\textbf{t}}}_t} \right\|}^2}},
\end{align}where the number of valid matching pairs is denoted by $N_e^1$. In addition, the edge feature point $\mathbf{q}_{{k_1}}^e \in \mathbb{F}_e$ corresponds the edge $\{\mathbf{p}_{{{k_1} \to i}}^w,  \mathbf{p}_{{{k_1} \to j}}^w\} \in {\cal M}_p$, and the matching relationship is established as follows
\begin{align}
	{\mathbf{F}_{e}^2} = \sum\limits_{{k_1} = 1}^{{N_e^2}} {\frac{{\left\| {\left( {\mathbf{q}_{{k_1}}^w - \mathbf{p}_{{{k_1} \to i}}^w} \right) \times \left( {\mathbf{q}_{{k_1}}^w - \mathbf{p}_{{{k_1} \to j}}^w} \right)} \right\|}^2}{{\left\| {\mathbf{p}_{{{k_1} \to i}}^w - \mathbf{p}_{{{k_1} \to j}}^w} \right\|}^2}},
\end{align}where $N_e^2$ is number of valid matching pairs, and $\mathbf{q}_{{k_1}}^w$ is calculated as follows
\begin{align}
	\mathbf{q}_{{k_1}}^w = \tilde{\emph{\textbf{R}}}_t \cdot \mathbf{q}_{{k_1}}^e + \tilde{\emph{\textbf{t}}}_t.
\end{align}In summary, the matching cost function of point-to-edge is represented as follows
\begin{align}
	\mathbf{F}_{e}(\tilde{\mathbf{s}}_t, \mathbb{F}_e, \mathbb{C}_e, {\cal M}_p) = {\mathbf{F}_{e}^1} + {\mathbf{F}_{e}^2}.
\end{align}

On this basis above, in order to obtain the accurate pose estimation, ERPoT performs nonlinear optimization similar to iterative closest point (ICP) methods. This optimization terminates either upon detecting convergence or once the predefined maximum number of iterations has been reached.

\subsubsection{Real-time pose tracking}
In this subsection, the real-time pose tracking process is introduced, anchored by the matching cost function that correlates the current observation with the prior polygon map. Especially, during the initial phase of pose tracking, it is essential to provide an initial pose estimate for the current observation as prior information, which is denoted as $\tilde{\mathbf{s}}_1$. Subsequently, according to equation (18), the optimal estimation about ${\mathbf{s}}_1$ is derived.

Subsequently, for the real-time pose tracking, the continuous pose sequence is recursively obtained according to the continuous observation sequence and the prior polygon map. Taking the time step at $t(>2)$ as an example, the corresponding pose guess $\tilde{\mathbf{s}}_t$ is obtained according to ${\mathbf{s}}_{t-1}$ and ${\mathbf{s}}_{t-2}$ using a constant velocity motion model,
\begin{align}
	\tilde{\emph{\textbf{R}}}_{t} = {\emph{\textbf{R}}}_{t-1} \cdot {\Delta}\tilde{\emph{\textbf{R}}}_{t-1}, \tilde{\emph{\textbf{t}}}_{t} = {\emph{\textbf{R}}}_{t-1} \cdot {\Delta}\tilde{\emph{\textbf{t}}}_{t-1} + {\emph{\textbf{t}}}_{t-1},
\end{align}where ${\Delta}\tilde{\emph{\textbf{R}}}_{t-1}$ and ${\Delta}\tilde{\emph{\textbf{t}}}_{t-1}$ are calculated as
\begin{align}
&{\Delta}\tilde{\emph{\textbf{R}}}_{t-1} = {\emph{\textbf{R}}}_{t-2}^T \cdot {\emph{\textbf{R}}}_{t-1}, \\
&{\Delta}\tilde{\emph{\textbf{t}}}_{t-1} = {\emph{\textbf{R}}}_{t-2}^T \cdot ({\emph{\textbf{t}}}_{t-1} - {\emph{\textbf{t}}}_{t-2}).
\end{align}Therefore, the optimal pose estimation can be obtained according to equation (18) based on current pose guess $\tilde{\mathbf{s}}_t$, current observation, and the prior polygon map. Subsequently, the reliability of the current tracking result is assessed based on the discrepancy between the solved pose information and the predicted pose information. At the same time, the distance and angle thresholds are set to $\lambda_d$ and $\lambda_a$, respectively.

\section{Experiments}
\begin{figure}[!htb]
	\centering
	\includegraphics[width=7cm]{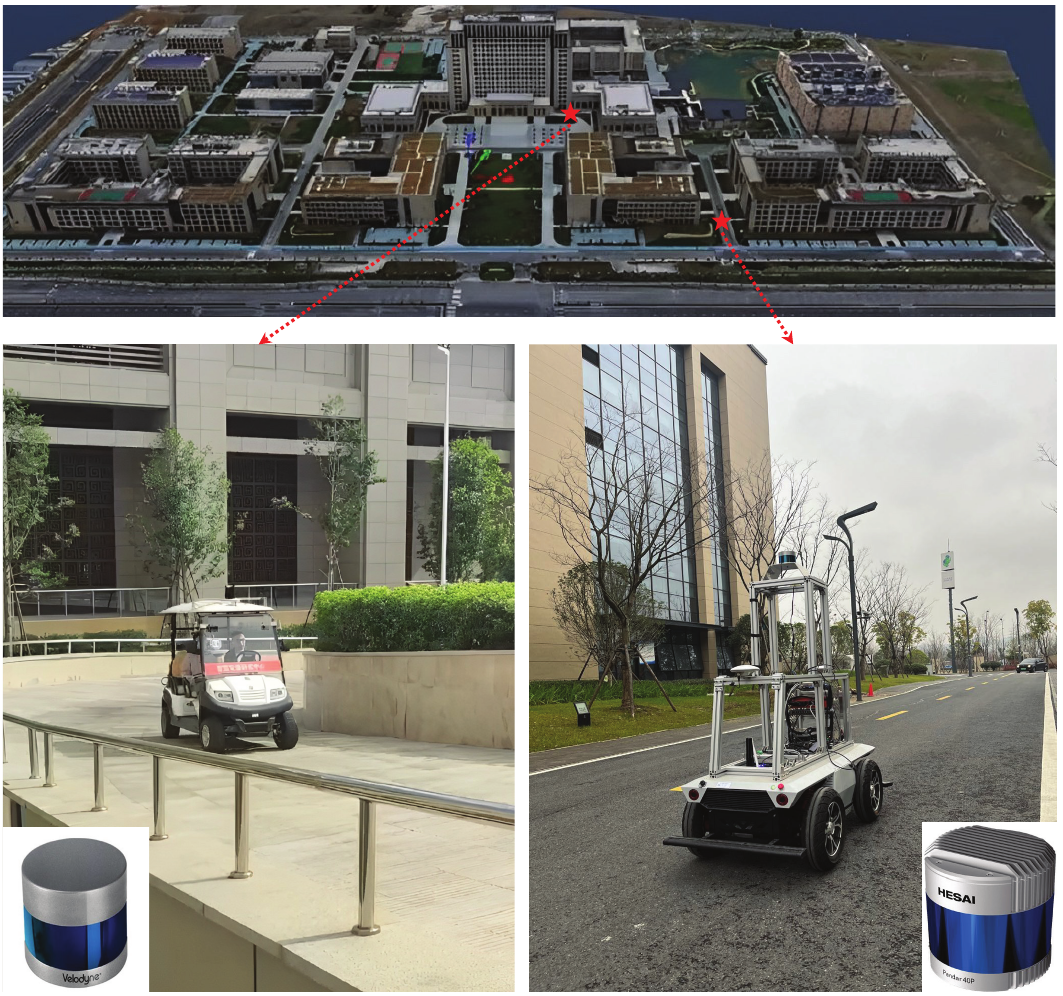}
	\caption{Experimental setup for the self-recorded dataset. In particular, the top figure depicts the experimental setting, which is representative of a typical campus environment. In addition, the bottom two figures show two dataset collection platforms, 204AKZ (Eagle) and FR-09 (YUHESEN), equipped with two different LiDAR sensors, named VLP-32C (Velodyne) and Pandar40P (HESAI), respectively.}
	\label{Fig_10}
\end{figure}

\begin{figure}[!htb]
	\centering
	\subfigure[KITTI dataset]{\includegraphics[width=7cm]{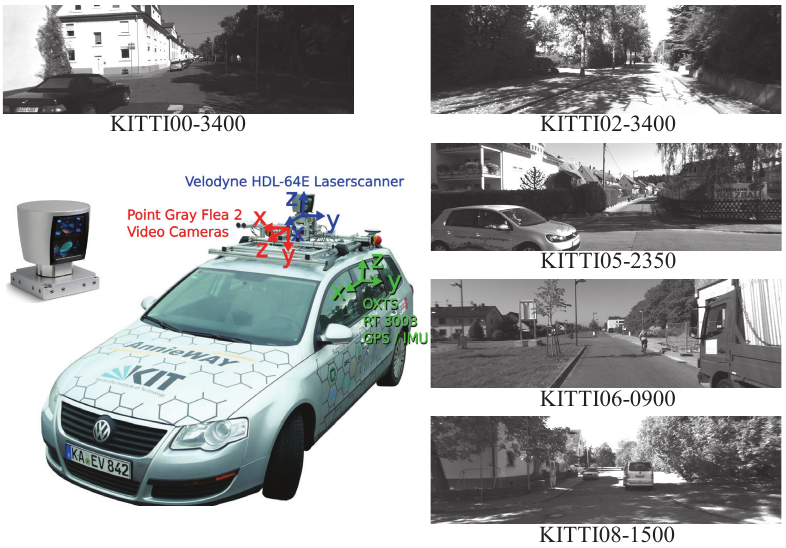}}
	\centering
	\subfigure[Newer College dataset]{\includegraphics[width=7cm]{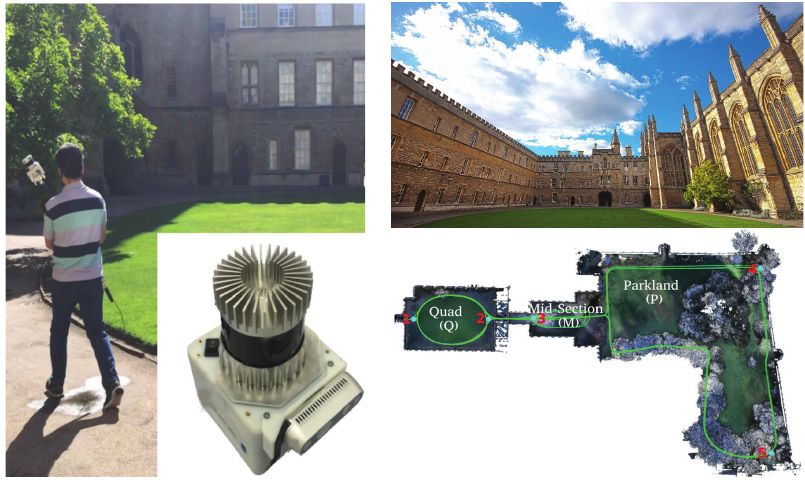}}\\
	\centering
	\subfigure[FusionPortable dataset]{\includegraphics[width=7cm]{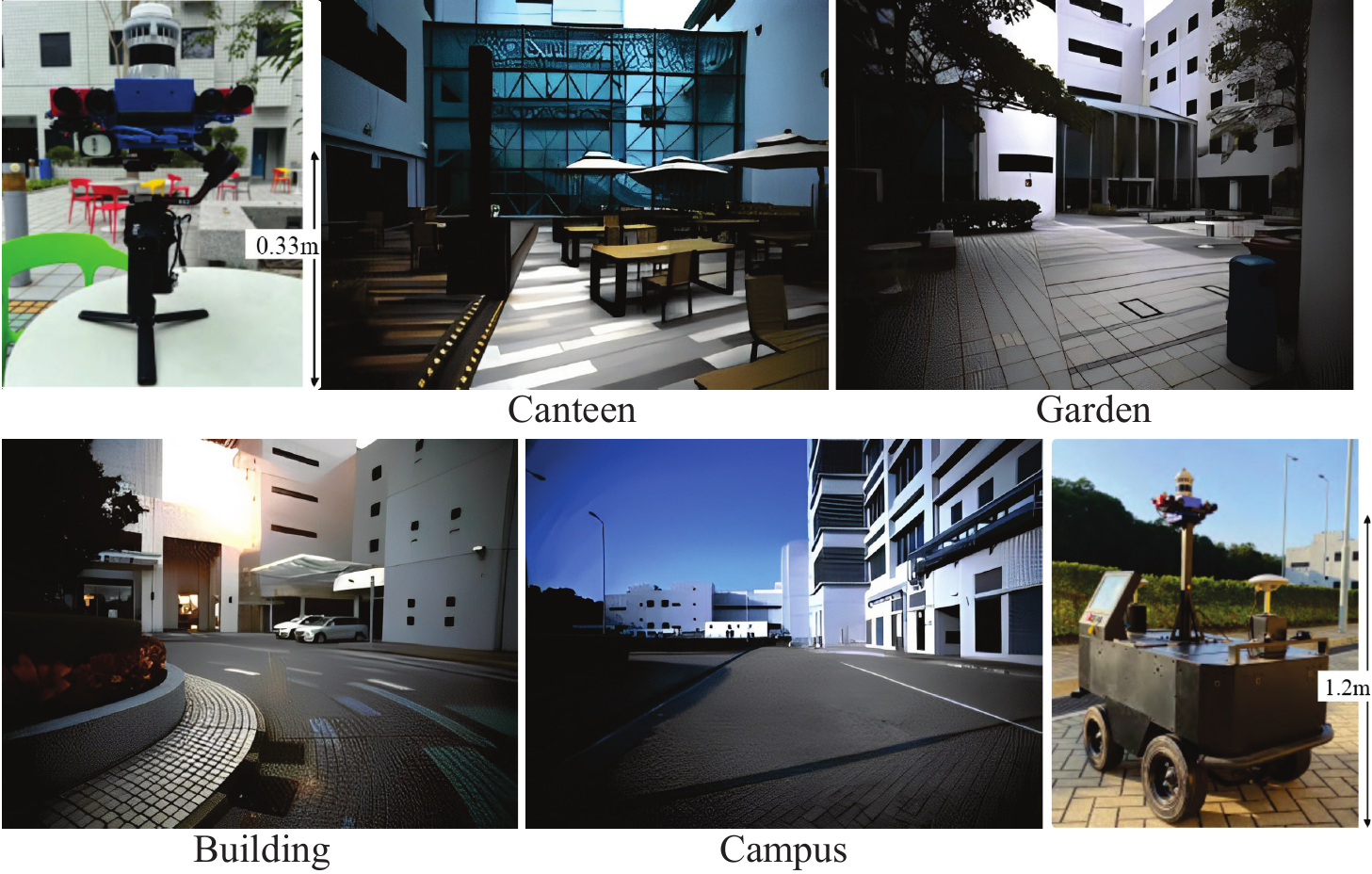}}\\
	\caption{The experimental environments and platforms for dataset collection of two publicly available datasets. For the KITTI dataset, the LiDAR sensor HDL-64E (Velodyne) is used to capture the corresponding point cloud dataset with typical street and road environments. The Newer College dataset is acquired using the handheld LiDAR sensor os1-64 (Ouster) at typical walking speeds through New College, Oxford for nearly $6.7$ km. Additionally, the FusionPortable dataset, captured using Ouster os1-128, is a typical campus-scene dataset that includes both indoor and outdoor environments.}
	\label{Fig_11}
\end{figure}

To show the superiority and effectiveness of the proposed ERPoT, both self-recorded datasets and publicly available datasets are used for the evaluation in this section. As shown in Fig. \ref{Fig_10}, the experimental environment exemplifies a typical campus setting, characterized by an abundance of architectural structures and lush vegetation. Moreover, publicly available datasets used in this evaluation include the KITTI dataset \cite{Geiger}, the Newer College dataset \cite{Ramezani}, and the FusionPortable dataset \cite{JJiao}, while corresponding environments and platforms are shown in Fig. \ref{Fig_11}.

\begin{table}[!htb]\small
	\renewcommand\arraystretch{0.8}
	\centering
	\caption{Hardware and software specification}
	{\fontsize{7pt}{7pt}
	\begin{tabular}{c|c}
		\hline \hline
		\textbf{Robot}           & \textbf{204AKZ and FR-09} \\
		\hline
		\textbf{Sensor}          & \textbf{VLP-32C and Pandar40P} \\
		\hline
		\textbf{Processor}       & \textbf{Intel Core i7-11800H @2.3Ghz} \\
		\hline
		\textbf{RAM}             & \textbf{32GB}  \\
		\hline
		\textbf{Operating system} & \textbf{Ubuntu 20.04}          \\
		\hline
		\textbf{ROS}             & \textbf{Noetic}        \\
		\hline \hline
	\end{tabular}}
	\label{table_1}
\end{table}

\begin{table}\small
	\renewcommand\arraystretch{0.8}
	\centering
	\caption{Systems parameters}
	{\fontsize{7pt}{7pt}
		\begin{tabular}{l|c}
					\hline \hline
					\textbf{Parameter:          }                              & \textbf{value}\\
					\hline
					\textbf{Grid map resolution ($r$)}                                  &   \\
					- \emph{Small scenes}                                      & $0.05$m\\
					- \emph{Large scenes}                                      & $0.10$m\\
					\hline
					\textbf{Pose tracking}                                & \\
					- \emph{Vertex data association threshold} ($\epsilon_v$)  & $2.0$m\\
					- \emph{Polygon data association threshold} ($\epsilon_p$) & $3.0$m\\
					- \emph{Distance difference threshold} ($\lambda_d$)       & $1.0$m\\
					- \emph{Angle difference threshold} ($\lambda_a$)          & $0.2$rad\\
					\hline
					\textbf{Feature extraction}                            & \\
					- \emph{Corner smoothness threshold} ($\sigma_c$)    & $10.0$\\
					- \emph{Edge smoothness threshold} ($\sigma_e$)      & $0.1$\\
					- \emph{Corner feature number threshold} ($\tau_c$)  & $10$\\
					- \emph{Edge feature number threshold} ($\tau_e$)    & $50$\\
					\hline \hline
		\end{tabular}}
	\label{table_2}
\end{table}

Furthermore, the Robot Operating System (ROS) is utilized for both hardware and software implementations. The code for constructing the polygon map and performing pose tracking is written in C/C++. Experiments are conducted on a computer with specifications detailed in Tab. \ref{table_1} and the system parameters of ERPoT are shown in Tab. \ref{table_2}. The approach has been benchmarked against six other methods. It is important to note that during the pose tracking process, the estimation of the current pose relies solely on the prior map (including the point cloud map, mesh map, distance field map, and the proposed polygon map) and data obtained from the LiDAR sensor.

\noindent {\ttfamily ALOAM\_MCL}: A-LOAM\footnote{https://github.com/HKUST-Aerial-Robotics/A-LOAM} is an advanced implementation of LOAM \cite{JZhang}, which makes use of Eigen and Ceres Solver to simplify code structure, providing the real-time odometry information in pure LiDAR mode. In addition, combined with the point cloud-based Monte Carlo localization\footnote{https://github.com/at-wat/mcl\_3dl} (MCL), we can obtain a probabilistic 6-DOF localization system {ALOAM\_MCL} for mobile robots with only LiDAR sensor.

\noindent {\ttfamily FLOAM\_MCL}: F-LOAM\footnote{https://github.com/wh200720041/floam} \cite{HWang} represents a streamlined and cost-efficient evolution of A-LOAM and LOAM, offering improved computational efficiency while maintaining robust LiDAR odometry for mobile robots. In addition, based on the point cloud-based MCL and the prior map, the process of global pose tracking {FLOAM\_MCL} can be carried out.

\noindent {\ttfamily KISS\_MCL}: KISS-ICP\footnote{https://github.com/PRBonn/kiss-icp} \cite{Vizzo} is a LiDAR odometry pipeline that just works on most of the cases without tuning any parameter. This system stands out as a remarkably effective solution, noted for its simplicity in implementation and versatility in operation across diverse environmental settings with different LiDAR sensors. Afterward, the global pose tracking approach called {KISS\_MCL} is obtained.

\noindent {\ttfamily HDL\_localization}: {HDL\_localization}\footnote{https://github.com/koide3/hdl\_localization} \cite{Koide1} is a ROS package for real-time 3D localization with different LiDAR sensors, such as Velodyne HDL-32E and VLP-16. Based on the voxelized GICP (VGICP) algorithm\footnote{https://github.com/koide3/fast\_gicp} \cite{Koide2}, this package performs Unscented Kalman Filter-based pose estimation.

\noindent {\ttfamily SM\_MCL}: The work in \cite{Akai} provides a ROS package\footnote{https://github.com/NaokiAkai/mcl3d\_ros} of 3D LiDAR-based pose tracking for mobile robots. It acknowledges the complementary strengths and weaknesses of Monte Carlo Localization (MCL) and Scan Matching (SM), integrating these two sophisticated localization methods. The fusion of MCL and SM within {SM\_MCL} is innovative, allowing the package to operate effectively under pure LiDAR mode.

\noindent {\ttfamily Range\_MCL}: The range image provides a lightweight 3D LiDAR scan representation, and the mesh map is more compact than the point cloud map. In \cite{XChen}, a novel sensor model combined with MCL is used for 3D LiDAR-based global localization and pose tracking \footnote{https://github.com/PRBonn/range-mcl}. This model compares current LiDAR range images to synthetic ones from the mesh to adjust particle weights, offering a simple yet versatile approach.

\subsection{Experimental configurations}

In the pursuit of pose tracking evaluation, we meticulously select segments from each dataset to construct prior maps, which act as essential baselines for the localization process. Specifically, we make use of the provided segment sequence including ground truth values and corresponding point clouds to transform all point clouds into the world coordinate system, thereby creating a prior point cloud map. Additionally, to manage the large scale of the point cloud map, we apply voxel filtering to facilitate efficient online pose tracking, the final prior map is used for ALOAM\_MCL, FLOAM\_MCL, KISS\_MCL, and HDL\_localization. Based on the obtained point cloud map, the distance field representation for SM\_MCL is generated using the provided open-source tool. Similarly, the prior mesh map for Range\_MCL is acquired using the open-source code.

\begin{figure}[!htb]
	\centering
	\subfigure[KITTI00]{\includegraphics[width=4.3cm]{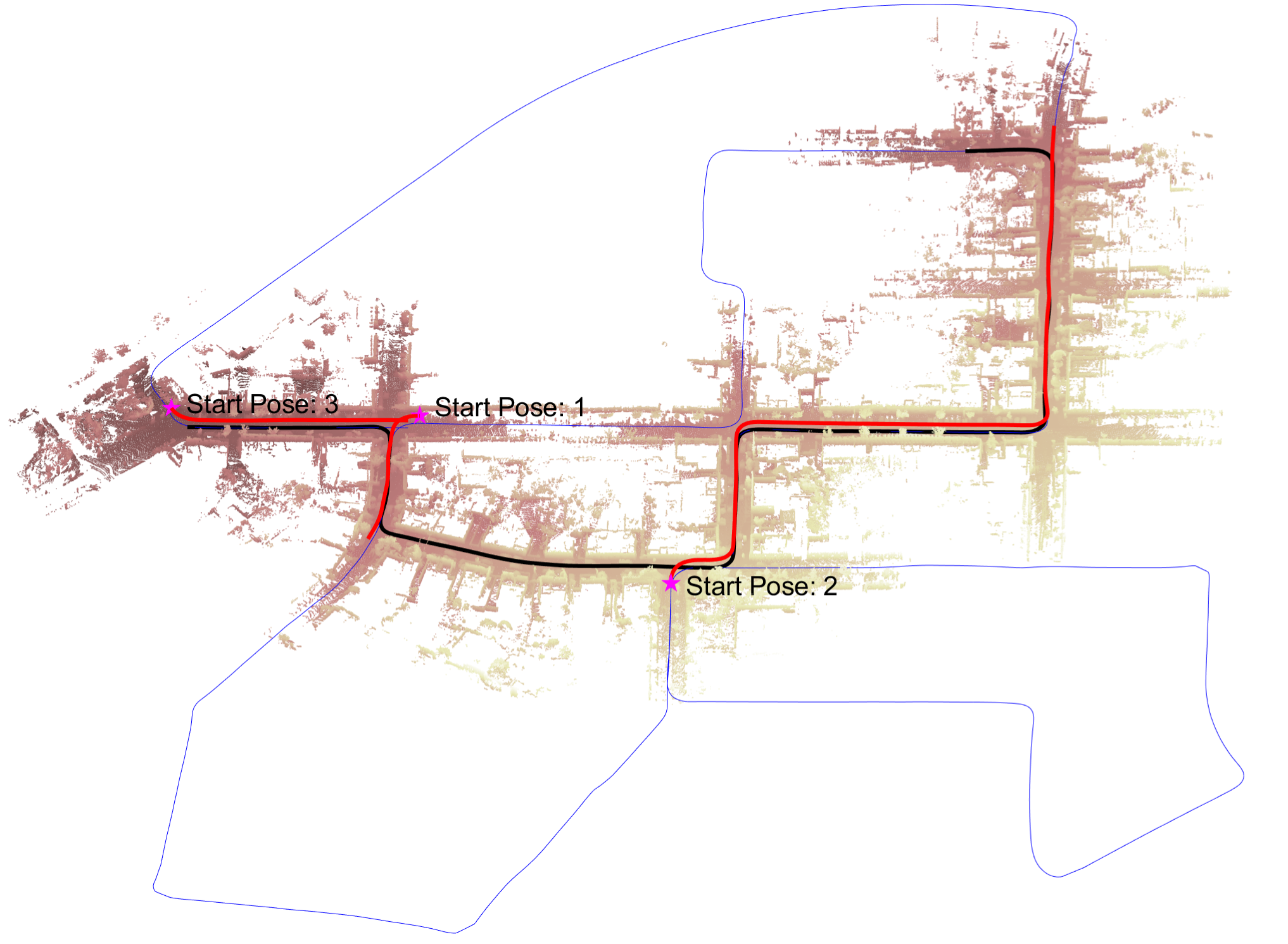}}
	\subfigure[KITTI02]{\includegraphics[width=4.3cm]{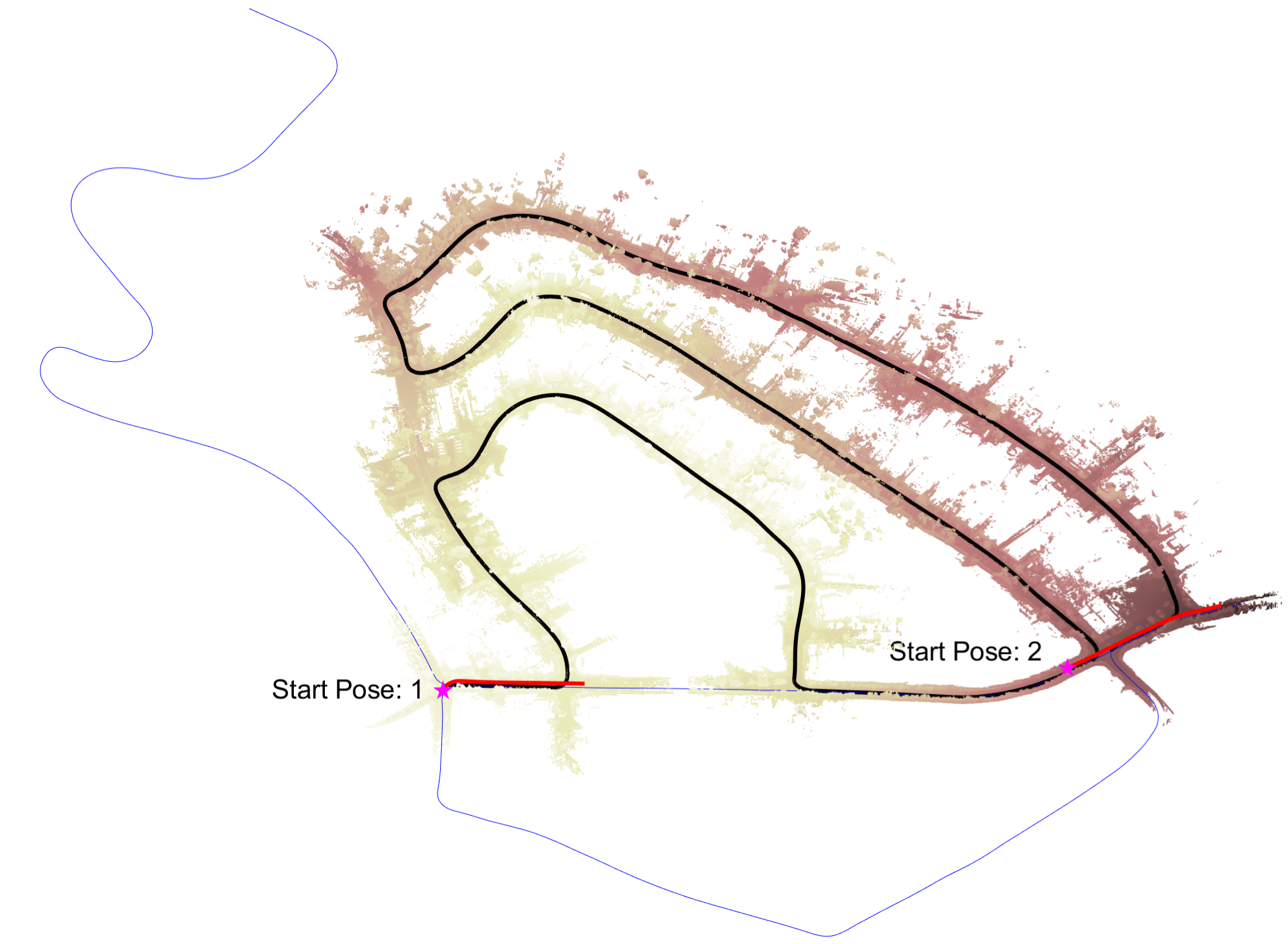}}\\
	\subfigure[KITTI05]{\includegraphics[width=4.3cm]{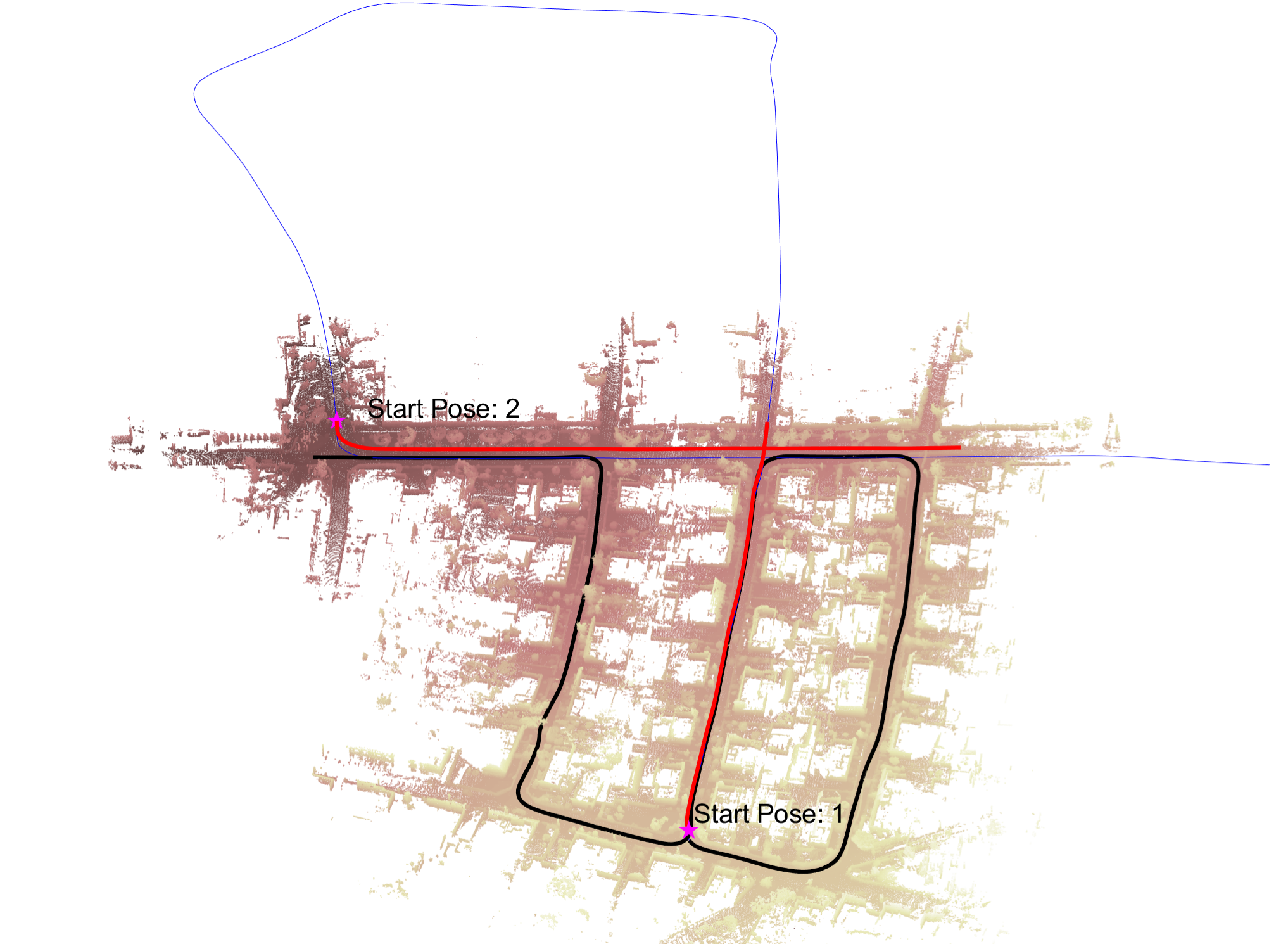}}
	\subfigure[KITTI08]{\includegraphics[width=4.3cm]{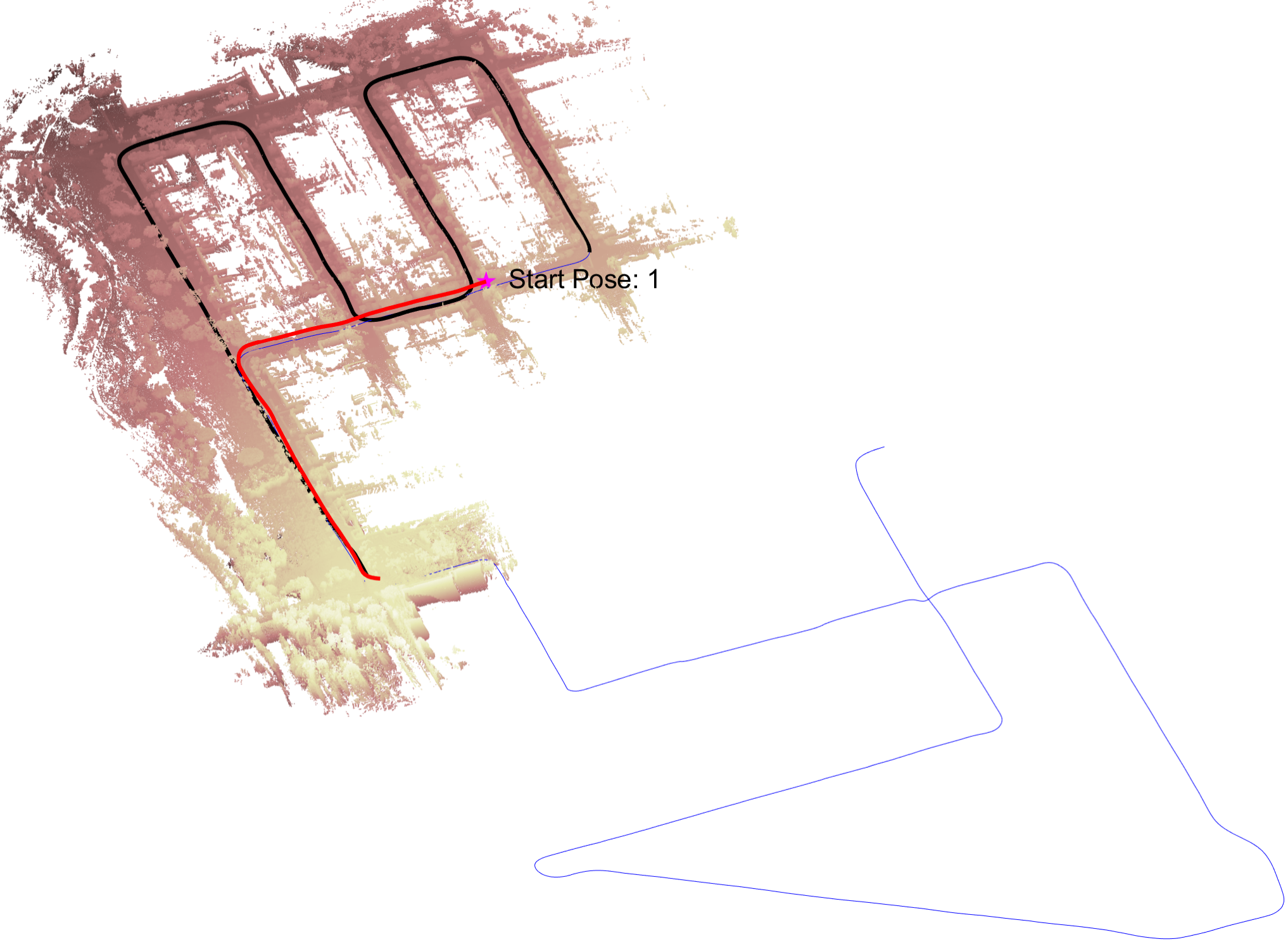}}\\
	\subfigure[KITTI06]{\includegraphics[width=8.0cm]{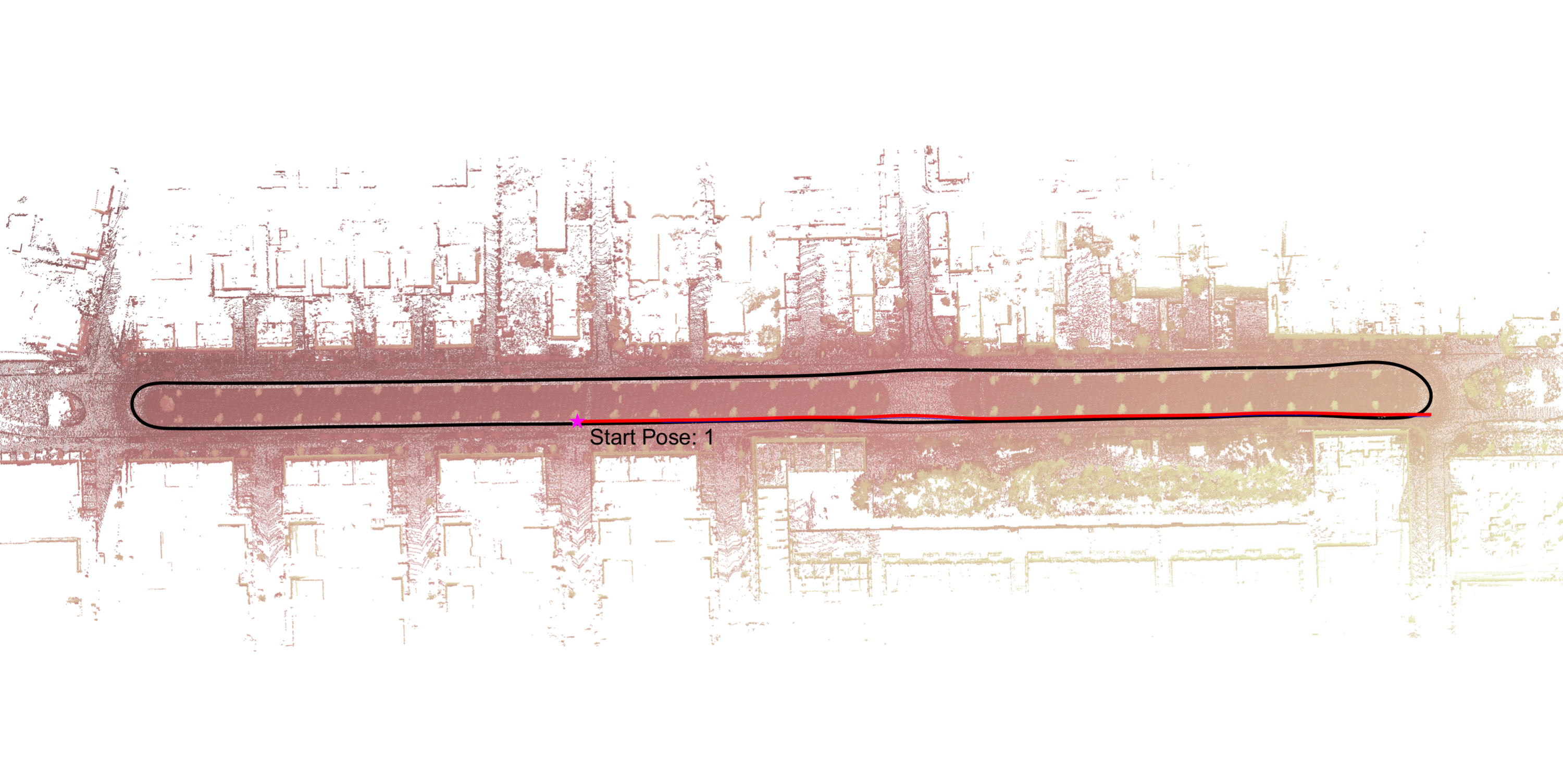}}
	\caption{The experimental configurations of the KITTI dataset. The blue thin line represents the complete trajectory of the dataset, and the black thick line represents the trajectory used to construct the prior map. In addition, the trajectory denoted by the red thick line is used for evaluation of pose tracking, and the corresponding start pose is marked by the magenta pentagram.}
	\label{Fig_12}
\end{figure}

For the KITTI dataset, sequences 00, 02, 05, 06, and 08 have been deliberately selected for our evaluative purposes, leveraging their overlapping trajectories to ensure a reliable and comprehensive assessment of pose tracking performance. On the other hand, it is crucial to recognize that although the KITTI dataset remains the famous vision benchmark, the inconsistencies in the baselines are on the order of meters since the ground truth system relies only on RTK-GPS measurements \cite{Ferrari}. Therefore, we combine point cloud registration with the ground truth provided by the KITTI dataset to further optimize the ground truth of the dataset for subsequent performance evaluation. The detailed experimental configurations are shown in Fig. \ref{Fig_12}, and the corresponding LiDAR sequences are shown in Tab. \ref{table_3}. 

\begin{table}[!htb]\footnotesize
	\renewcommand{\arraystretch}{0.8}
	\centering
	\caption{Experimental configurations of KITTI dataset}
	\setlength{\tabcolsep}{2.2mm}{
		\begin{tabular}{c|c|c}
			\hline \hline
			\multicolumn{1}{c|}{Dataset No.} &\multicolumn{1}{c|}{Map construction} &\multicolumn{1}{c}{Pose tracking} \\
			\hline
			{00} 		&{0000-1000} 	    &{1540-1650, 3360-3860, 4420-4540} \\
			\hline
			{02} 		&{0920-3408} 	    &{4180-4280, 4549-4649} \\
			\hline
			{05} 		&{0000-1300} 	    &{1301-1581, 2300-2650} \\
			\hline
			{06} 		&{0000-0833} 	    &{0834-1100} \\
			\hline
			{08} 		&{0000-1306} 	    &{1400-1850} \\
			\hline \hline
	\end{tabular}}
	\label{table_3}
\end{table}

\begin{figure}[!htb]
	\centering
	\includegraphics[width=7.5cm]{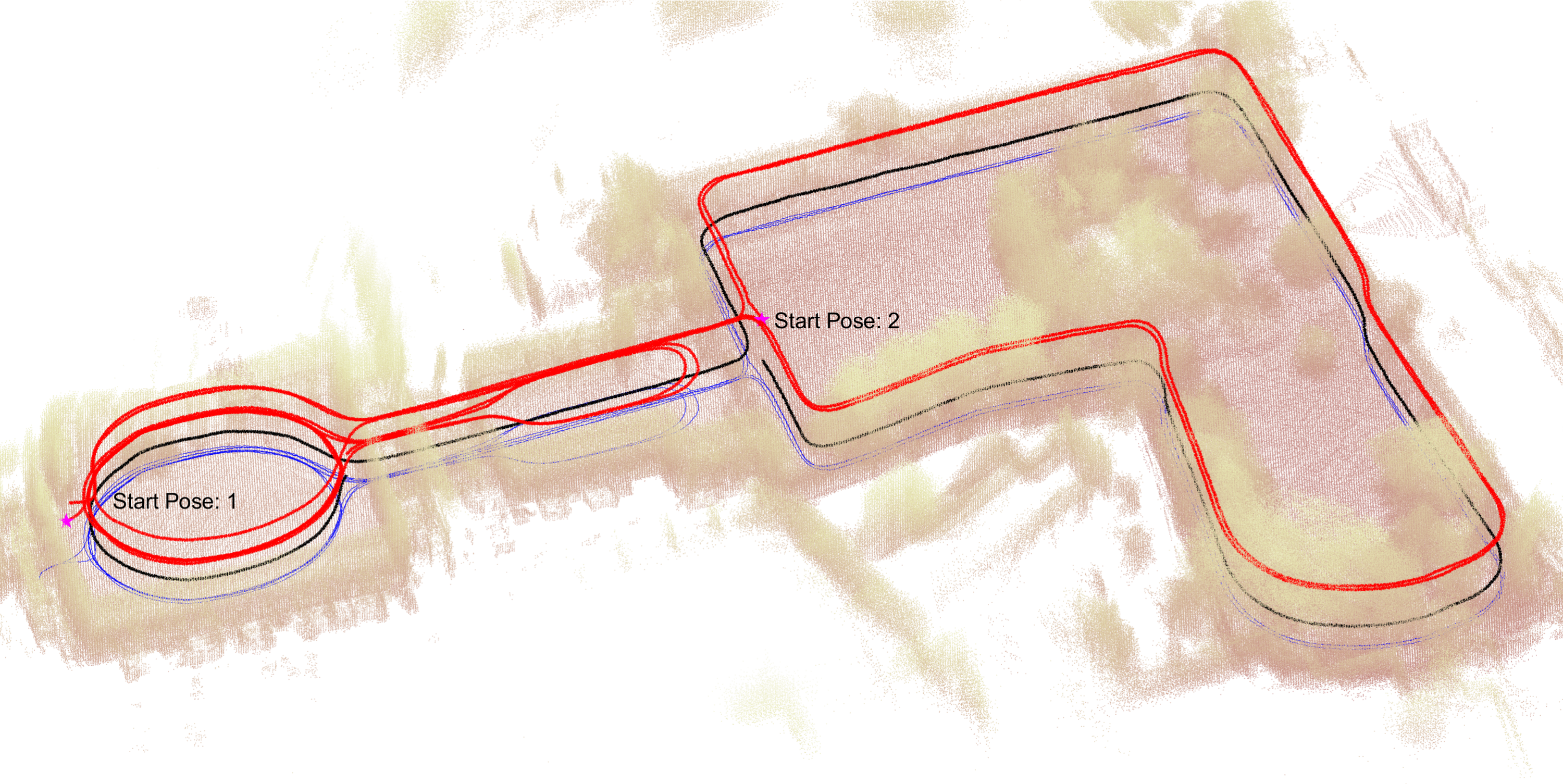}
	\caption{The experimental configurations of the Newer College dataset.}
	\label{Fig_13}
\end{figure}

For the Newer College dataset, a comprehensive collection of $26559$ LiDAR frames is meticulously captured throughout the environment, as depicted in Fig. \ref{Fig_11}(b). The sequence of LiDAR frames within the range of $[8700, 14550]$ is utilized to construct the prior map essential for pose tracking, represented by the black thick trajectory in Fig. \ref{Fig_13}. Furthermore, the sequences $[500, 8699]$ and $[14551, 25901]$ are used for experimental evaluation of pose tracking, denoted by the red thick trajectories in Fig. \ref{Fig_13}.

\begin{figure}[!htb]
	\centering
	\subfigure[Portable-campus]{\includegraphics[width=8.5cm]{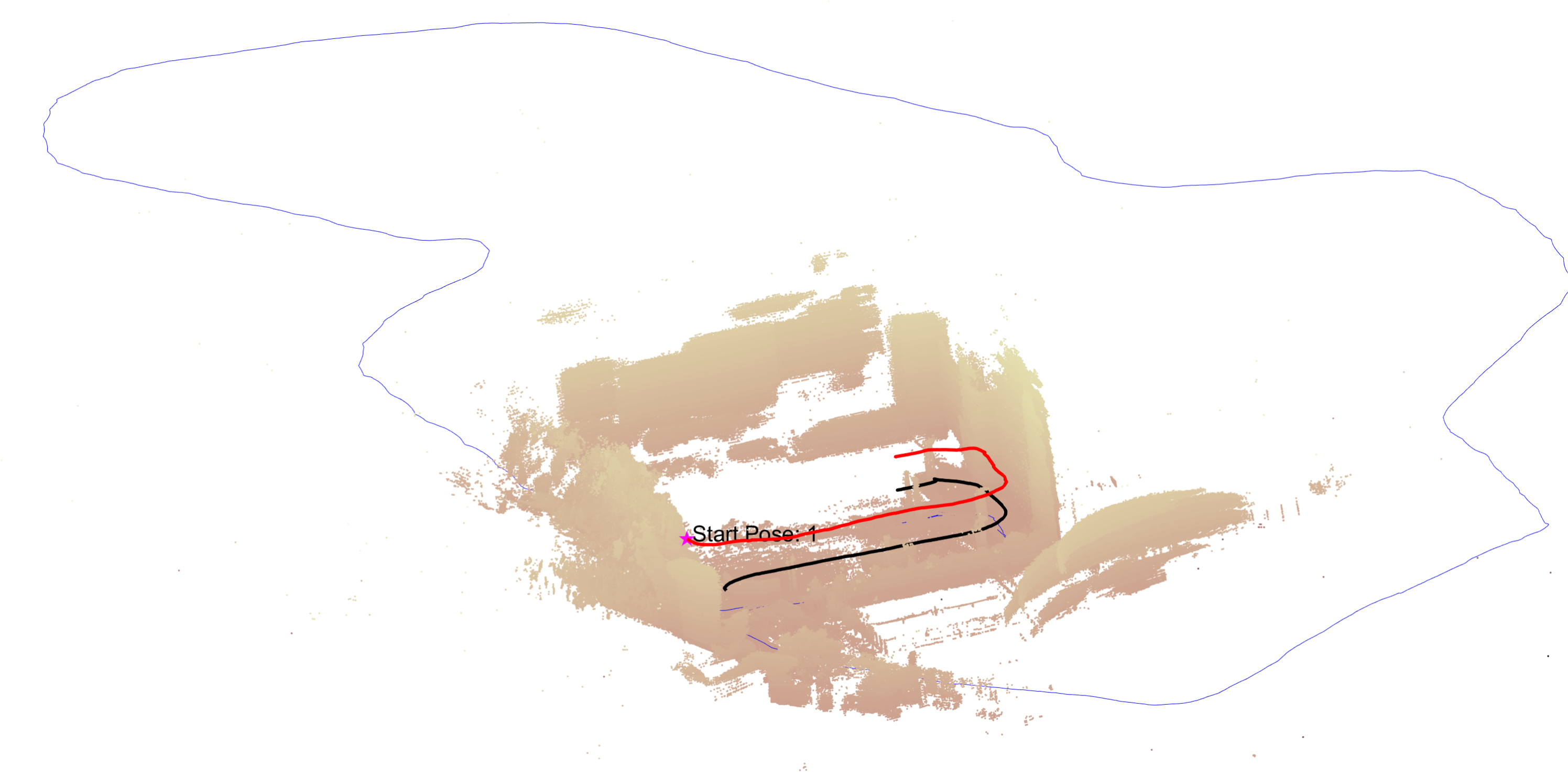}}\\
	\subfigure[Portable-building]{\includegraphics[width=8.5cm]{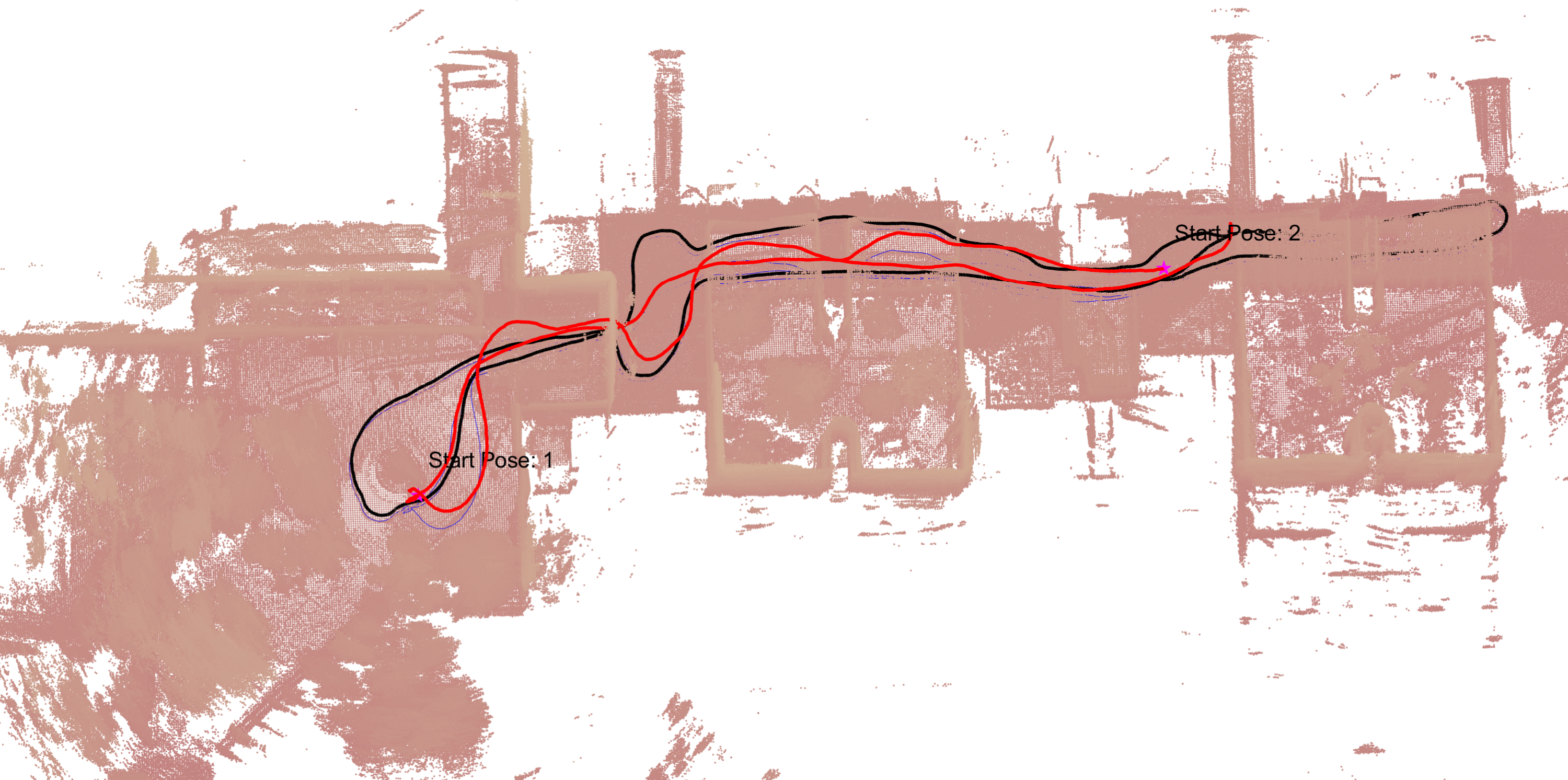}}
	\subfigure[Portable-garden]{\includegraphics[width=4.3cm]{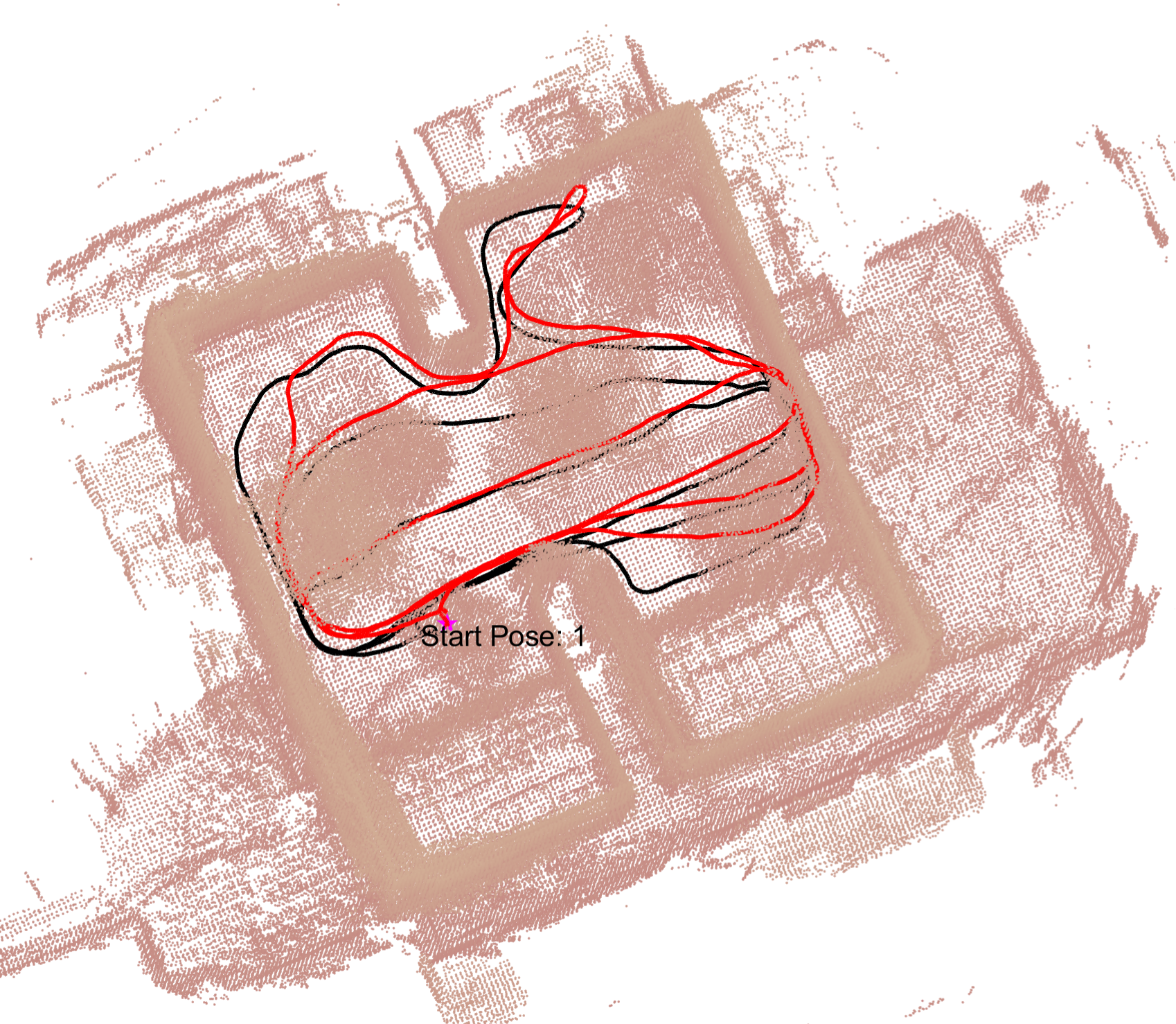}}
	\subfigure[Portable-canteen]{\includegraphics[width=4.3cm]{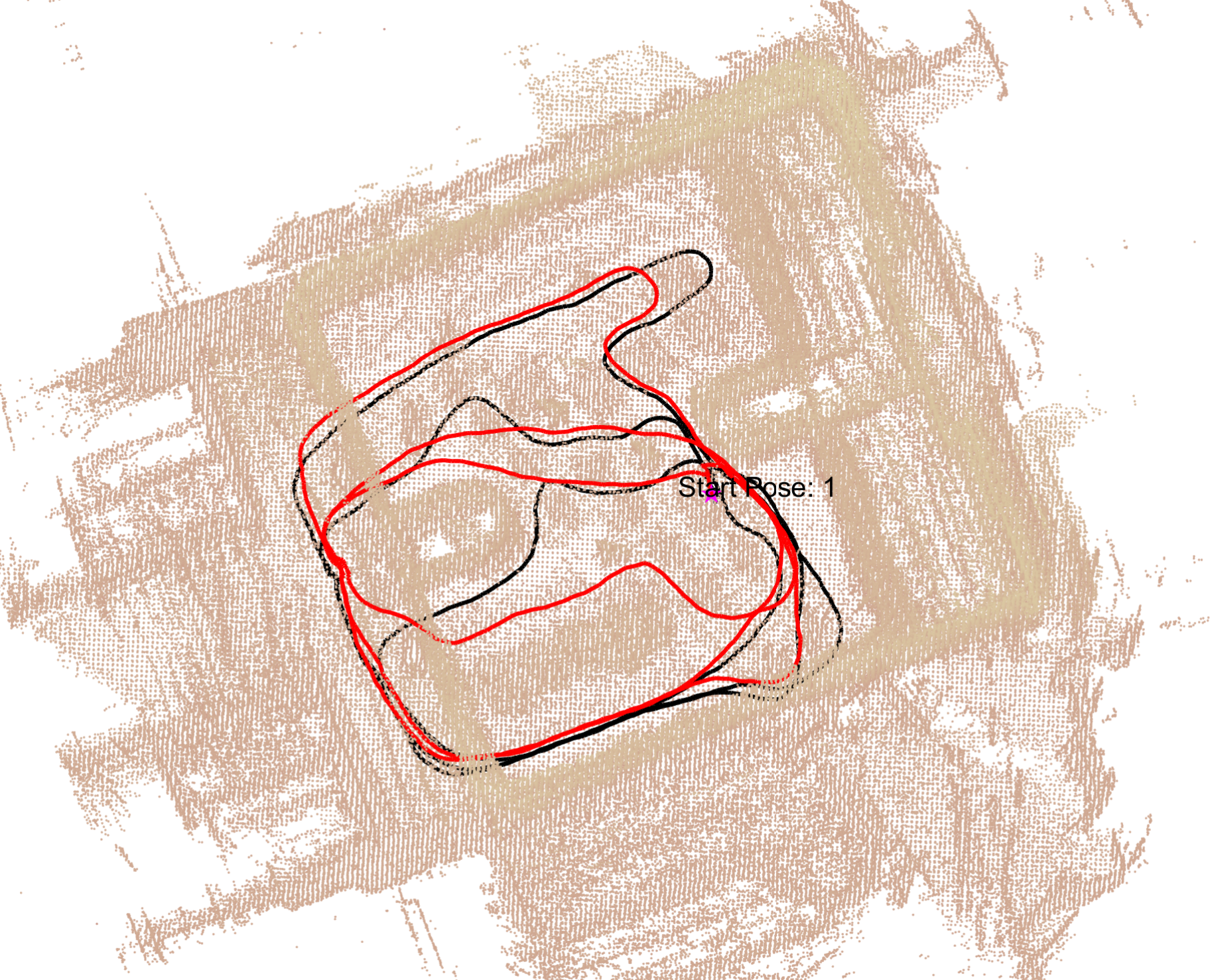}}\\
	\caption{The experimental configurations of the FusionPortable dataset.}
	\label{Fig_14}
\end{figure}

For the FusionPortable dataset, the experimental configurations are illustrated in Fig. \ref{Fig_14}. Specifically, for the Portable-canteen and Portable-garden datasets, which encompass both day and night scenes, the night datasets (20220215\_canteen\_night and 20220215\_garden\_night) are used to construct the prior maps, while the day datasets (20220216\_canteen\_day and 20220216\_garden\_day) are employed for pose tracking evaluation. For Portable-campus (20220226\_campus\_road\_day), the sequence of LiDAR frames within the range $[1,1100]$ is used for prior map construction, and the sequence $[11063,12263]$ is utilized for experimental evaluation. For Portable-building, which includes both indoor and outdoor scenes, the range $[1201,4500]$ is used for prior map construction, and the sequences $[1,1200]$ and $[4501,5996]$ are used for evaluation.

\begin{figure}[!htb]
	\centering
	\subfigure[Self dataset-01]{\includegraphics[width=8cm]{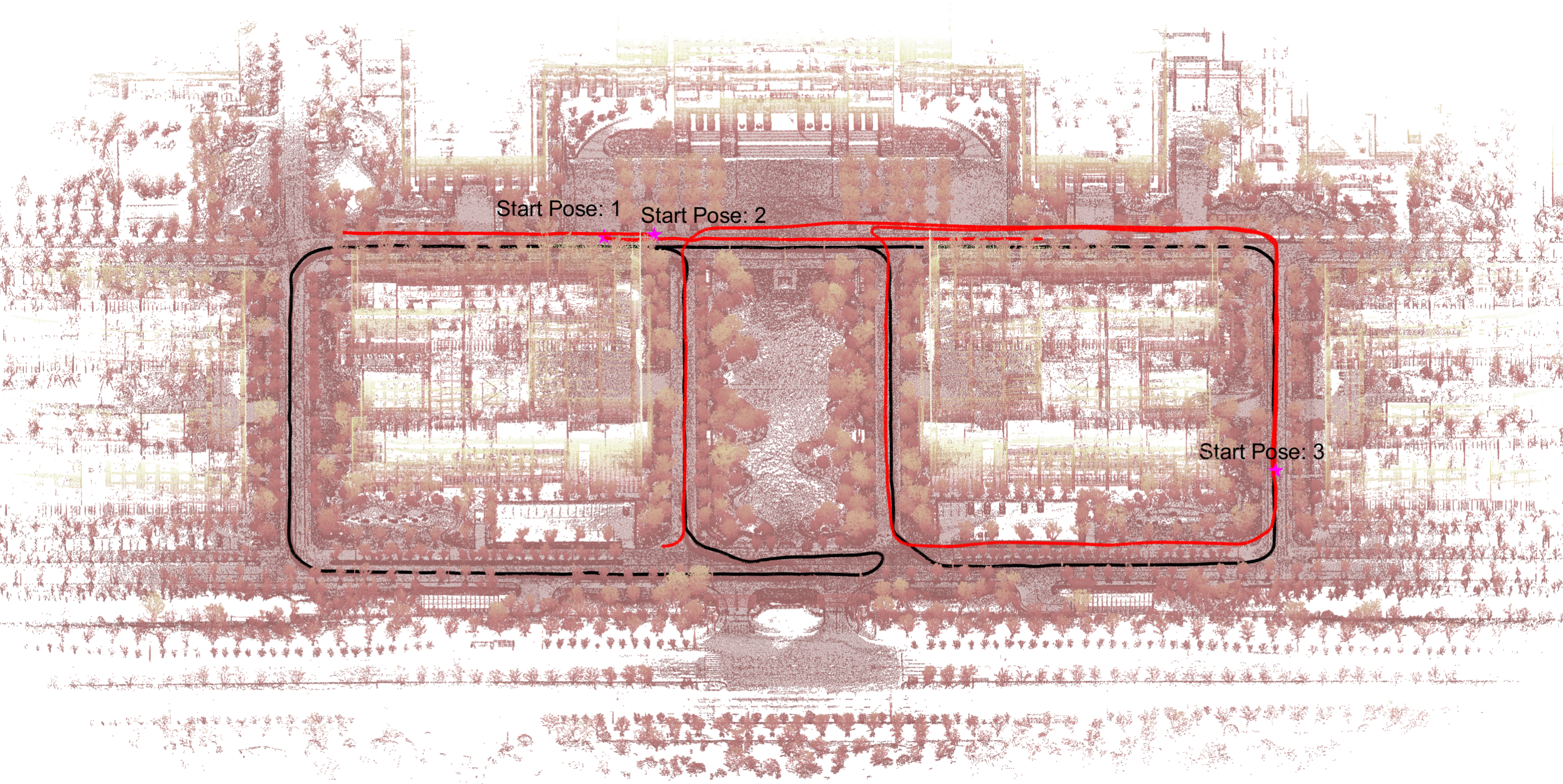}}
	\centering
	\subfigure[Self dataset-02]{\includegraphics[width=8cm]{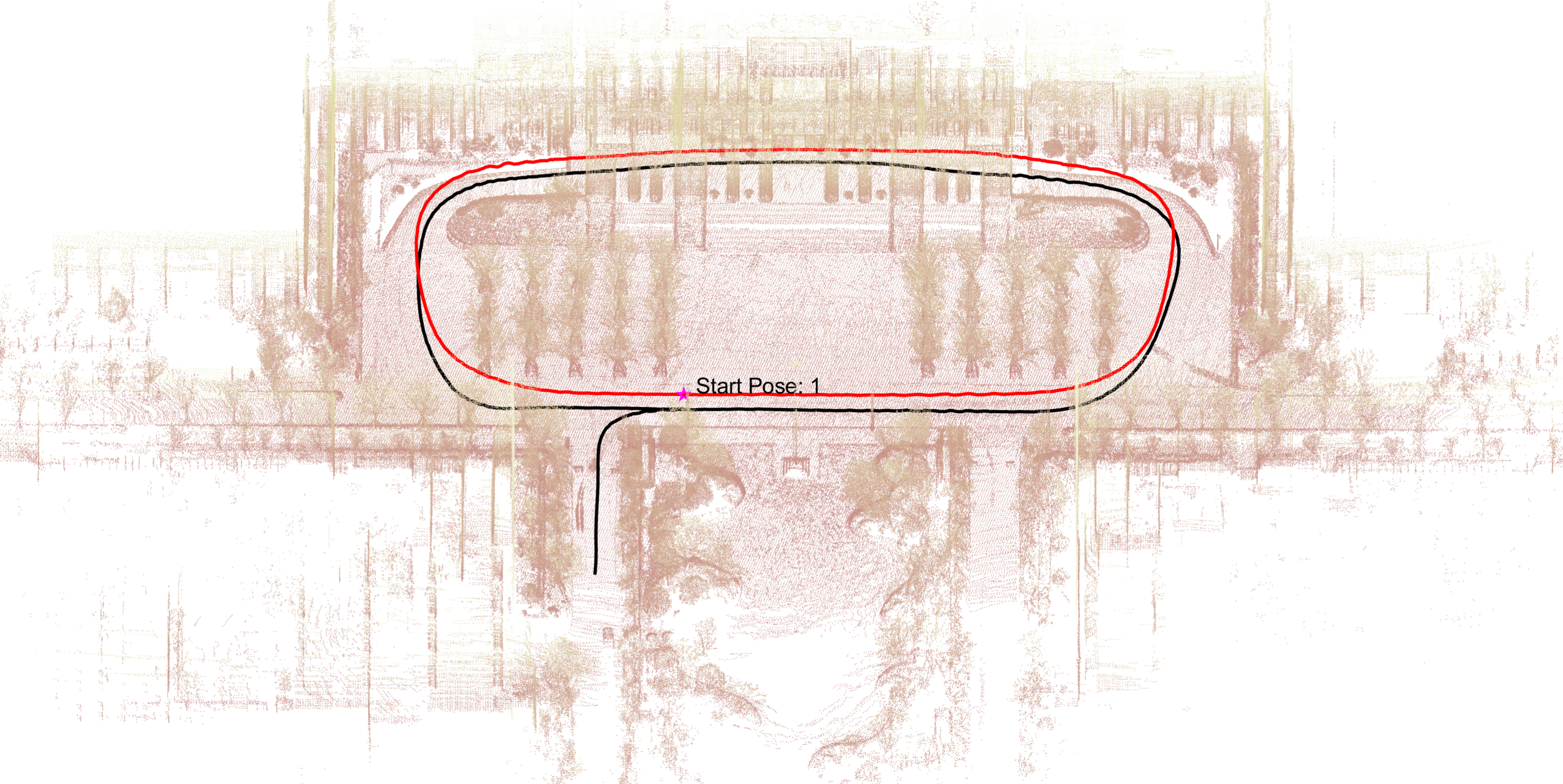}}
	\caption{The experimental configurations of self-recorded dataset.}
	\label{Fig_15}
\end{figure}

To further validate the effectiveness of the proposed approach, we have utilized two different platforms with different LiDAR sensors to collect multiple datasets at various times in the environment shown in Fig. \ref{Fig_10}. Specifically, for Self data-01, two sequences denoted by `Start Pose: 1' and `Start Pose: 2' in Fig. \ref{Fig_15}(a) are recorded at the same time (2021-09-07) as the sequence used to construct the prior map. While the third sequence denoted by `Start Pose: 3' is recorded a year and a half later (2023-03-05) to verify the long-term viability of the proposed polygon map and ERPoT. On the other hand, considering the complexity of the actual environment, especially scenarios involving uphill and downhill sections, and Self dataset-02 has been recorded via the golf cart equipped with VLP-32, as shown in Fig. \ref{Fig_10}.

\subsection{Qualitative experimental results}

\begin{figure*}[!b]
	\centering
	\subfigure[Self dataset-01]{
		\includegraphics[width=4.2cm]{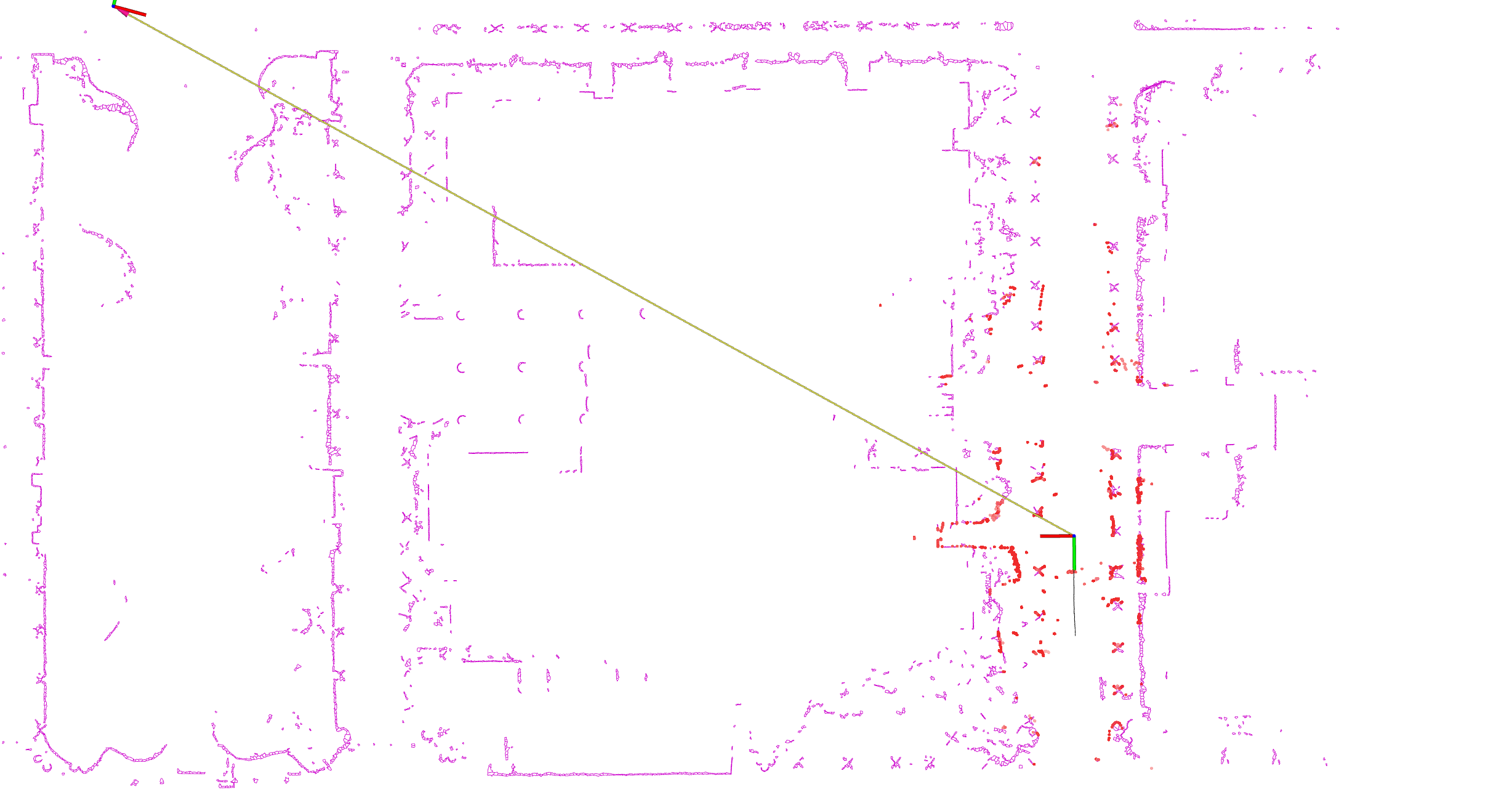}
		\includegraphics[width=4.2cm]{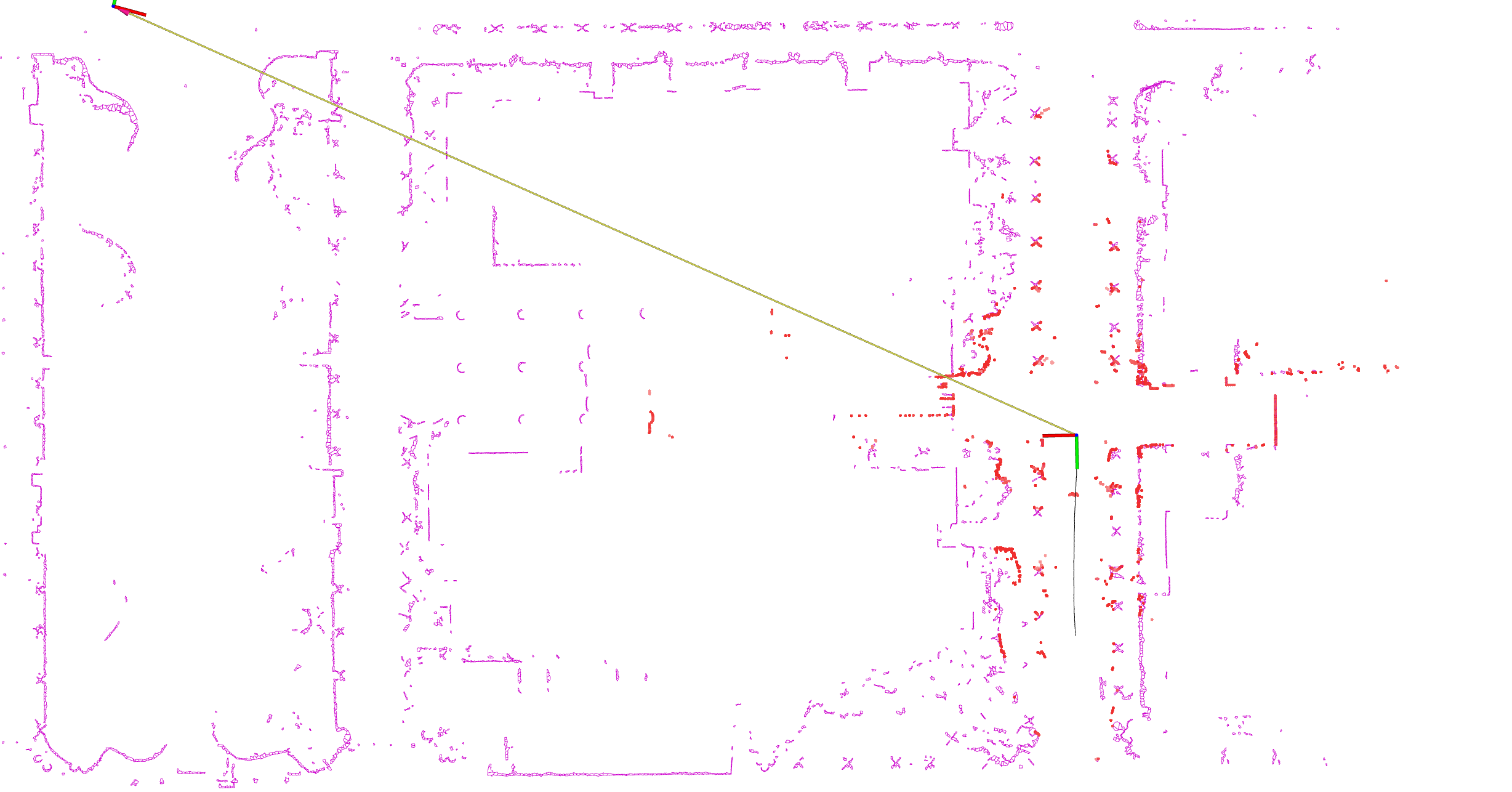}
		\includegraphics[width=4.2cm]{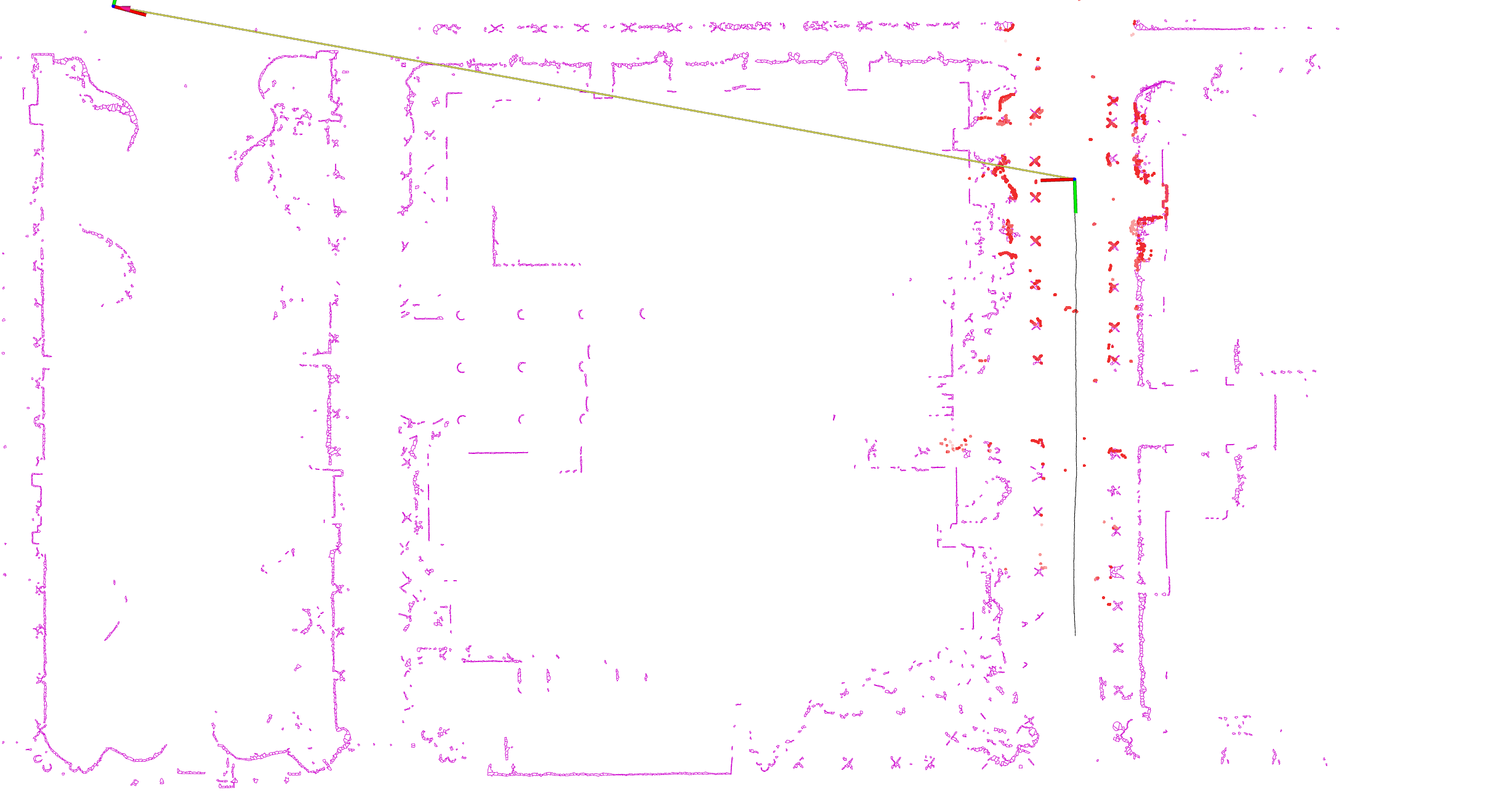}
		\includegraphics[width=4.2cm]{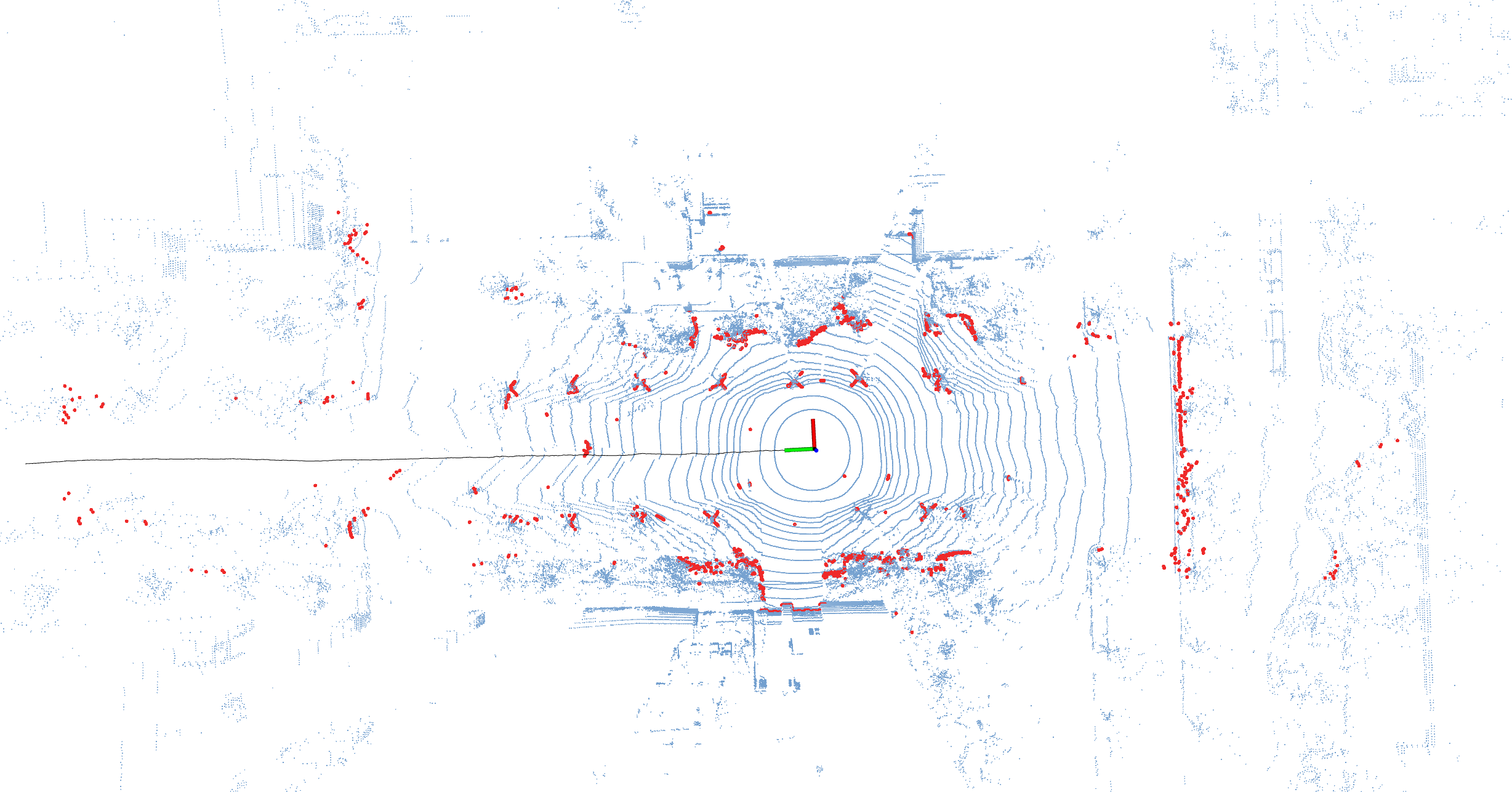}}
	\subfigure[Self dataset-02]{
		\includegraphics[width=4.2cm]{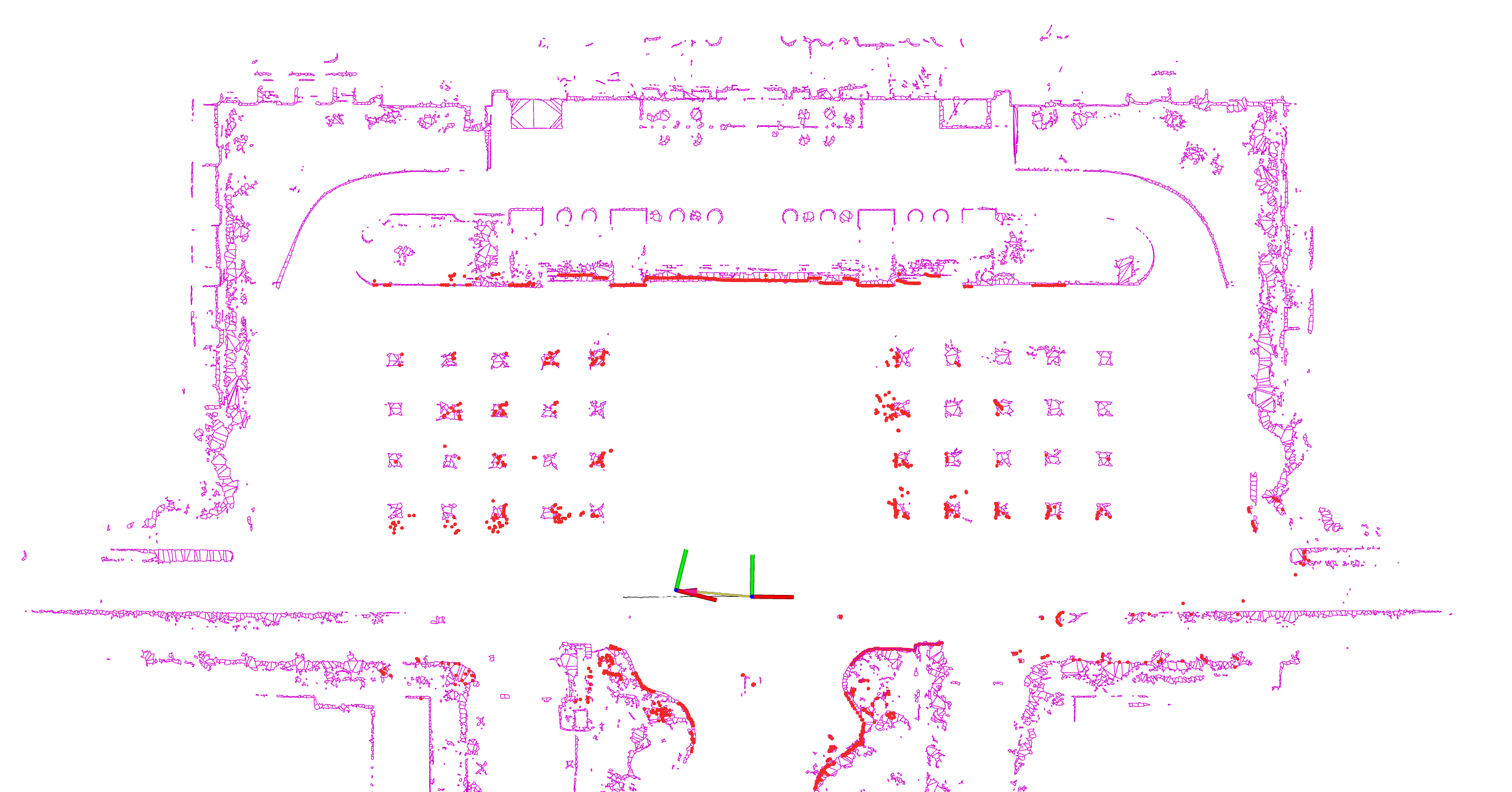}
		\includegraphics[width=4.2cm]{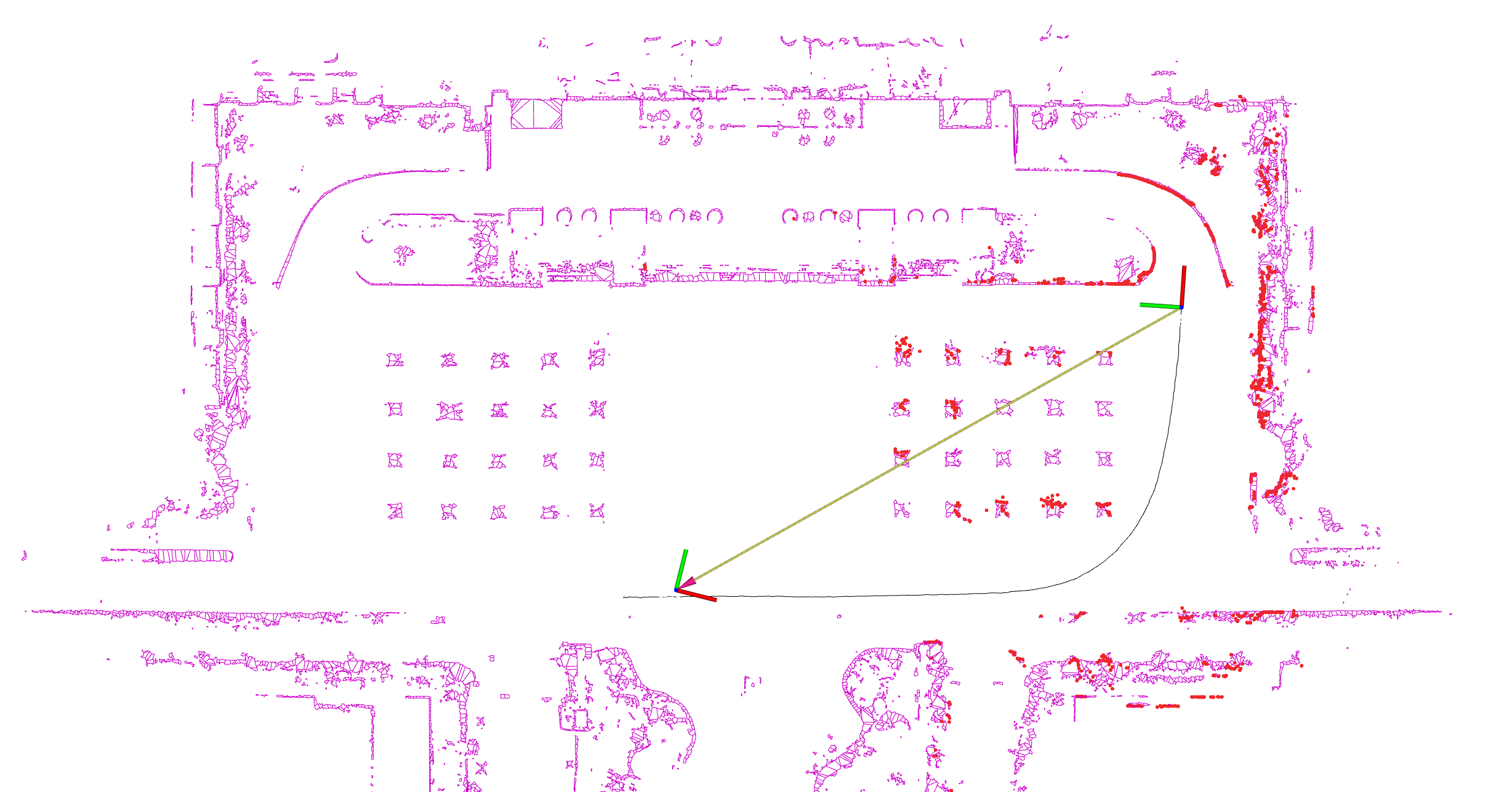}
		\includegraphics[width=4.2cm]{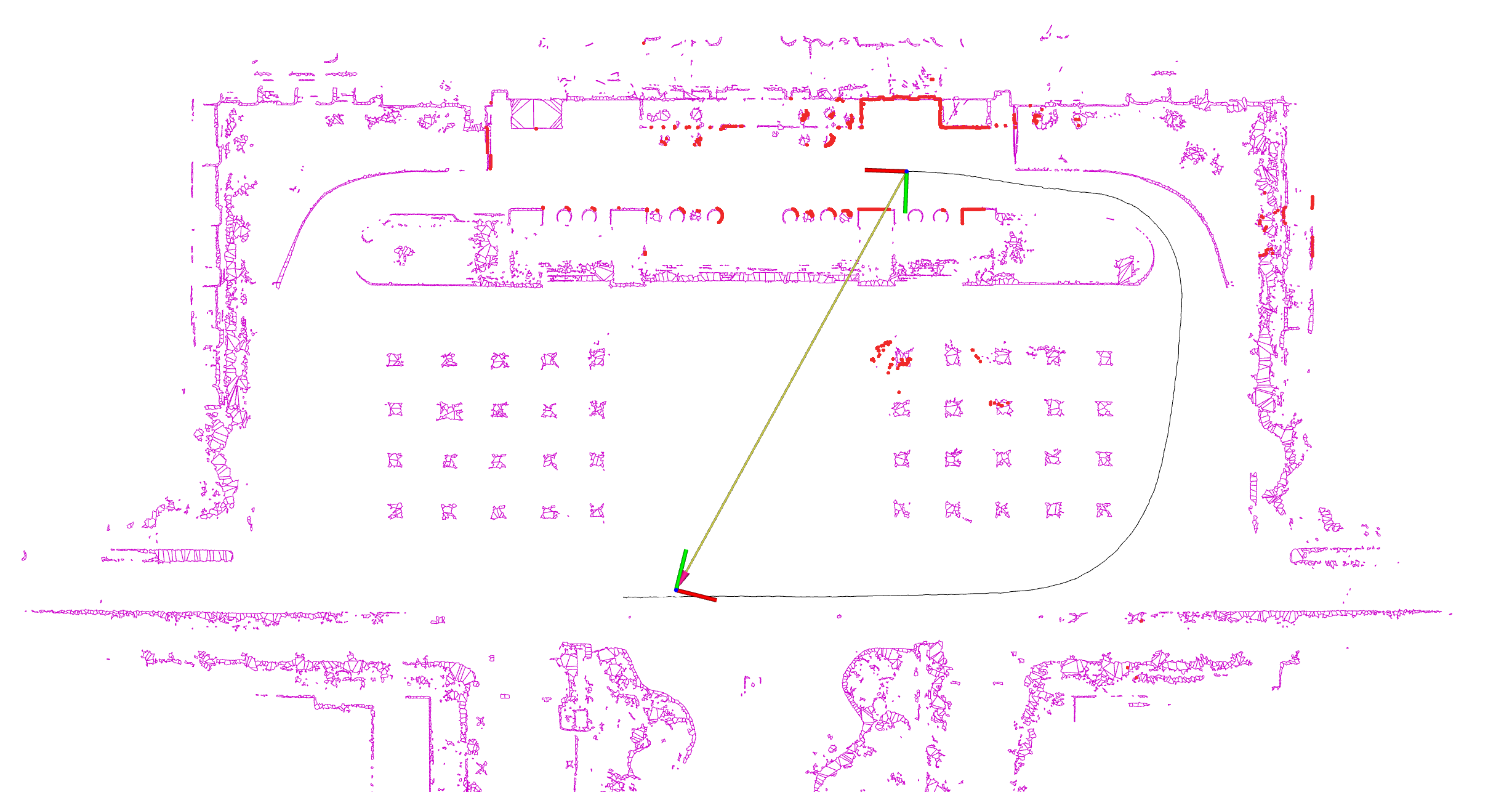}
		\includegraphics[width=4.2cm]{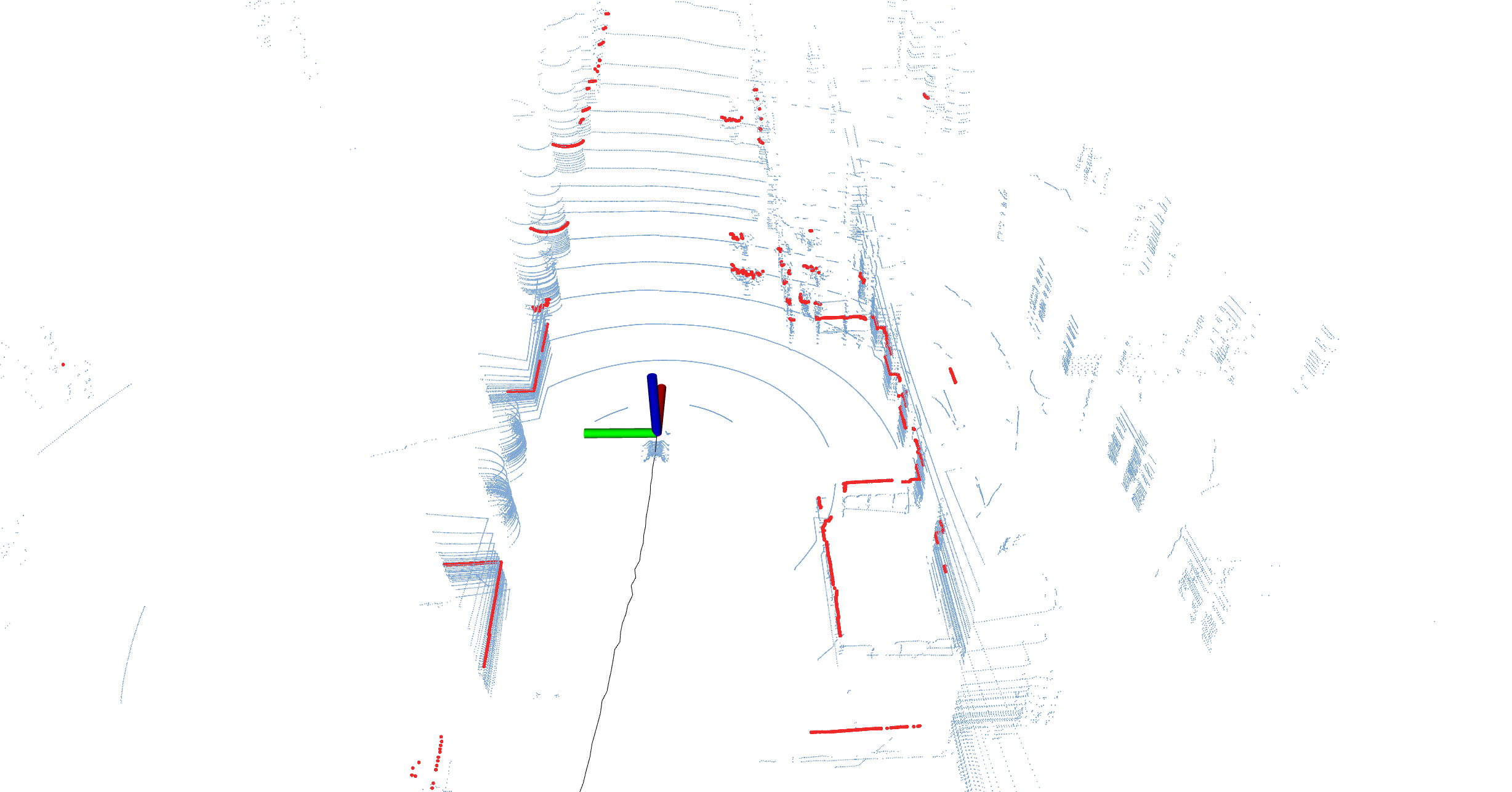}}
	\subfigure[KITTI00]{
		\includegraphics[width=4.2cm]{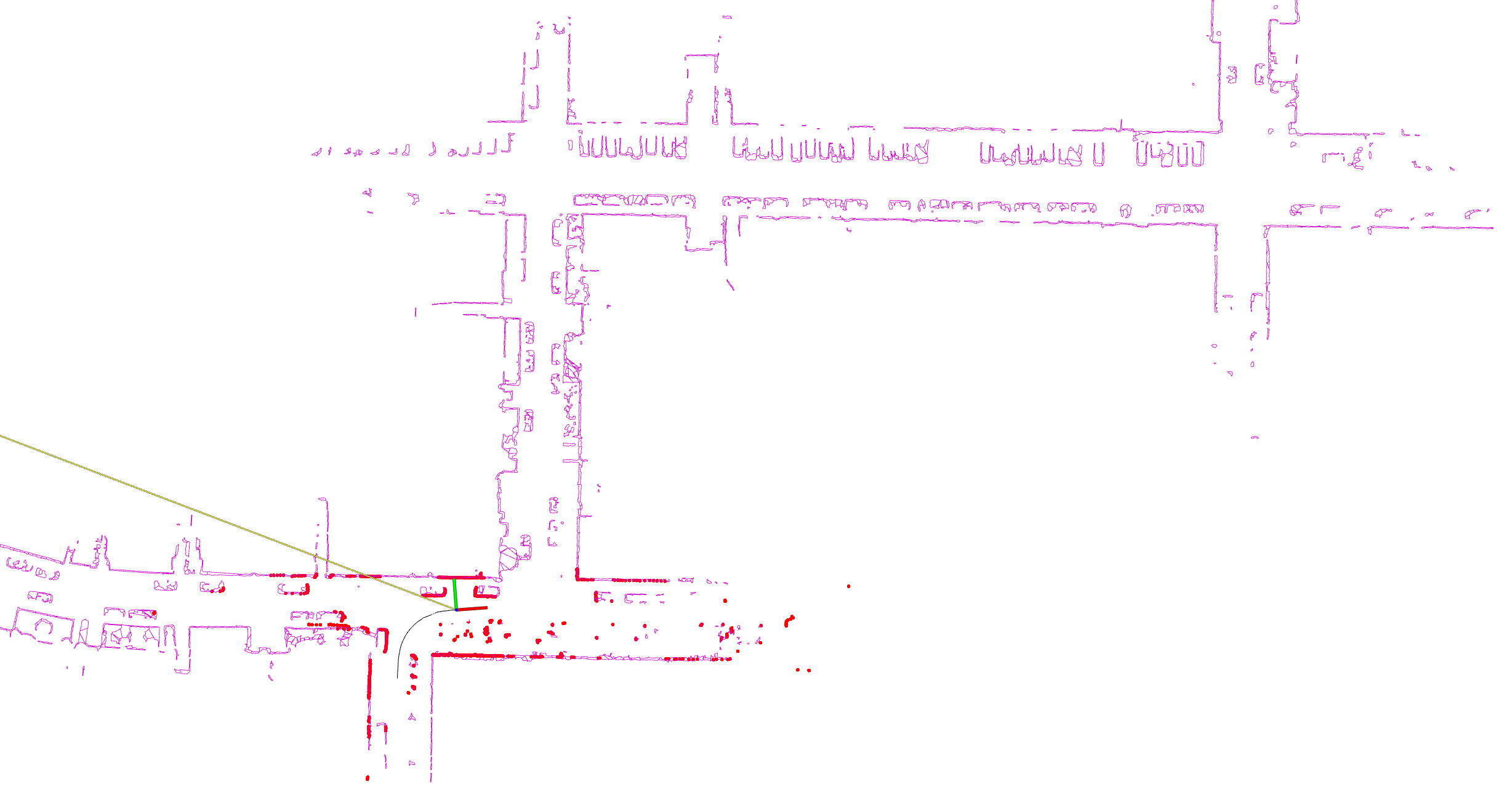}
		\includegraphics[width=4.2cm]{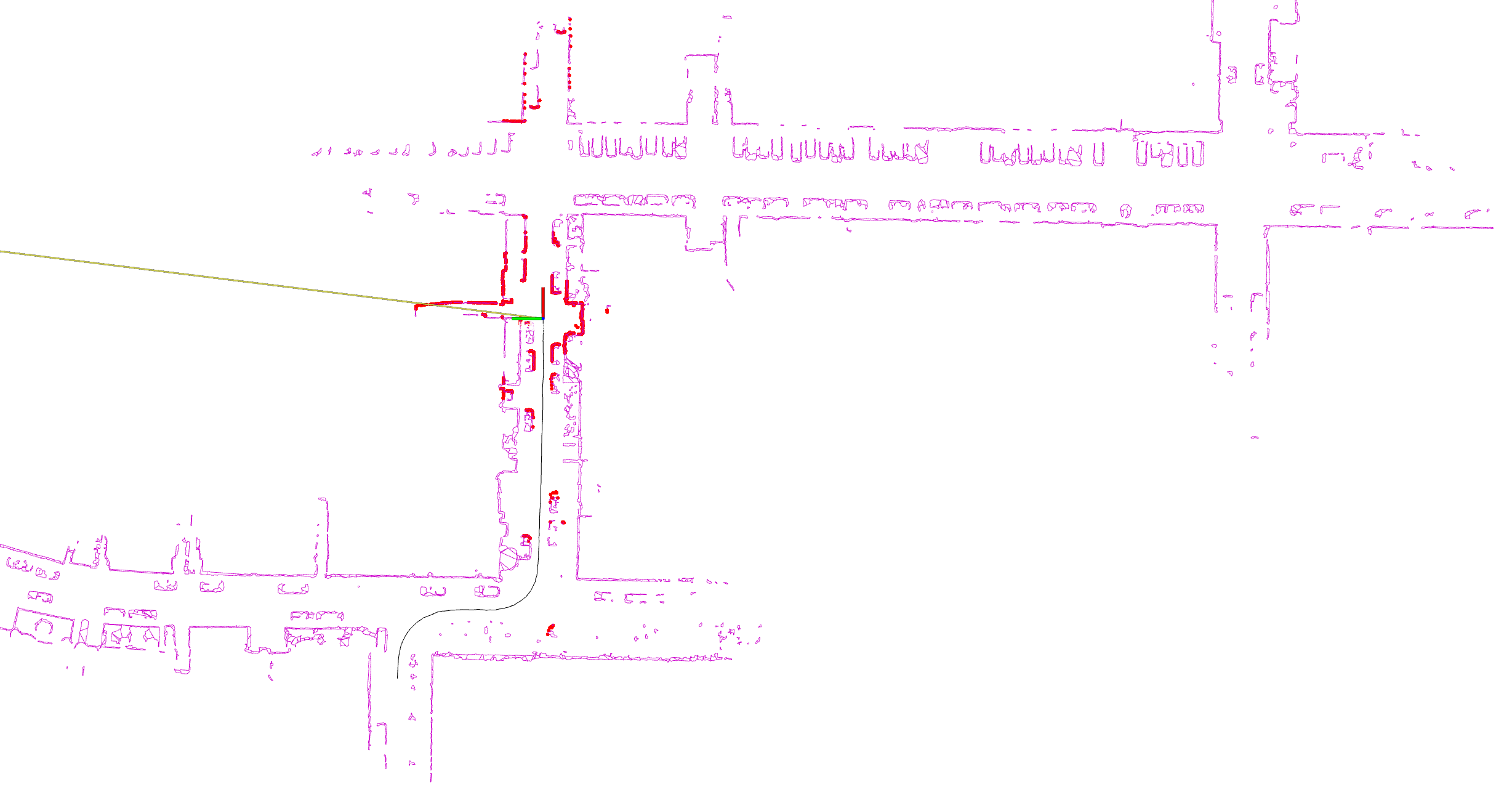}
		\includegraphics[width=4.2cm]{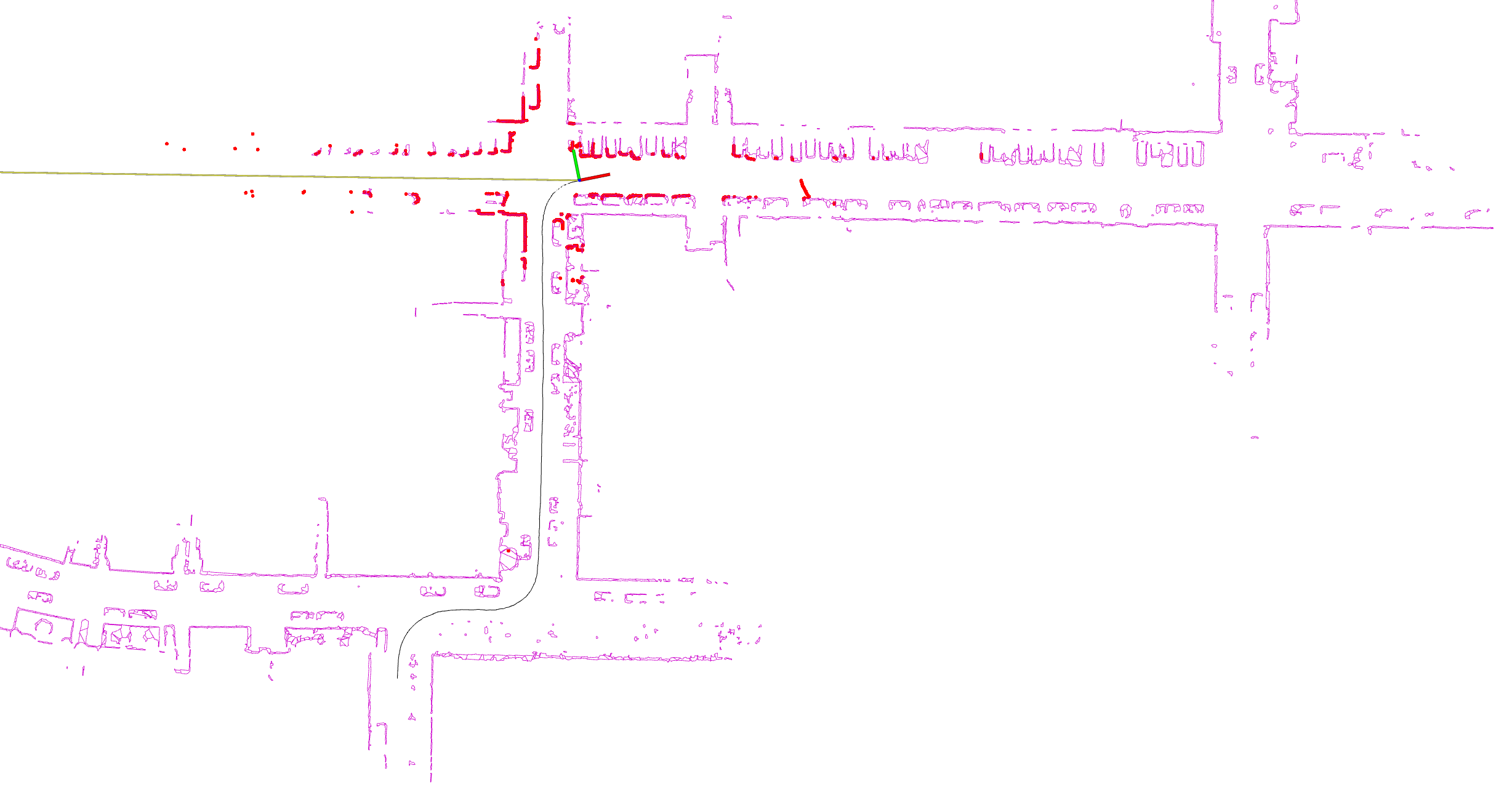}
		\includegraphics[width=4.2cm]{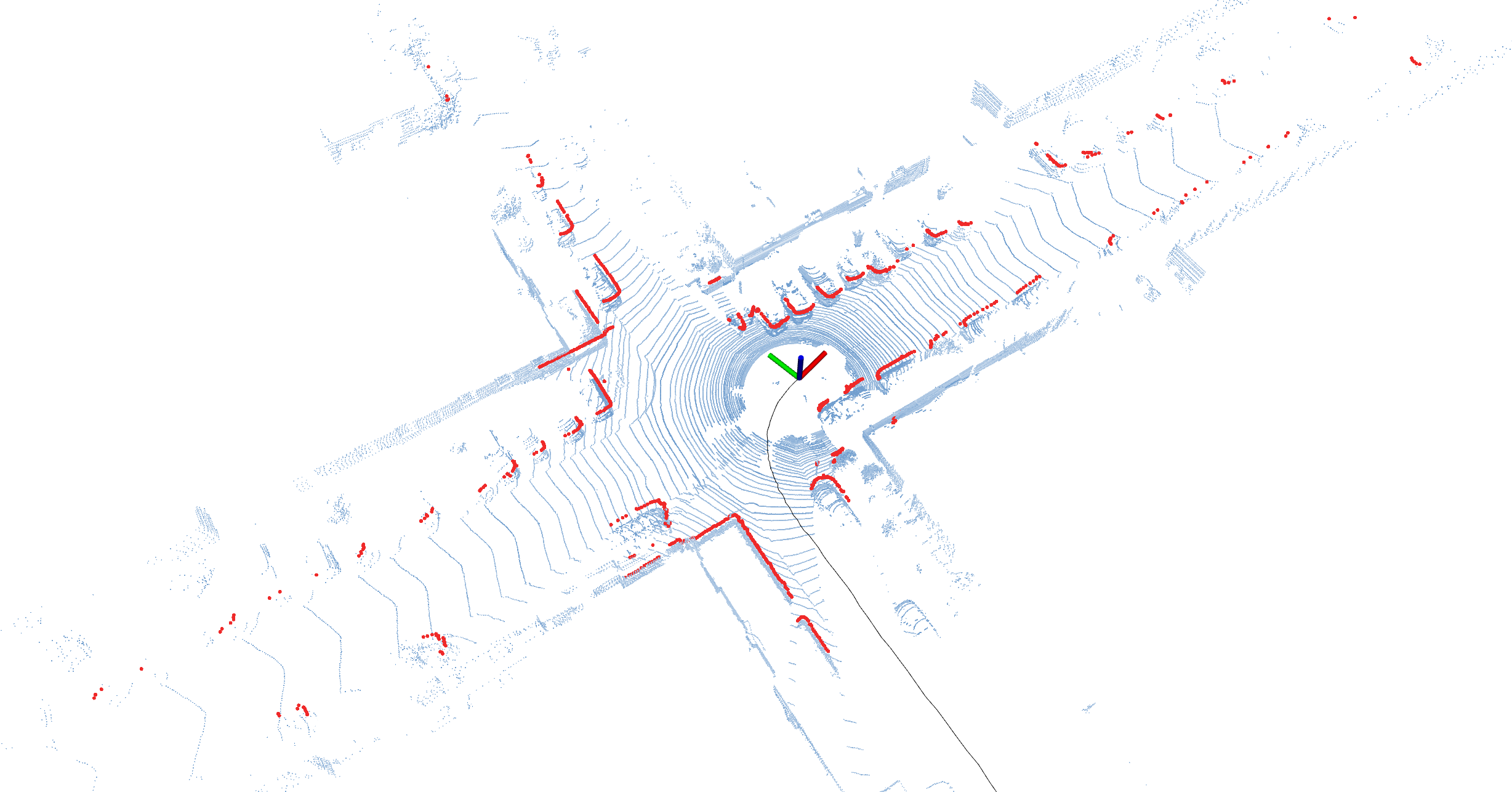}}
	\subfigure[Newer College dataset]{
		\includegraphics[width=4.2cm]{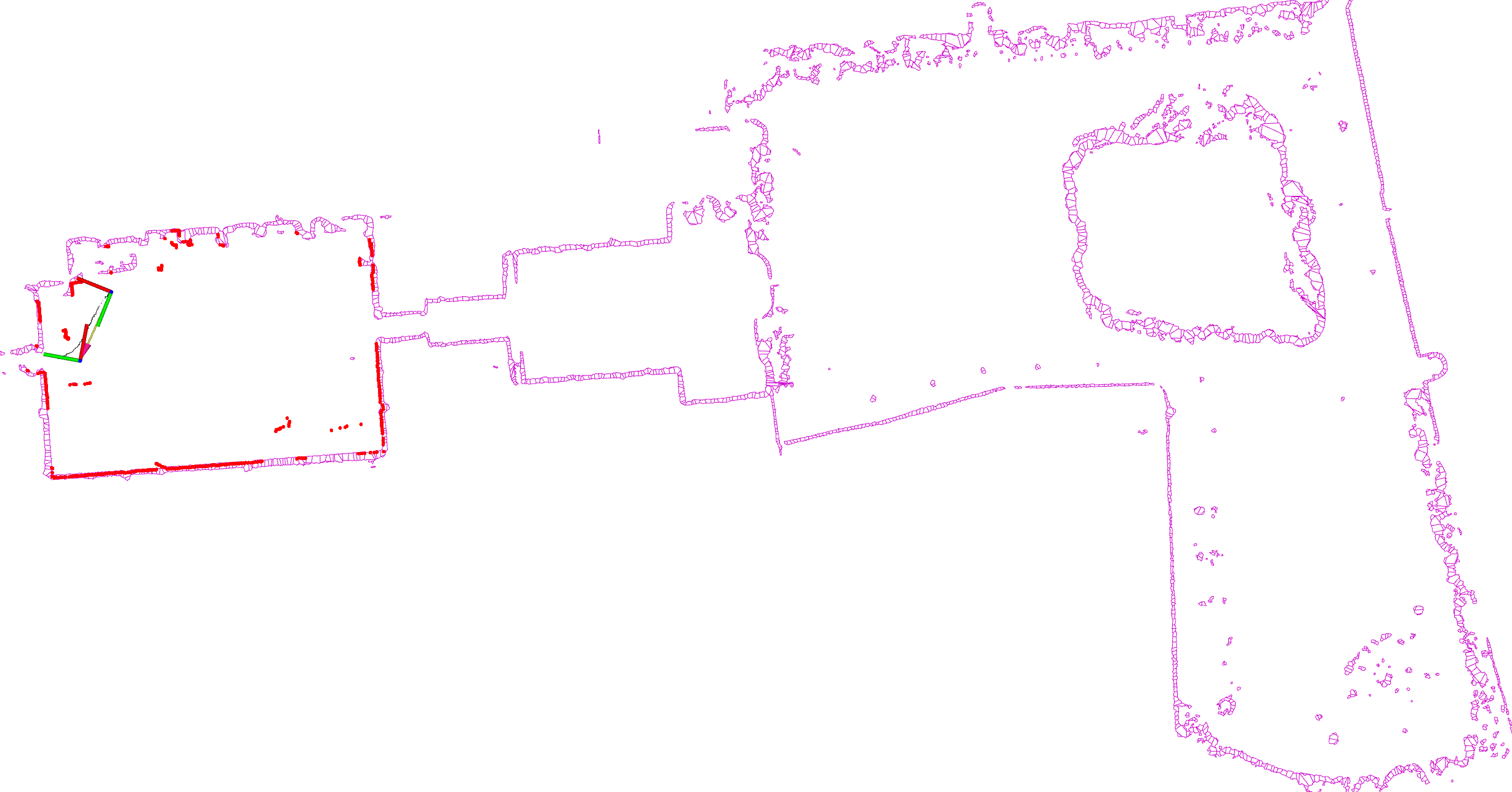}
		\includegraphics[width=4.2cm]{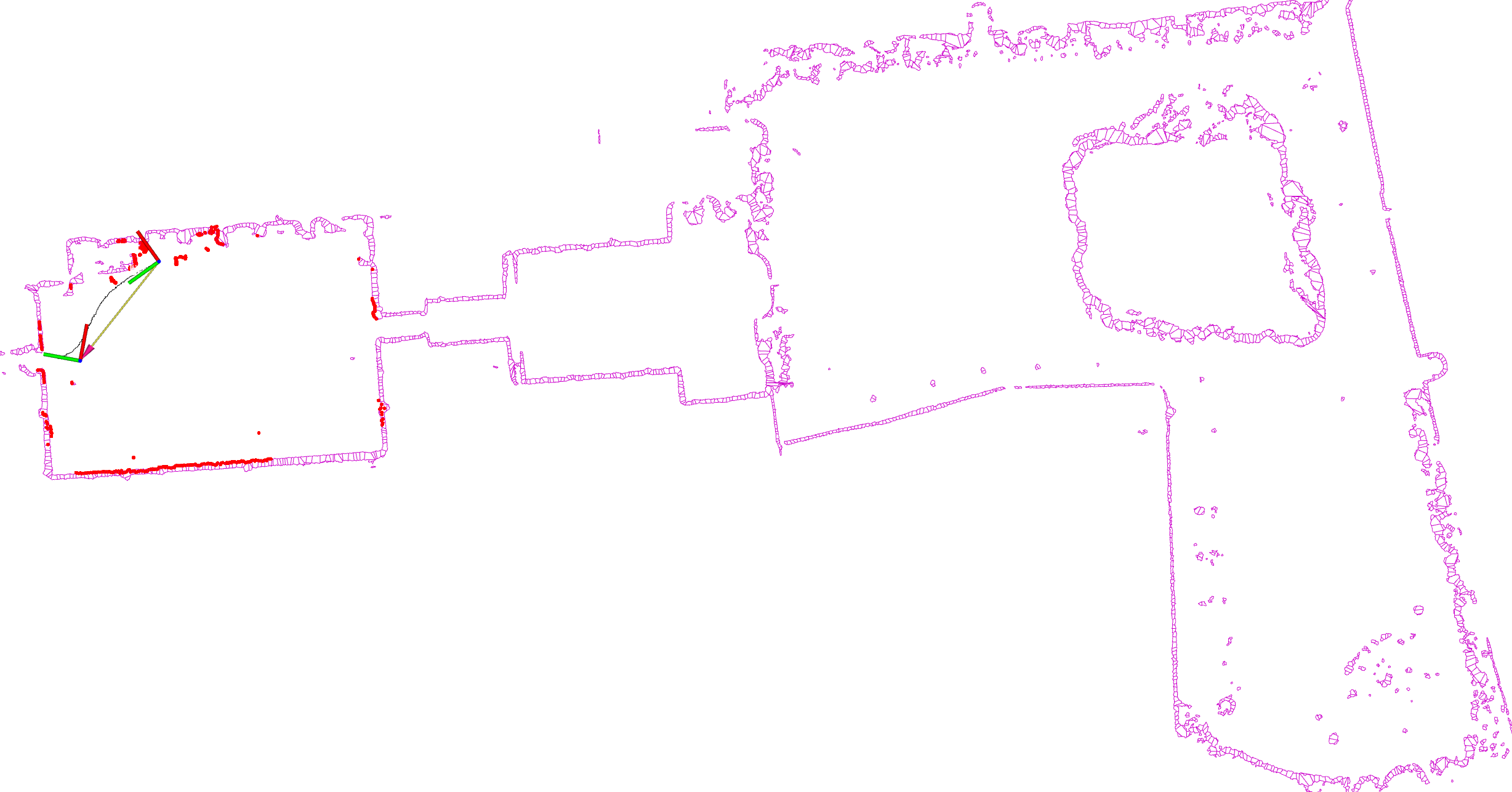}
		\includegraphics[width=4.2cm]{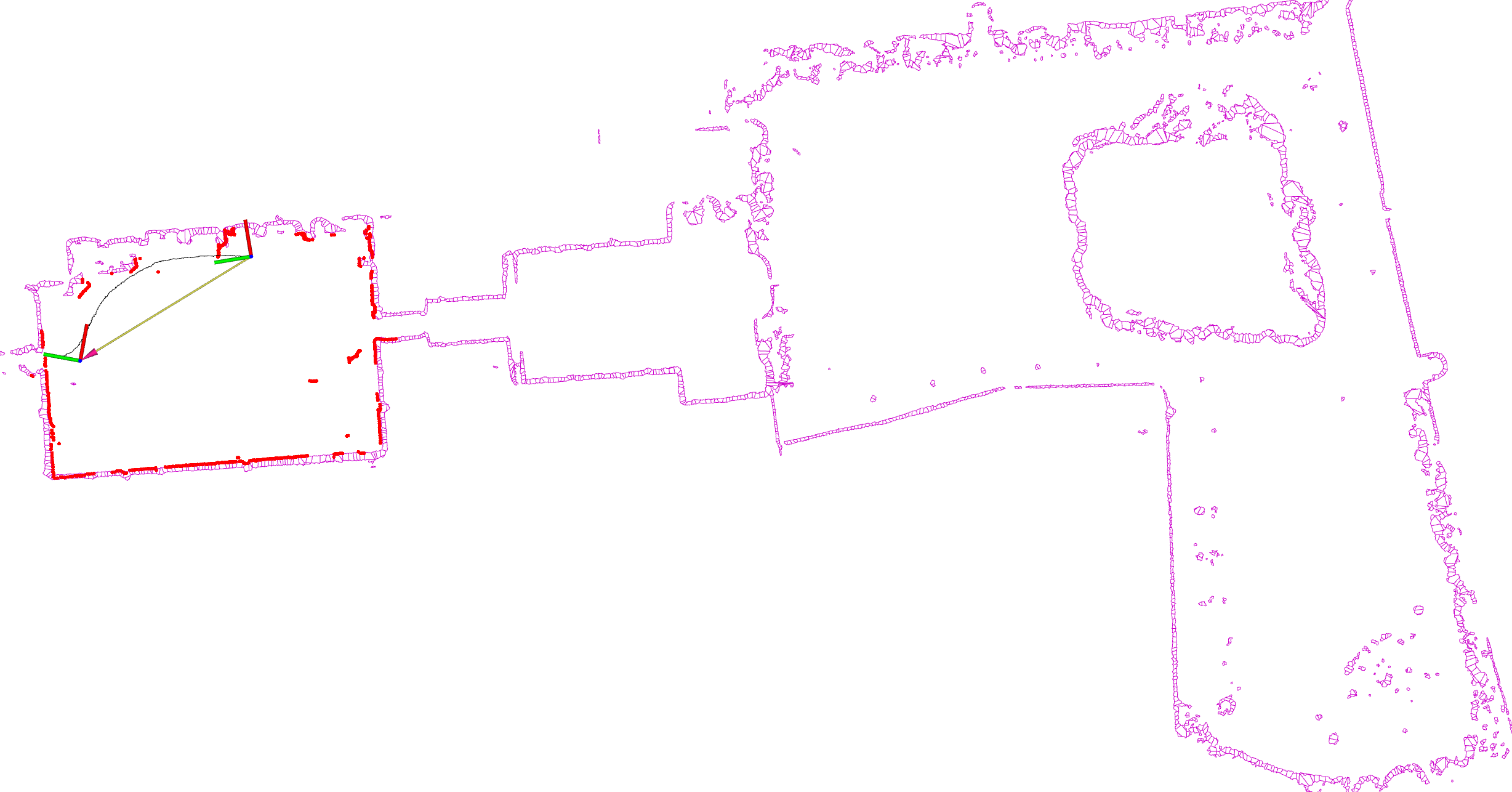}
		\includegraphics[width=4.2cm]{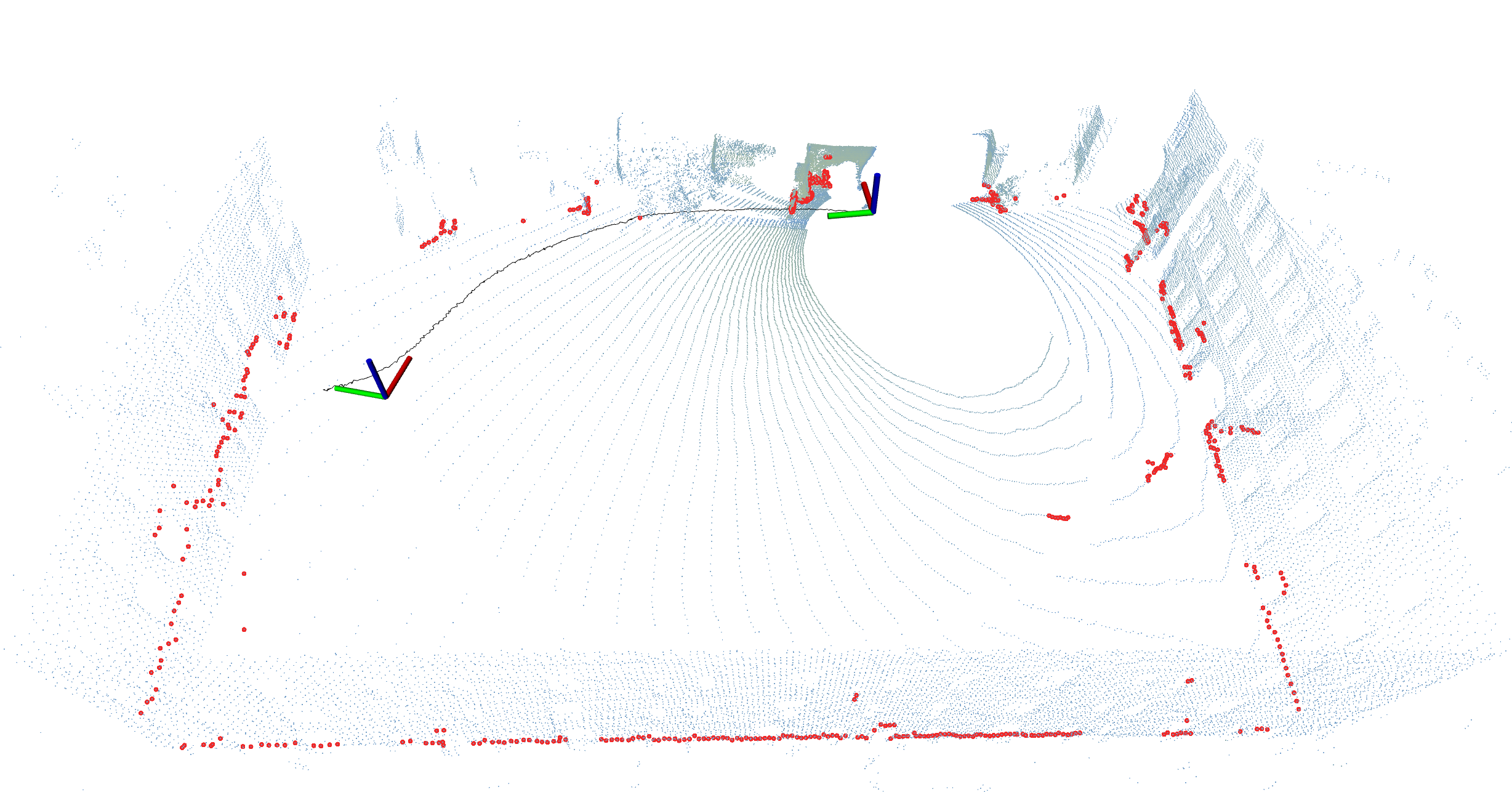}}
	\subfigure[Portable-building dataset]{
		\includegraphics[width=4.2cm]{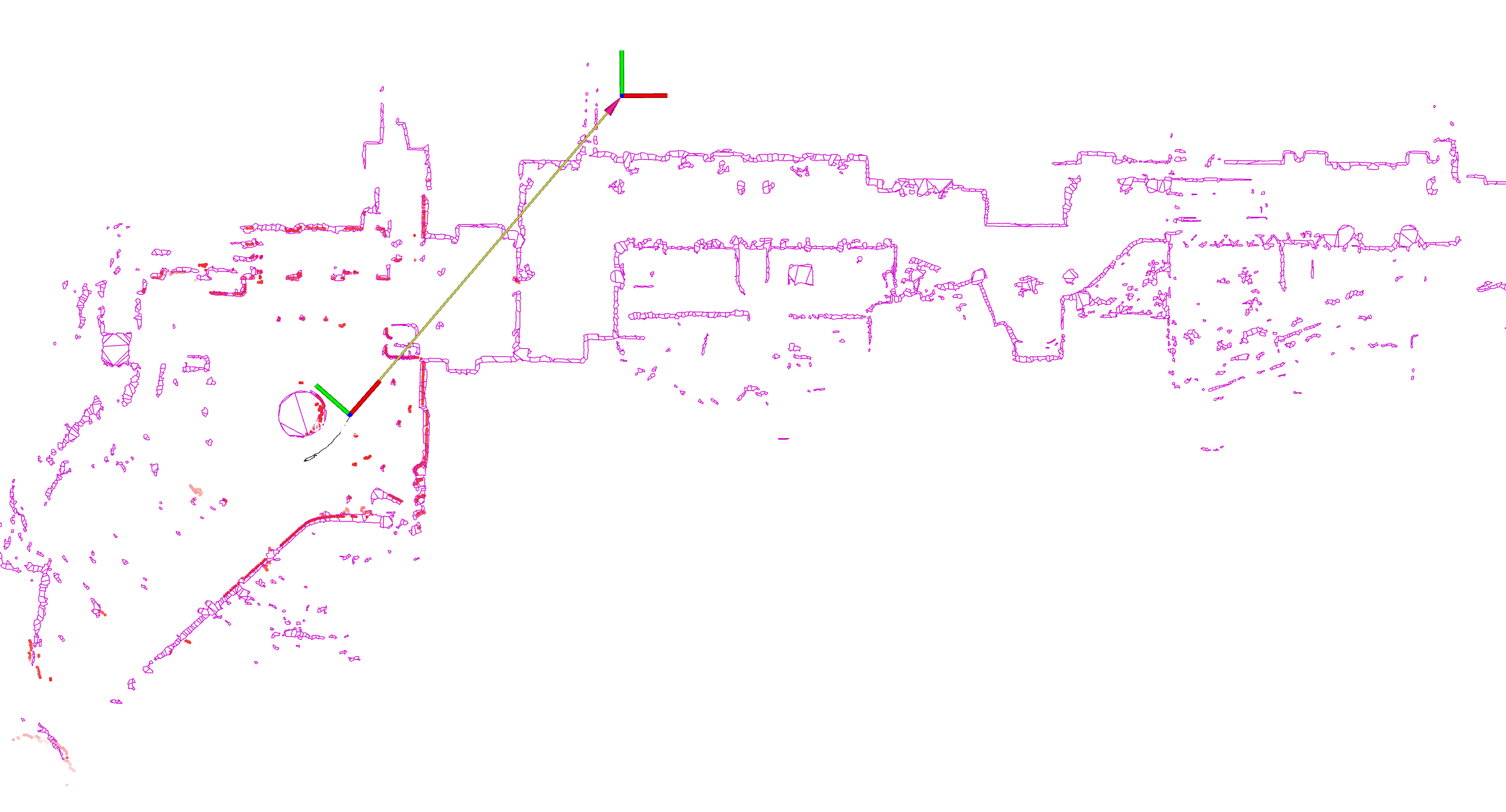}
		\includegraphics[width=4.2cm]{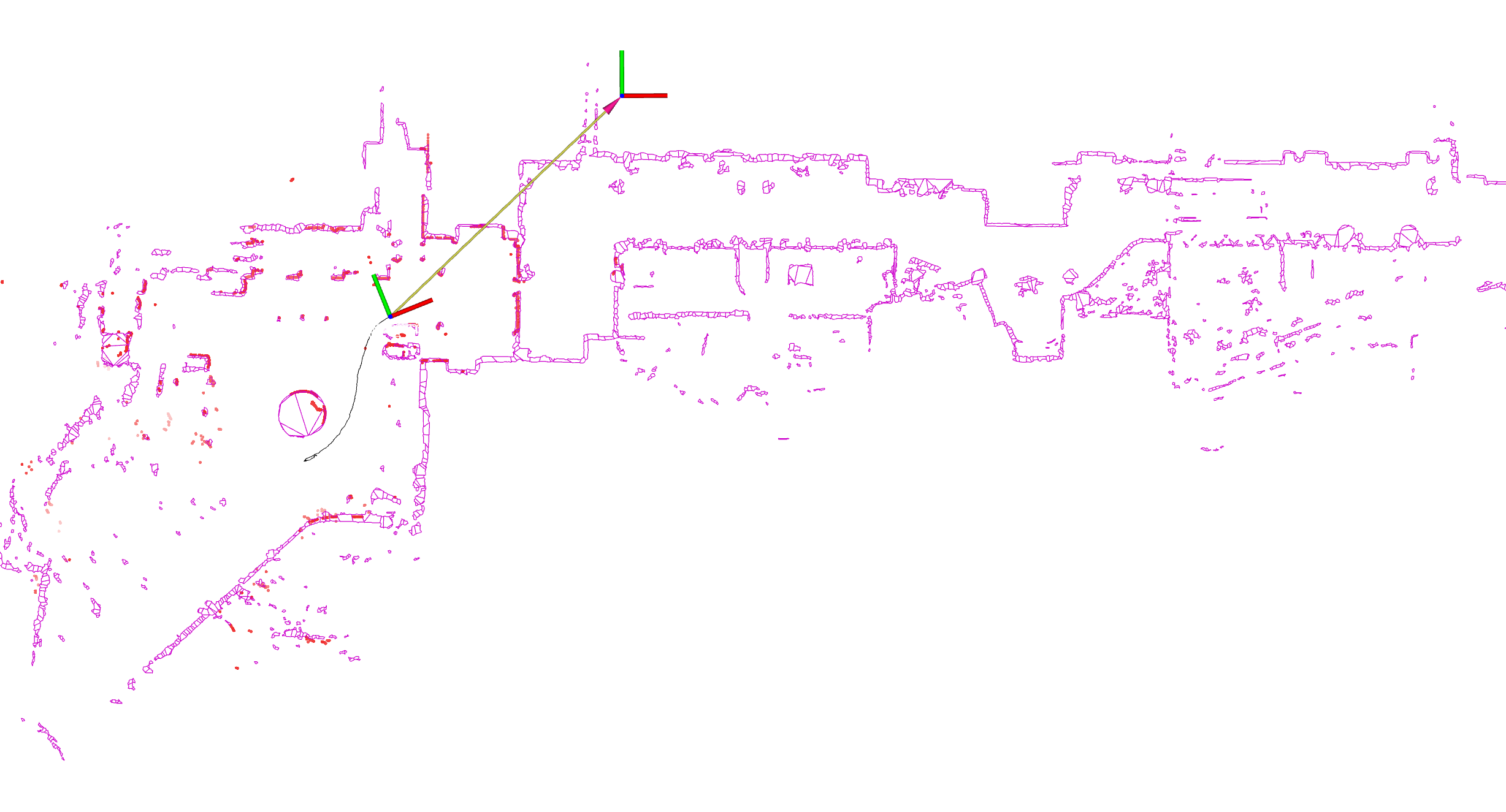}
		\includegraphics[width=4.2cm]{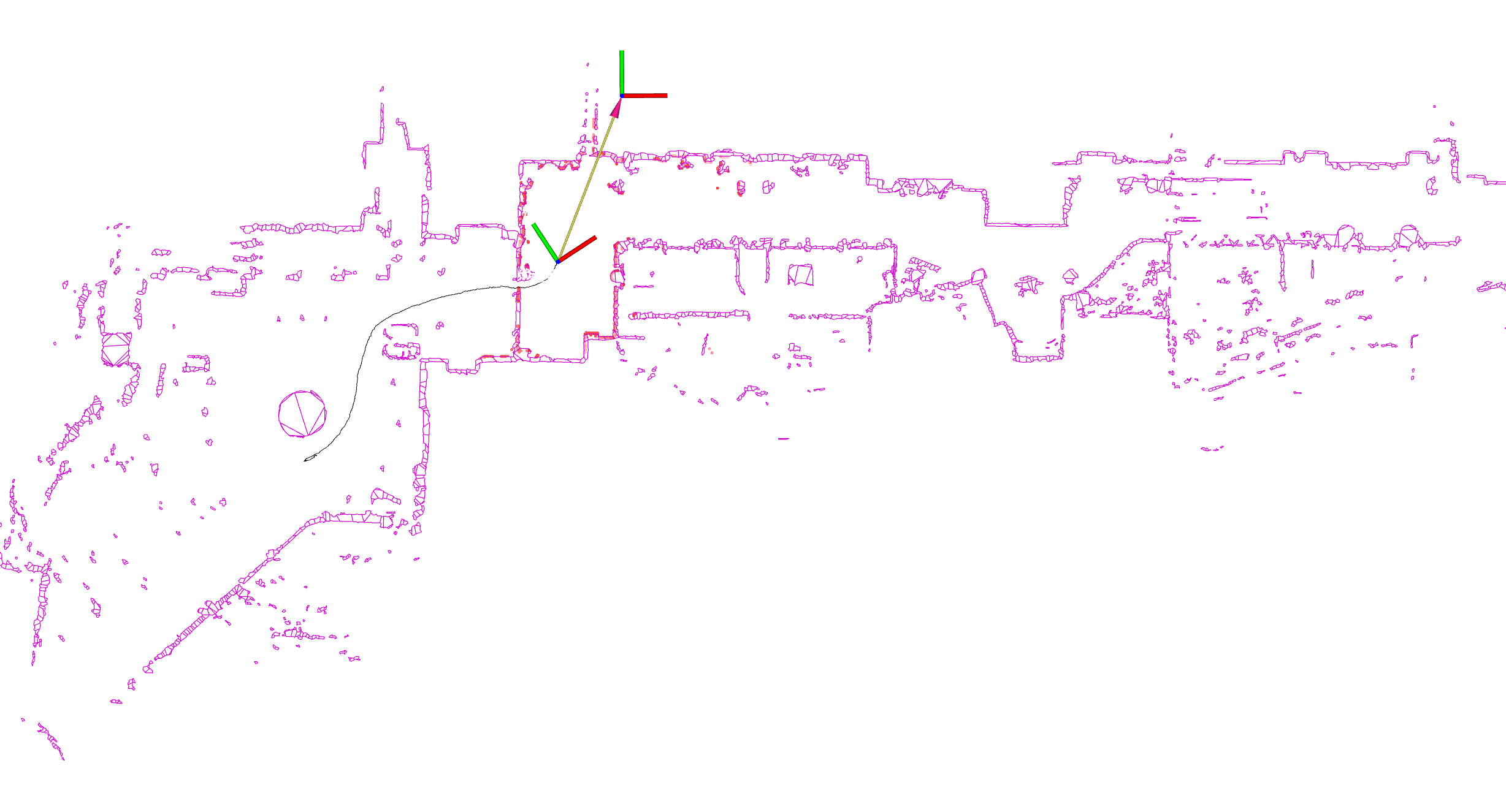}
		\includegraphics[width=4.2cm]{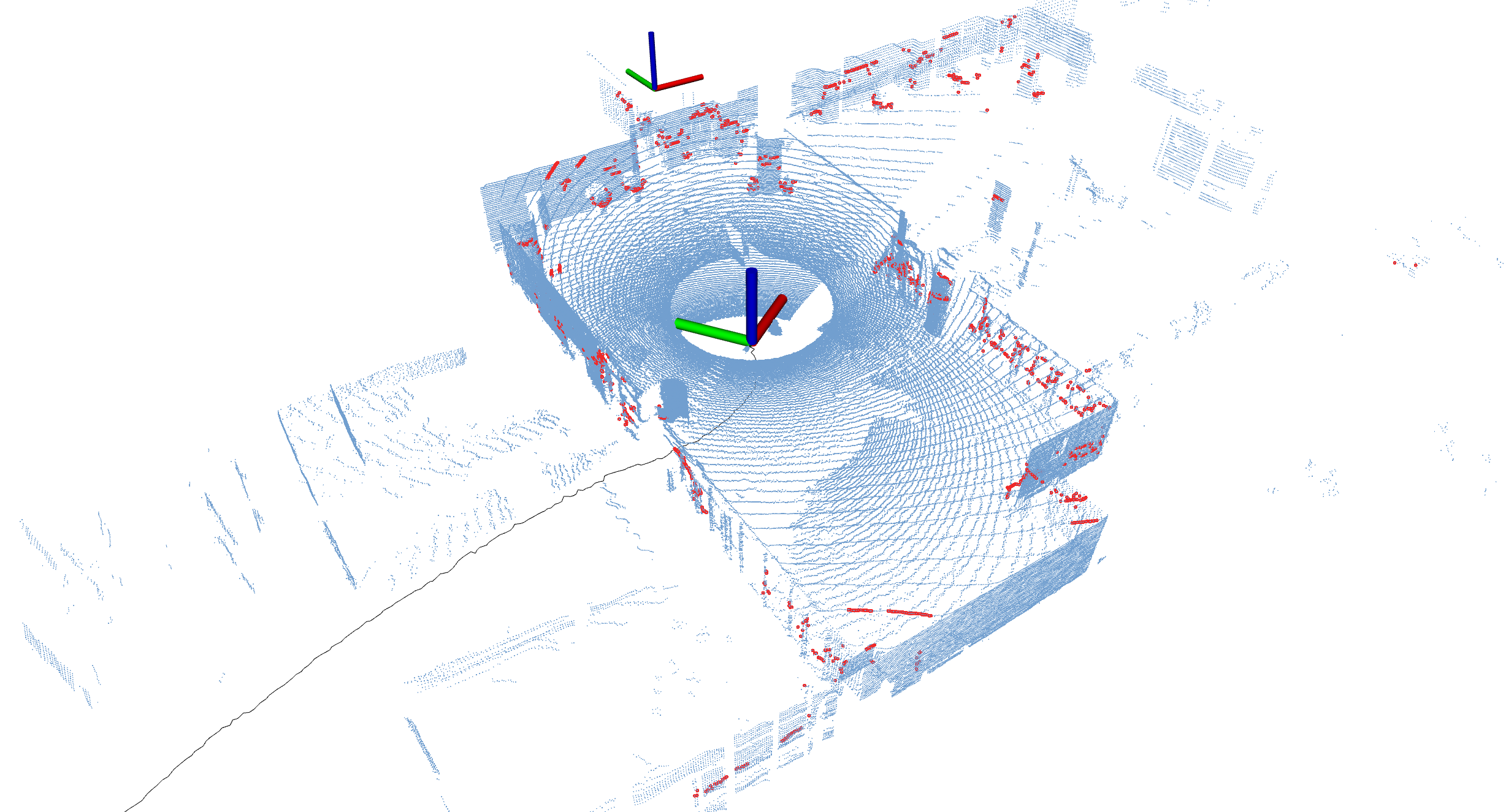}}
	\caption{The experimental results of the pose tracking process based on the prior maps for both self-recorded datasets and publicly available datasets. The thin black line denotes the trajectory of pose tracking, and the 2D scan information based on the nearest obstacle selection is represented by the red point, while all the magenta polygons make up the prior map. Furthermore, the three left subfigures in each row depict the process of pose tracking, and the right subfigure describes the 2D scan information corresponding to the third subfigure.}
	\label{Fig_16}
\end{figure*}

In this section, the qualitative experimental results of pose tracking based on the prior polygon map are provided. At first, based on the experimental configurations for self-recorded datasets and publicly available datasets, the corresponding prior polygon maps for the proposed ERPoT are obtained. Subsequently, the pose tracking process based on these prior maps is carried out and illustrated in Fig. \ref{Fig_16}.

\begin{figure*}[!htb]
	\centering
	{\includegraphics[width=3.54cm]{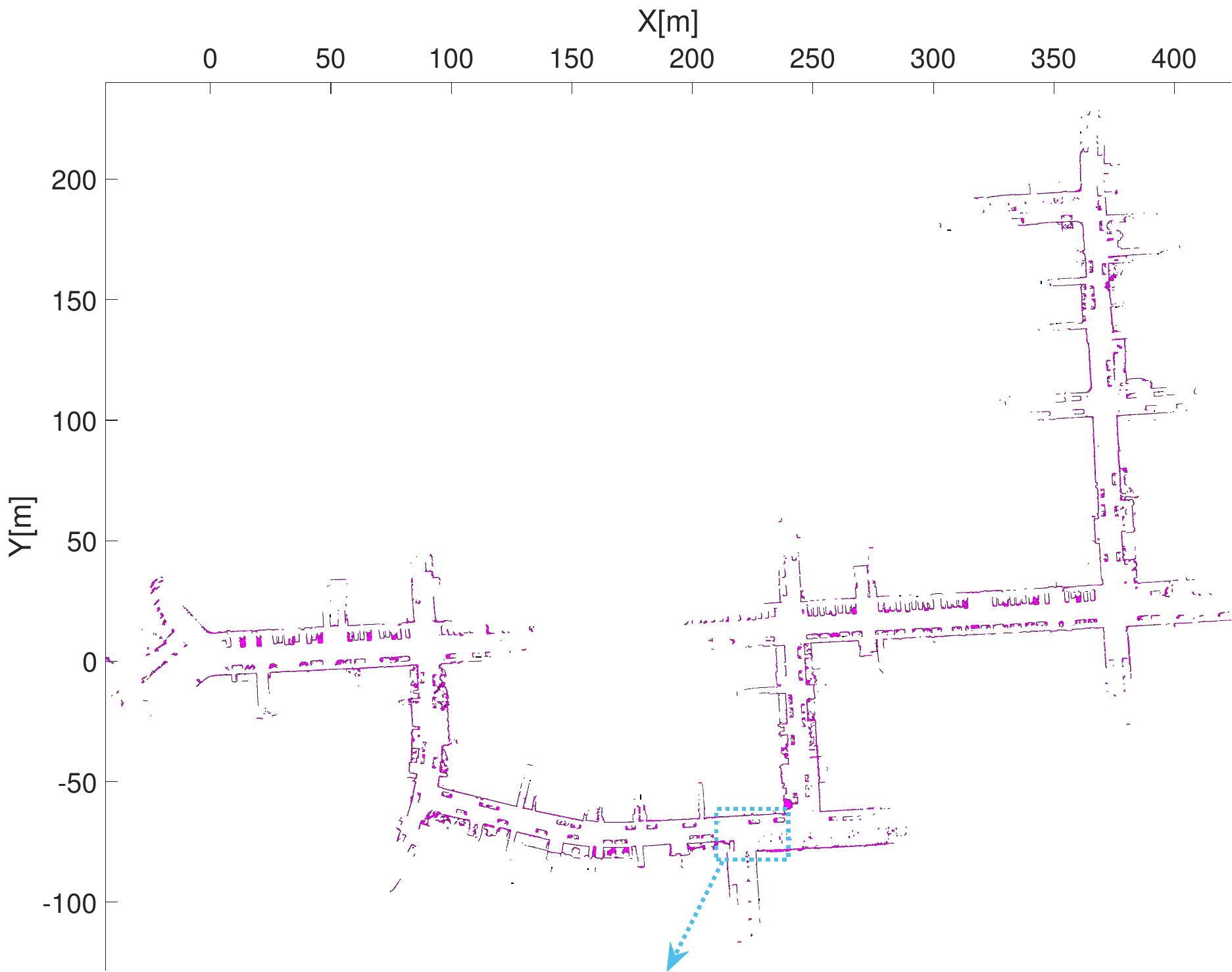}}
	{\includegraphics[width=3.54cm]{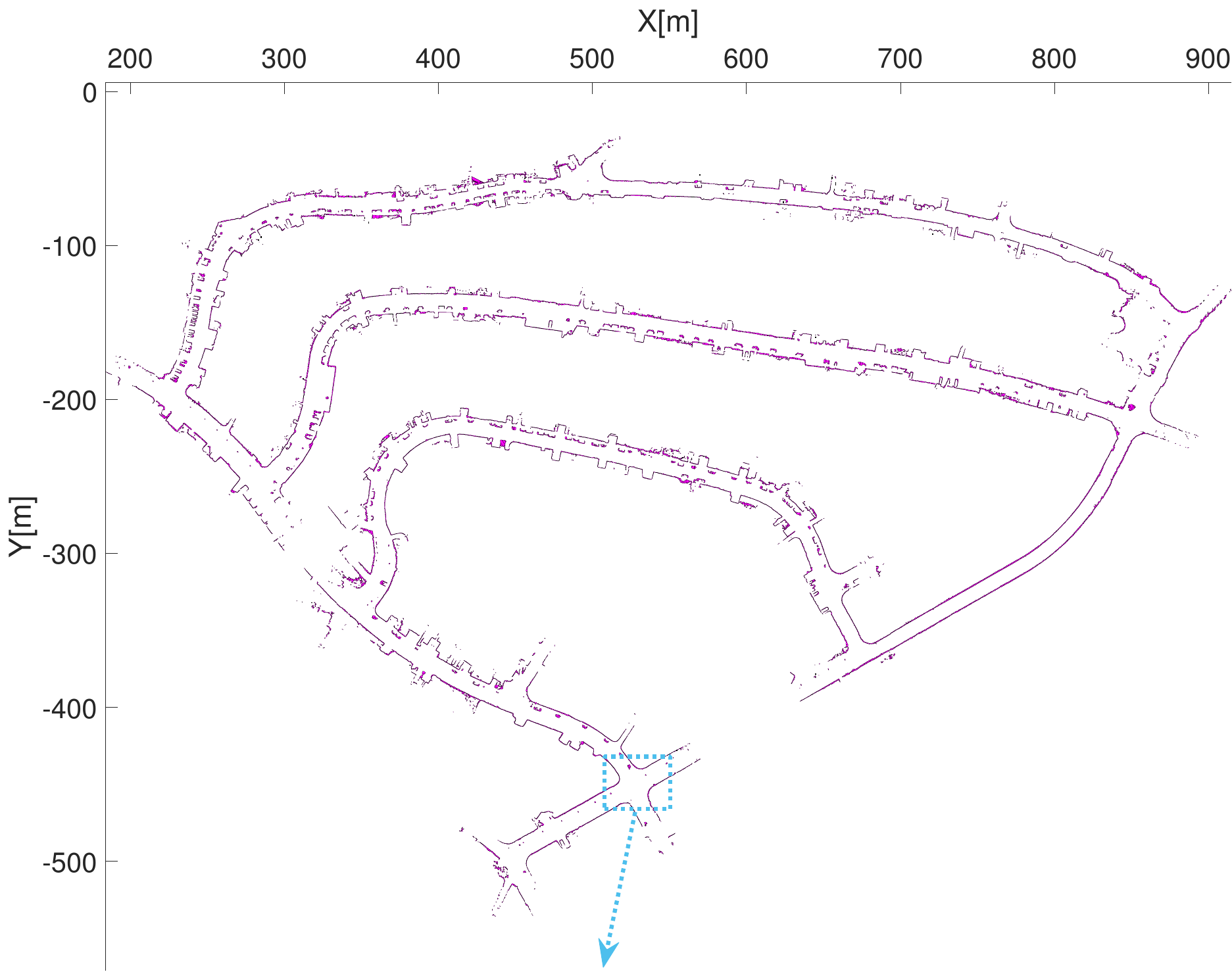}}
	{\includegraphics[width=3.54cm]{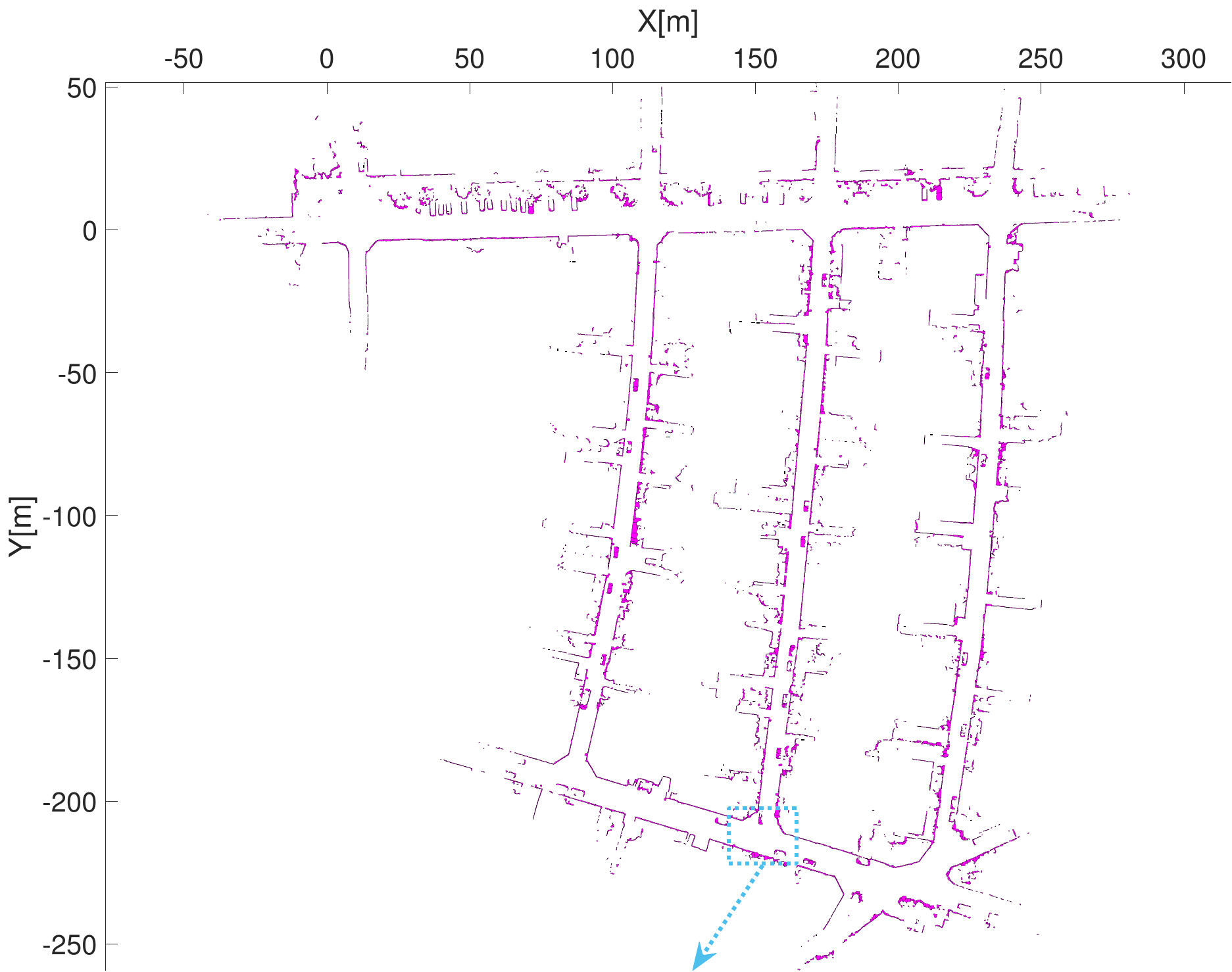}}
	{\includegraphics[width=3.54cm]{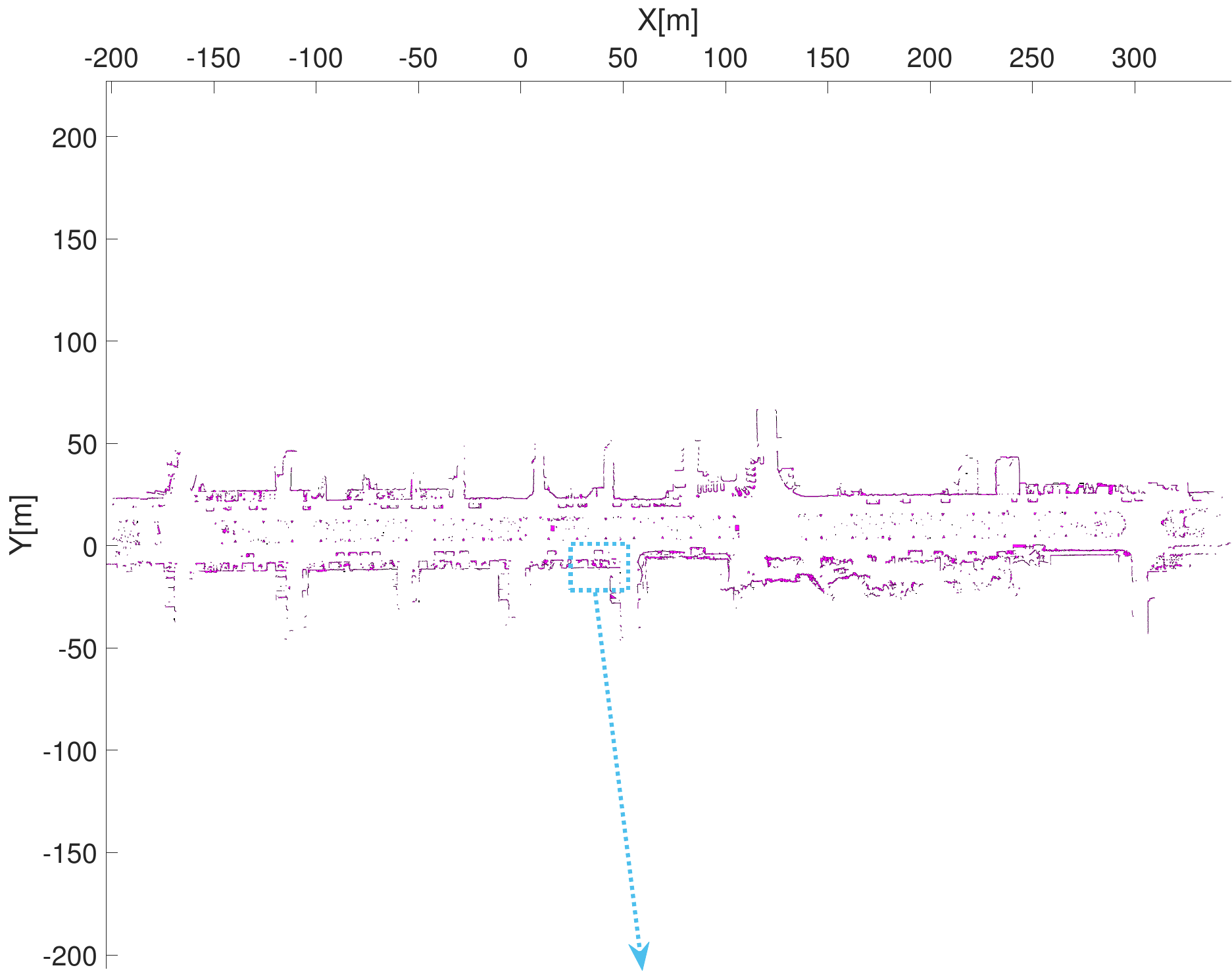}}
	{\includegraphics[width=3.54cm]{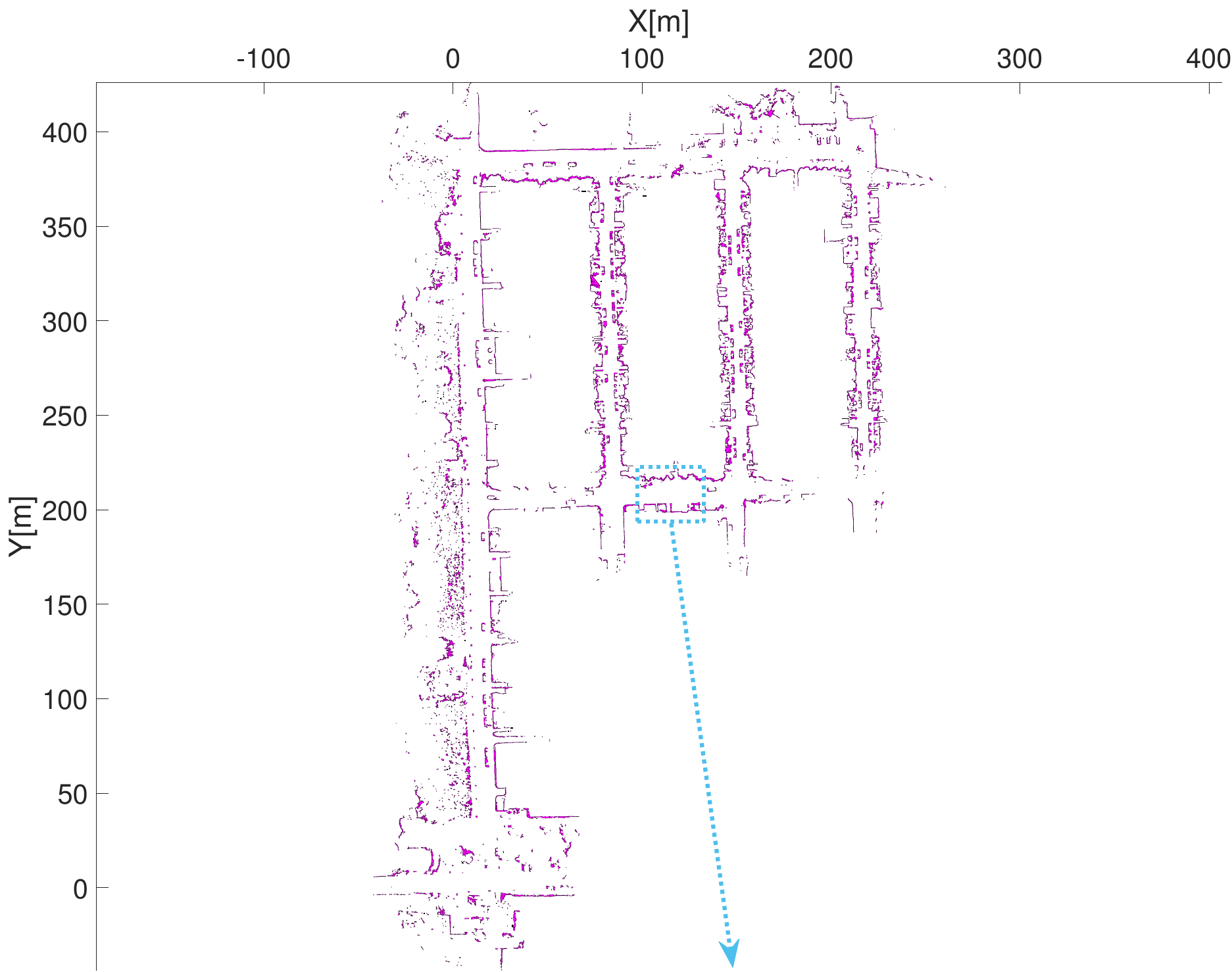}}\\
	\subfigure[KITTI00]{\includegraphics[width=3.54cm]{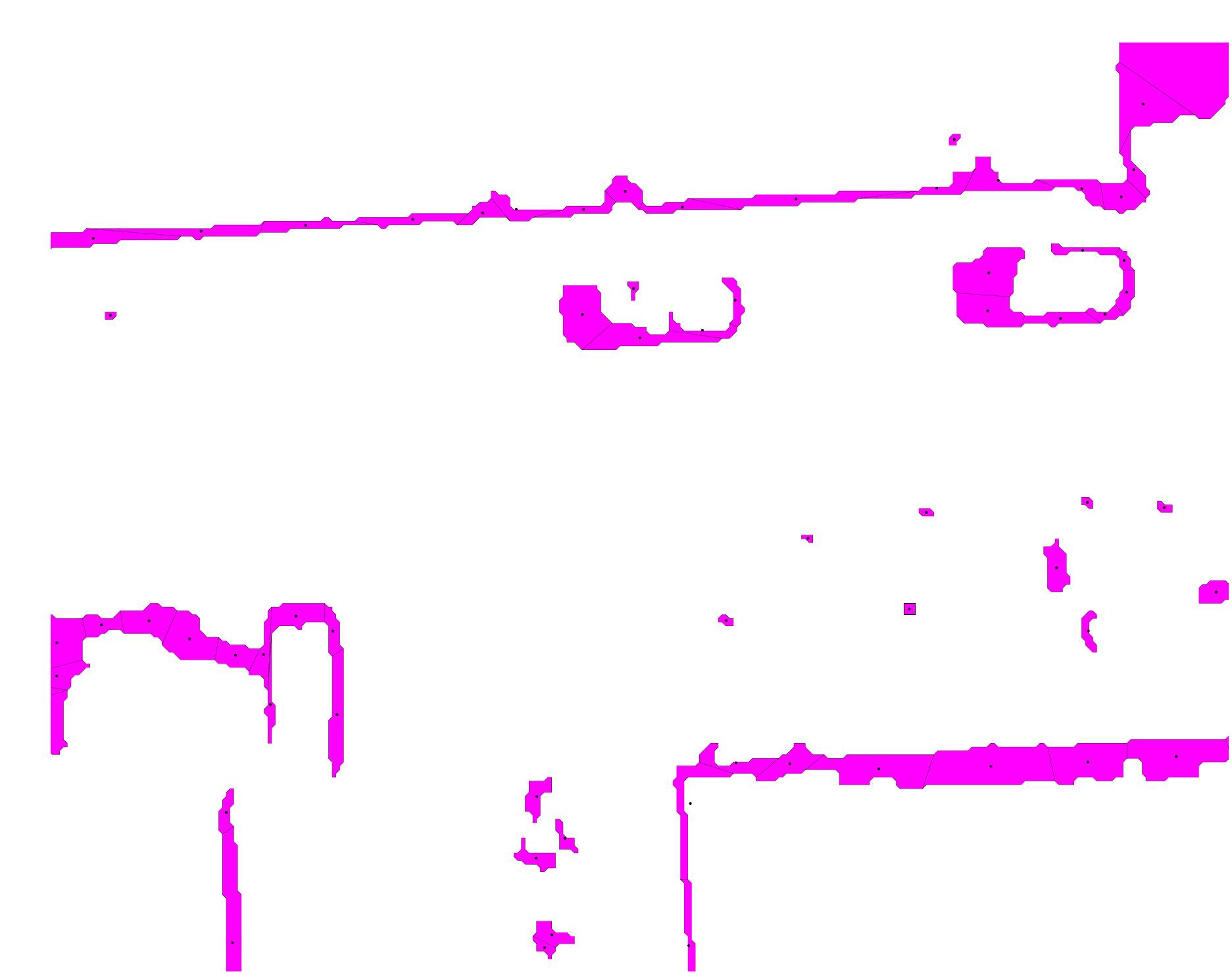}}
	\subfigure[KITTI02]{\includegraphics[width=3.54cm]{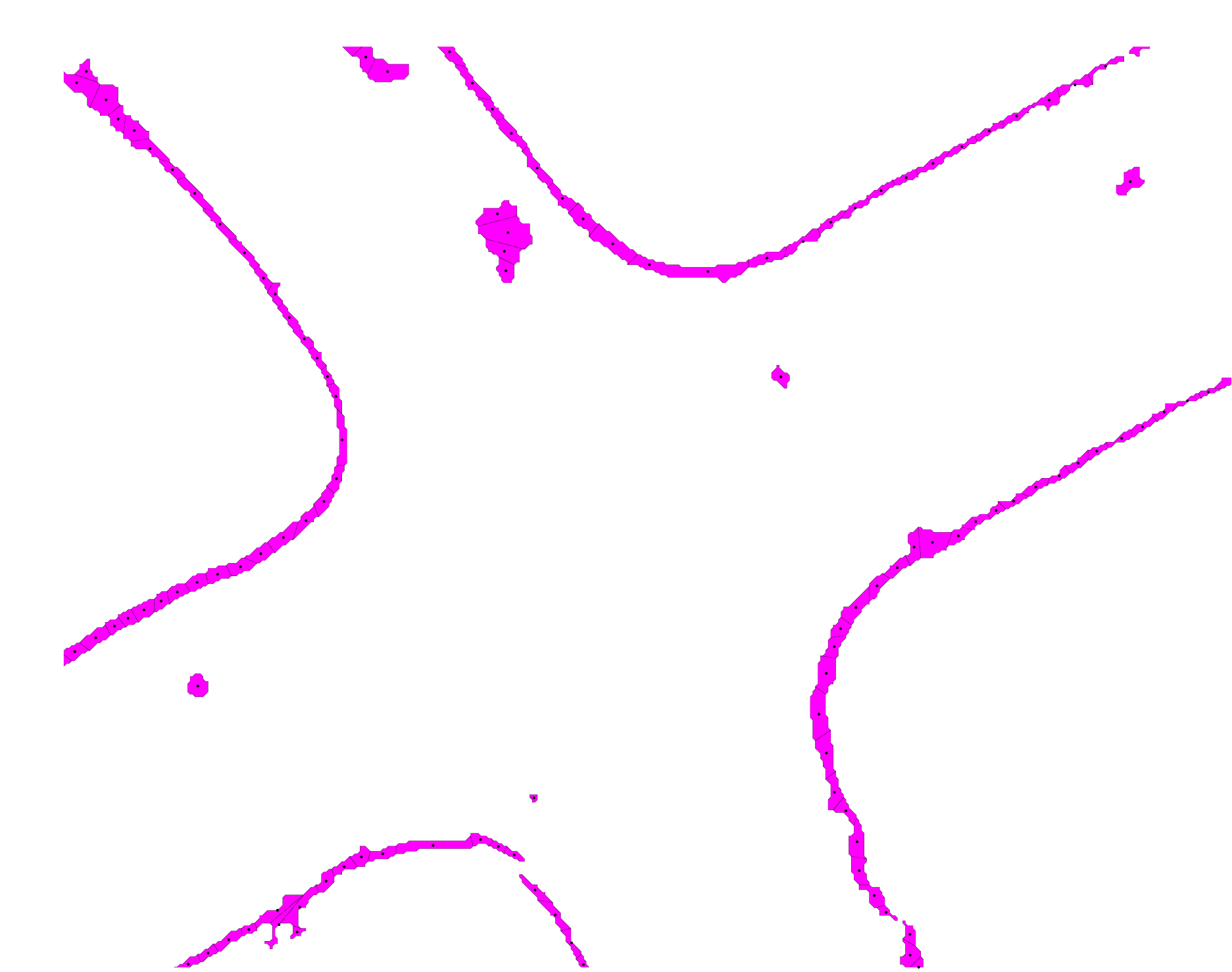}}
	\subfigure[KITTI05]{\includegraphics[width=3.54cm]{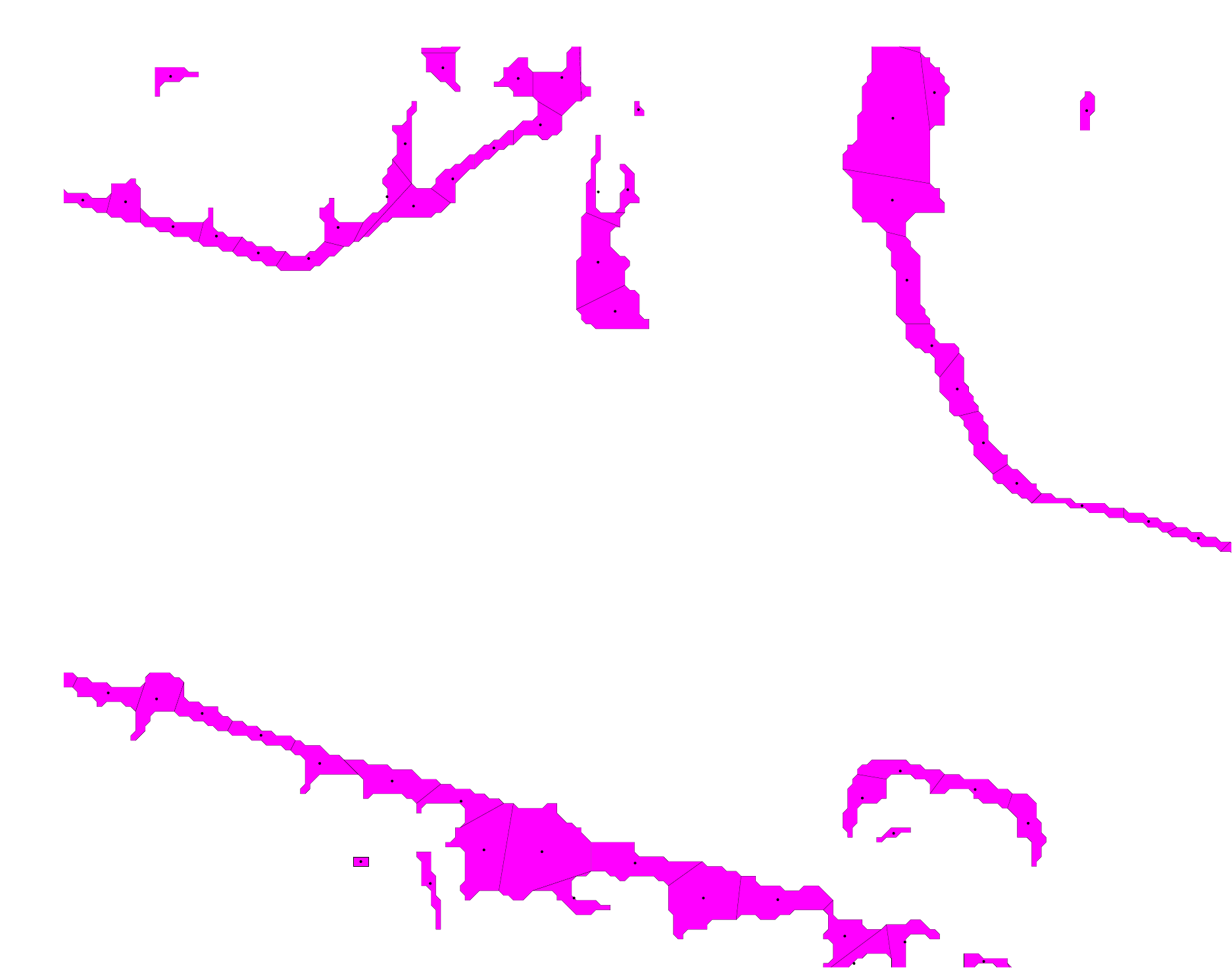}}
	\subfigure[KITTI06]{\includegraphics[width=3.54cm]{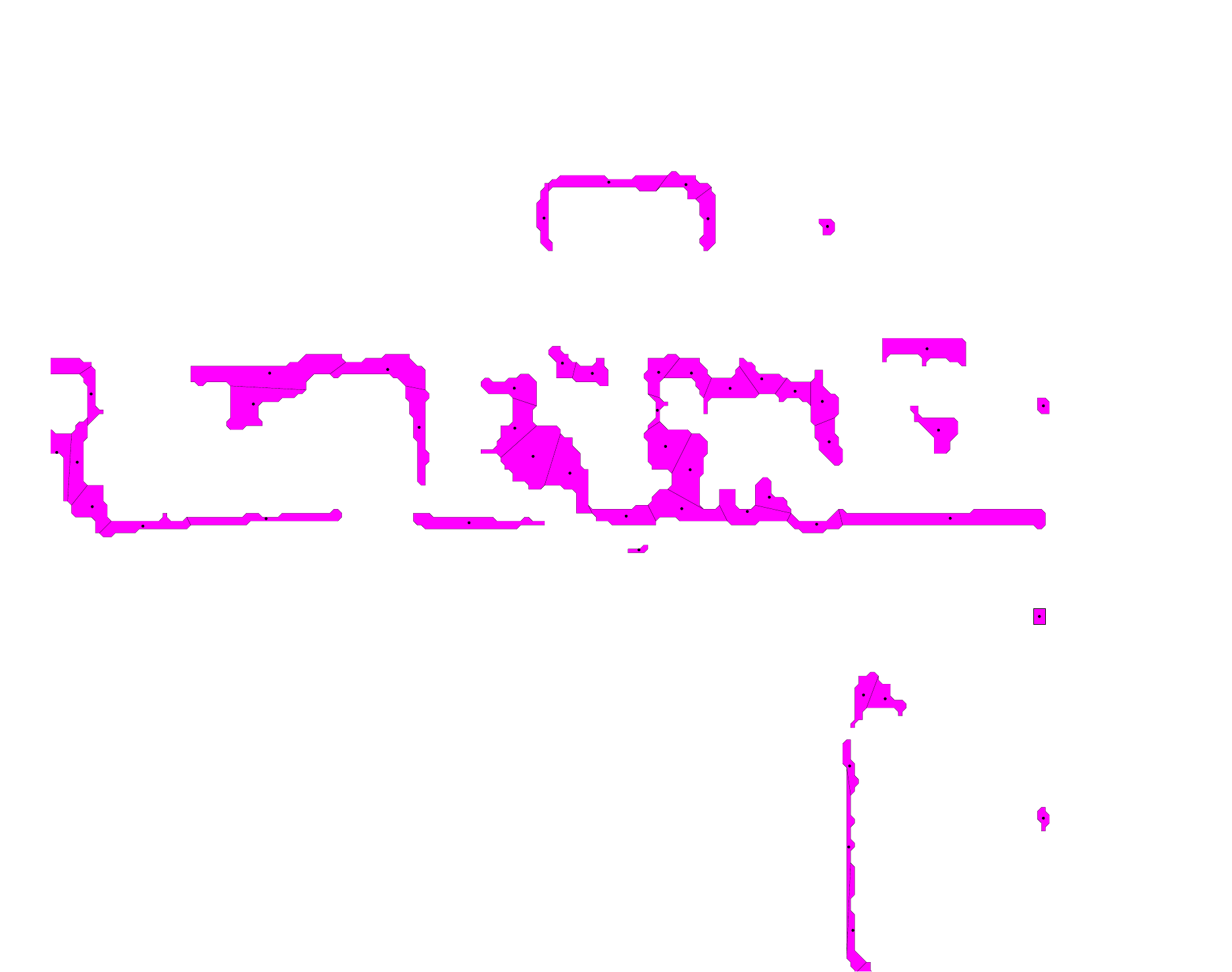}}
	\subfigure[KITTI08]{\includegraphics[width=3.54cm]{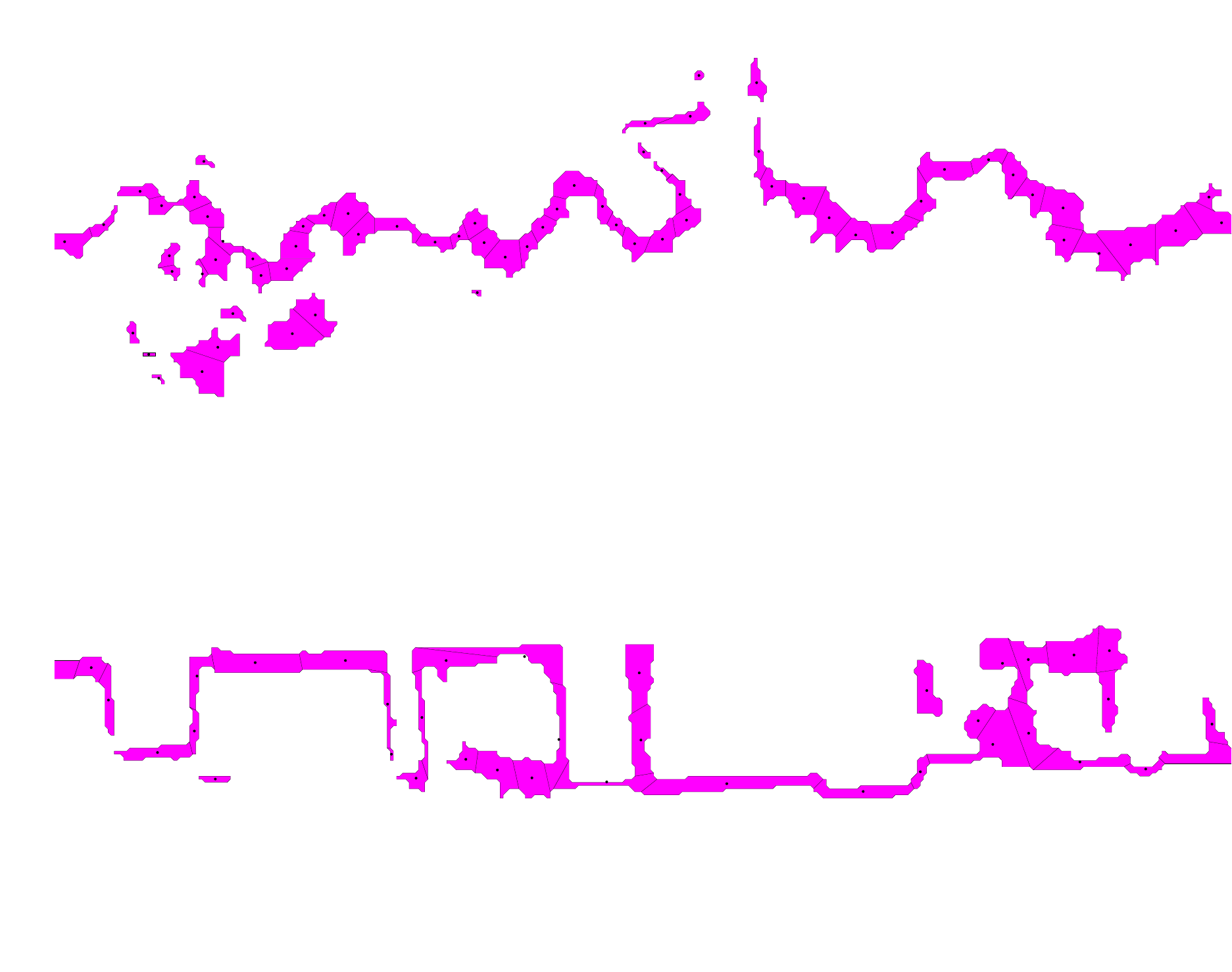}}\\
	\caption{Polygon maps for the KITTI dataset. Each polygon map is composed of multiple polygons and the corresponding centroid points, marked by the magenta filled polygons and black points. In addition, the area within the blue box is enlarged for display, as shown in the bottom subfigure. Due to the utilization of vector graphics for map rendering, the polygon map details can be viewed upon zooming in.}
	\label{Fig_17}
\end{figure*}

For Self dataset-01, the process of pose tracking is represented by the three left subfigures in Fig. \ref{Fig_16}(a), and the obtained trajectory is smooth. As shown in the right subfigure in Fig. \ref{Fig_16}(a), the obtained 2D scan information is sufficient to describe the surrounding environmental information, including vegetation and buildings, which has a good matching relationship with the prior polygon map. Furthermore, within the Self dataset-02 environment, which includes uphill and downhill terrains, the proposed approach, while estimating only the 3-DoF $(x, y, \theta)$ of the pose, can still accurately estimate the pose during uphill and downhill movements, as demonstrated by the smooth trajectory in the Fig. \ref{Fig_16}(b).

On the other hand, for the KITTI00 dataset, our proposed ERPoT can realize effective and reliable pose tracking, even in scenarios where the collection platform navigates at high velocities within road environments. This capability is evident from the smooth trajectory illustrated in Fig. \ref{Fig_16}(c). In addition, in both Fig. \ref{Fig_16}(d) and Fig. \ref{Fig_16}(e), for the handheld data collection platform with different LiDAR sensors, the smooth pose estimation result can also be obtained.

Drawing from the aforementioned qualitative experimental outcomes, the proposed approach ERPoT has demonstrated its capacity to achieve effective and reliable pose tracking.

\subsection{Experimental evaluation metrics}
Before the detailed evaluation, this section provides an overview of the qualitative measures used to assess performance. A comprehensive set of metrics evaluates the pose tracking process based on a lightweight polygon map.

\textbf{Translation error and rotation error}: For pose tracking using a prior map, the system has a global reference with a global position and yaw angle. The accuracy of pose estimation is evaluated using the Absolute Pose Error (APE) \cite{ZZhang}. In this study, the $i$-th estimated pose is denoted by $\mathbf{ {s}}_i = \left[ {{{ x}_i}, {{ y}_i}, {{ \theta }_i}} \right]$, and the ground truth is $\mathbf{\hat {s}}_i = \left[ {{{\hat x}_i}, {{\hat y}_i}, {{\hat \theta }_i}} \right]$, then the error between $\mathbf{\hat {s}}_i$ and $\mathbf{s}_i$ can be parameterized as $\Delta \mathbf{s}_i = \left[ {\Delta {x_i},\Delta {y_i},\Delta {\theta_i}} \right]$, and satisfies the following equation
\begin{align}
{\emph{\textbf{T}}}\left( \Delta \mathbf{s}_i \right) = {\emph{\textbf{T}}}{\left( \mathbf{\hat {s}}_i \right)^{ - 1}} \cdot {\emph{\textbf{T}}}\left( \mathbf{ {s}}_i \right),
\end{align}where the transformation ${\emph{\textbf{T}}}(\cdot) \in SE(2)$.

\textbf{Runtime of pose tracking}: In the realm of approach performance evaluation, runtime is a crucial metric, reflecting the efficiency of algorithms in processing information in various environments. Assessing runtime is essential for understanding the practical applicability of an approach, especially in time-sensitive environments. In comparative experiments, we report only CPU times, focusing on CPU performance to ensure consistency across different algorithms. The times cover the full pose tracking cycle, including data preprocessing, feature extraction, feature matching, and pose estimation.

\textbf{Size of the prior map}: Moreover, as a foundational element for accurate pose tracking, the prior map is designed to balance detail and efficiency, especially in large-scale, resource-constrained environments. We advocate for a map that is rich in essential features yet optimized for minimal storage and swift retrieval, facilitating real-time pose estimation without compromising computational resources.

\subsection{Quantitative comparative results using public datasets}

Based on the experimental configurations and metrics, comparative experiments are conducted between ERPoT and six other approaches. The prior polygon and point cloud maps are constructed using the LiDAR sequences listed in Tab. \ref{table_3} for pose tracking. For SM\_MCL, a distance field representation, which requires significant precomputation based on the point cloud map, is used for efficient pose tracking. In contrast, the polygon maps for KITTI sequences 00, 02, 05, 06, and 08, shown in Figure \ref{Fig_17}, are lightweight and compact. They capture essential spatial cues while omitting unnecessary details. This design allows ERPoT to operate with minimal memory usage, which is crucial for applications with hardware limitations.

\begin{table*}[!htb]\small
	\renewcommand\arraystretch{0.7}
	\centering
	\caption{Experimental results of different approaches using the KITTI dataset}
	\setlength{\tabcolsep}{1.2mm}{
	{\fontsize{6pt}{6pt}
	\begin{tabular}{|l|c|c|c|c|c|c|c|c|c|c|c|}
			\hline
			\multirow{2}{*}{\textbf{Dataset No.}} &\textbf{Approach} &\textbf{Prior map} &\multicolumn{3}{c|}{\textbf{Translational error (cm)}} &\multicolumn{3}{c|}{\textbf{Rotational error (deg)}} &\multicolumn{3}{c|}{\textbf{Runtime (ms)}} \\
			\cline{4-12}
			 &\textbf{name}   &\textbf{size (MB) $\downarrow$} &\textbf{Max} $\downarrow$  &\textbf{Mean} $\downarrow$  &\textbf{RMSE} $\downarrow$  &\textbf{Max} $\downarrow$ &\textbf{Mean} $\downarrow$ &\textbf{RMSE} $\downarrow$ &\textbf{Max} $\downarrow$ &\textbf{Mean} $\downarrow$ &\textbf{SD}$\downarrow$ \\
			\hline \hline
			\multirow{7}{*}{\textbf{KITTI00-1}}
			& \textbf{ALOAM\_MCL}        &\multirow{4}{*}{$56.44$}  &$92.85$  &$58.44$  &$63.70$   &$2.72$   &$0.96$   &$1.16$   &$221.08$   &$175.54$   &$26.91$    \\
			& \textbf{FLOAM\_MCL}        &    &$136.08$  &$61.08$   &$64.49$   &$5.24$   &$1.35$   &$1.78$   &$156.81$    &$69.51$   &$14.74$ \\
			& \textbf{KISS\_MCL}         &   &$74.88$  &$49.18$   &$51.27$   &$4.66$   &$1.76$   &$2.11$   &$\mathbf{58.58}^{\star}$    &$43.09$   &$7.19$  \\
			& \textbf{HDL\_localization} &   &$361.09$  &$24.95$   &$60.52$   &$8.36$   &$0.62$   &$1.48$   &$559.63$    &$68.94$   &$81.78$ \\
			& \textbf{SM\_MCL}           &$6709.49$   &$99.72$  &$33.00$   &$38.15$   &$2.07$   &$0.55$   &$0.66$   &$68.03$    &$\mathbf{36.09}^{\star}$   &${5.60}$  \\
			& \textbf{Range\_MCL}       &${12.67}$   &${100.11}$  &${73.31}$   &${74.76}$   &${6.07}$   &${2.40}$   &${2.83}$   &$135.73$   &$132.13$   &$\mathbf{4.59}^{\star}$\\
			& \textbf{ERPoT(ours)}     &$\mathbf{0.28}^{\star}(2584/34708)$   &$\mathbf{47.96}^{\star}$  &$\mathbf{13.14}^{\star}$   &$\mathbf{16.33}^{\star}$   &$\mathbf{0.79}^{\star}$   &$\mathbf{0.38}^{\star}$   &$\mathbf{0.43}^{\star}$   &$100.65$   &$47.44$   &$8.61$\\
			\hline
			\multirow{7}{*}{\textbf{KITTI00-2}} 
			& \textbf{ALOAM\_MCL}        &\multirow{4}{*}{$56.44$}   &$158.69$  &$40.92$  &$47.96$   &$5.76$   &$1.24$   &$1.55$   &$261.92$   &$179.10$   &$56.19$  \\
			& \textbf{FLOAM\_MCL}        &   &$138.17$  &$49.73$   &$58.74$   &$4.34$   &$1.19$   &$1.47$   &$170.37$    &$76.94$   &$12.61$\\
			& \textbf{KISS\_MCL}         &   &$208.34$  &$62.20$   &$81.73$   &$5.35$   &$1.22$   &$1.53$   &$84.26$    &$42.58$   &$8.92$\\
			& \textbf{HDL\_localization} &   &$149.60$  &$\mathbf{7.72}^{\star}$   &$11.32$   &$5.08$   &$0.49$   &$0.67$   &$334.79$    &$61.64$   &$46.85$ \\
			& \textbf{SM\_MCL}           &$6709.49$   &$86.20$  &$21.69$   &$25.17$   &$1.17$   &$0.32$   &$0.38$   &$\mathbf{50.33}^{\star}$    &$\mathbf{36.76}^{\star}$   &$\mathbf{4.75}^{\star}$ \\
			& \textbf{Range\_MCL}        &${12.67}$    &$-$  &$-$  &$-$   &$-$   &$-$   &$-$   &$-$   &$-$    &$-$  \\
			& \textbf{ERPoT(ours)}     &$\mathbf{0.28}^{\star}(2584/34708)$    &$\mathbf{23.66}^{\star}$  &$8.36$   &$\mathbf{9.53}^{\star}$   &$\mathbf{0.78}^{\star}$   &$\mathbf{0.21}^{\star}$   &$\mathbf{0.26}^{\star}$   &$61.55$    &$40.81$   &$5.67$   \\
			\hline
			\multirow{7}{*}{\textbf{KITTI00-3}} 
			& \textbf{ALOAM\_MCL}        &\multirow{4}{*}{$56.44$}  &$186.50$  &$114.40$  &$125.01$   &$4.78$   &$1.38$   &$1.82$   &$225.06$   &$175.33$    &$46.56$    \\
			& \textbf{FLOAM\_MCL}        &   &$170.32$  &$113.97$   &$123.72$   &$5.32$   &$1.44$   &$1.84$   &$199.90$    &$80.45$   &$18.15$  \\
			& \textbf{KISS\_MCL}         &   &$265.02$  &$131.44$   &$152.89$   &$8.28$   &$1.82$   &$2.47$   &$65.65$    &$44.06$   &$9.13$  \\
			& \textbf{HDL\_localization} &   &$48.52$  &$\mathbf{8.83}^{\star}$   &$\mathbf{10.79}^{\star}$   &$3.72$   &$0.31$   &$0.74$   &$444.92$    &$77.79$   &$75.35$     \\
			& \textbf{SM\_MCL}           &$6709.49$   &$51.84$  &$19.46$   &$22.20$   &$0.79$   &$0.29$   &$0.33$   &$\mathbf{59.71}^{\star}$    &$\mathbf{38.39}^{\star}$   &$5.68$      \\
			& \textbf{Range\_MCL}       &${12.67}$   &${152.38}$  &${112.33}$   &${118.33}$   &${3.66}$   &${1.09}$   &${1.39}$   &$135.92$   &$132.38$   &$\mathbf{1.72}^{\star}$\\
			& \textbf{ERPoT(ours)}     &$\mathbf{0.28}^{\star}(2584/34708)$    &$\mathbf{34.46}^{\star}$  &$12.88$   &$16.64$   &$\mathbf{0.41}^{\star}$   &$\mathbf{0.12}^{\star}$   &$\mathbf{0.15}^{\star}$   &$61.71$    &$45.28$   &$6.64$     \\
			\hline \hline
			\multirow{7}{*}{\textbf{KITTI02-1}} 
			& \textbf{ALOAM\_MCL}        &\multirow{4}{*}{$110.47$}   &$130.04$  &$74.96$  &$76.71$   &$5.28$   &$1.33$   &$1.63$   &$246.59$   &$180.41$    &$44.14$   \\
			& \textbf{FLOAM\_MCL}        &   &$130.24$  &$69.36$   &$73.32$   &$4.68$   &$1.19$   &$1.55$   &$202.30$    &$81.64$   &$19.11$  \\
			& \textbf{KISS\_MCL}         &   &$170.96$  &$84.73$   &$90.04$   &$5.02$   &$1.59$   &$1.95$   &$\mathbf{60.13}^{\star}$    &$\mathbf{43.01}^{\star}$   &$7.13$  \\
			& \textbf{HDL\_localization} &  &$-$  &$-$  &$-$   &$-$   &$-$   &$-$   &$-$   &$-$    &$-$  \\
			& \textbf{SM\_MCL}           &$13838.15$   &$75.97$  &$19.10$   &$22.83$   &$0.69$   &$0.31$   &$0.35$   &$69.77$    &$36.27$   &$9.26$   \\
			& \textbf{Range\_MCL}        &${27.38}$   &$-$  &$-$  &$-$   &$-$   &$-$   &$-$   &$-$   &$-$    &$-$  \\
			& \textbf{ERPoT(ours)}     &$\mathbf{0.85}^{\star}(7582/106065)$    &$\mathbf{33.13}^{\star}$  &$\mathbf{11.58}^{\star}$   &$\mathbf{13.39}^{\star}$   &$\mathbf{0.64}^{\star}$   &$\mathbf{0.23}^{\star}$   &$\mathbf{0.26}^{\star}$   &$68.50$    &$43.83$   &$\mathbf{6.84}^{\star}$   \\
			\hline
			\multirow{7}{*}{\textbf{KITTI02-2}} 
			& \textbf{ALOAM\_MCL}        &\multirow{4}{*}{$110.47$}  &$-$  &$-$  &$-$   &$-$   &$-$   &$-$   &$-$   &$-$    &$-$   \\
			& \textbf{FLOAM\_MCL}        &   &$-$  &$-$  &$-$   &$-$   &$-$   &$-$   &$-$   &$-$    &$-$   \\
			& \textbf{KISS\_MCL}         &   &$181.80$  &$140.22$   &$142.50$   &$5.36$   &$1.97$   &$2.34$   &$76.80$    &$\mathbf{42.29}^{\star}$   &$11.26$   \\
			& \textbf{HDL\_localization} &   &$-$  &$-$  &$-$   &$-$   &$-$   &$-$   &$-$   &$-$    &$-$   \\
			& \textbf{SM\_MCL}           &$13838.15$  &$-$  &$-$  &$-$   &$-$   &$-$   &$-$   &$-$   &$-$    &$-$   \\
			& \textbf{Range\_MCL}        &${27.38}$   &$-$  &$-$  &$-$   &$-$   &$-$   &$-$   &$-$   &$-$    &$-$  \\
			& \textbf{ERPoT(ours)}     &$\mathbf{0.85}^{\star}(7582/106065)$    &$\mathbf{83.44}^{\star}$  &$\mathbf{18.70}^{\star}$   &$\mathbf{22.10}^{\star}$   &$\mathbf{0.85}^{\star}$   &$\mathbf{0.29}^{\star}$   &$\mathbf{0.35}^{\star}$   &$\mathbf{74.72}^{\star}$    &$48.40$   &$\mathbf{8.01}^{\star}$  \\
			\hline \hline
			\multirow{7}{*}{\textbf{KITTI05-1}} 
			& \textbf{ALOAM\_MCL}        &\multirow{4}{*}{$70.93$}	&$63.42$  &$32.76$  &$35.56$   &$3.60$   &$1.15$   &$1.41$   &$254.67$   &$177.73$    &$52.50$   \\
			& \textbf{FLOAM\_MCL}        &   &$127.64$  &$40.94$   &$50.25$   &$6.08$   &$1.25$   &$1.58$   &$189.57$   &$77.71$   &$14.50$   \\
			& \textbf{KISS\_MCL}         &   &$96.40$  &$48.07$   &$53.91$   &$3.96$   &$1.16$   &$1.43$   &$80.88$    &$48.65$   &$9.70$    \\
			& \textbf{HDL\_localization} &  &$149.28$  &$8.12$  &$19.02$   &$4.02$   &$0.32$   &$0.57$   &$429.17$   &$56.83$    &$58.34$   \\
			& \textbf{SM\_MCL}           &$6647.48$   &$44.22$  &$15.55$   &$17.78$   &$\mathbf{1.52}^{\star}$   &$0.35$   &$0.41$   &$\mathbf{55.58}^{\star}$    &$\mathbf{34.17}^{\star}$   &$5.05$    \\
			& \textbf{Range\_MCL}        &${23.37}$   &$-$  &$-$  &$-$   &$-$   &$-$   &$-$   &$-$   &$-$    &$-$  \\
			& \textbf{ERPoT(ours)}   	&$\mathbf{0.36}^{\star}(3321/44174)$   	&$\mathbf{25.46}^{\star}$  &$\mathbf{7.94}^{\star}$  &$\mathbf{9.01}^{\star}$   &$1.58$   &$\mathbf{0.24}^{\star}$   &$\mathbf{0.33}^{\star}$   &$58.91$   &$42.63$    &$\mathbf{4.71}^{\star}$   \\
			\hline
			\multirow{7}{*}{\textbf{KITTI05-2}} 
			& \textbf{ALOAM\_MCL}   		&\multirow{4}{*}{$70.93$}    &$101.97$  &$43.56$  &$50.42$  &$5.46$   &$1.28$   &$1.67$     &$269.04$   &$179.49$    &$60.88$   \\
			& \textbf{FLOAM\_MCL}        &    &$216.58$  &$65.23$  &$85.12$   &$7.99$   &$1.22$   &$1.66$   &$215.86$   &$83.66$    &$14.59$  \\
			& \textbf{KISS\_MCL}         &    &$117.28$  &$51.24$   &$60.81$   &$4.52$   &$1.21$   &$1.57$   &$80.50$    &$50.80$   &$9.57$   \\
			& \textbf{HDL\_localization} &    &$103.83$  &$10.68$  &$13.14$   &$1.86$   &$0.33$   &$0.45$   &$512.56$   &$56.26$    &$46.64$    \\
			& \textbf{SM\_MCL}           &$6647.48$  &$40.92$  &$15.30$  &$17.87$   &$0.95$   &$0.43$   &$0.48$   &$81.48$   &$43.76$    &$8.43$    \\
			& \textbf{Range\_MCL}        &${23.37}$   &${176.95}$  &${83.49}$   &${91.05}$   &${5.80}$   &${1.32}$   &${1.66}$   &$235.29$   &$151.41$   &$88.27$\\
			& \textbf{ERPoT(ours)}     &$\mathbf{0.36}^{\star}(3321/44174)$    &$\mathbf{26.84}^{\star}$  &$\mathbf{10.14}^{\star}$   &$\mathbf{11.14}^{\star}$   &$\mathbf{0.60}^{\star}$   &$\mathbf{0.26}^{\star}$   &$\mathbf{0.29}^{\star}$   &$\mathbf{68.72}^{\star}$    &$\mathbf{41.59}^{\star}$   &$\mathbf{5.97}^{\star}$   \\
			\hline \hline
			\multirow{7}{*}{\textbf{KITTI06-1}} 
			& \textbf{ALOAM\_MCL}        &\multirow{4}{*}{$56.90$}  &$-$  &$-$  &$-$   &$-$   &$-$   &$-$   &$-$   &$-$    &$-$   \\
			& \textbf{FLOAM\_MCL}        &   &$547.68$  &$243.60$   &$276.71$   &$9.21$   &$2.16$   &$2.80$   &$278.36$  &$95.46$   &$18.56$   \\
			& \textbf{KISS\_MCL}         &   &$-$  &$-$  &$-$   &$-$   &$-$   &$-$   &$-$   &$-$    &$-$   \\
			& \textbf{HDL\_localization} &   &$-$  &$-$  &$-$   &$-$   &$-$   &$-$   &$-$   &$-$    &$-$   \\
			& \textbf{SM\_MCL}           &$5791.79$   &$43.86$  &$17.81$   &$19.64$   &$0.90$   &$0.38$   &$0.42$   &$102.60$    &$\mathbf{45.58}^{\star}$   &$7.35$    \\
			& \textbf{Range\_MCL}        &${10.80}$   &${153.86}$  &${110.38}$   &${112.82}$   &${2.89}$   &${0.75}$   &${0.95}$   &$221.04$   &$175.73$    &$23.96$\\
			& \textbf{ERPoT(ours)}     &$\mathbf{0.27}^{\star}(2644/33504)$   	&$\mathbf{28.78}^{\star}$  &$\mathbf{7.25}^{\star}$  &$\mathbf{8.08}^{\star}$   &$\mathbf{0.51}^{\star}$   &$\mathbf{0.23}^{\star}$   &$\mathbf{0.24}^{\star}$   &$\mathbf{63.79}^{\star}$   &$46.55$    &$\mathbf{5.63}^{\star}$   \\
			\hline \hline
			\multirow{7}{*}{\textbf{KITTI08-1}} 
			& \textbf{ALOAM\_MCL}   		&\multirow{4}{*}{$133.01$}  &$-$  &$-$  &$-$   &$-$   &$-$   &$-$   &$-$   &$-$    &$-$   \\
			& \textbf{FLOAM\_MCL}        &  &$-$  &$-$  &$-$   &$-$   &$-$   &$-$   &$-$   &$-$    &$-$   \\
			& \textbf{KISS\_MCL}         &  &$-$  &$-$  &$-$   &$-$   &$-$   &$-$   &$-$   &$-$    &$-$   \\
			& \textbf{HDL\_localization} &  &$-$  &$-$  &$-$   &$-$   &$-$   &$-$   &$-$   &$-$    &$-$   \\
			& \textbf{SM\_MCL}           &$11061.94$  &$\mathbf{61.26}^{\star}$  &$24.60$  &$28.24$   &$1.24$   &$0.48$   &$0.53$   &$\mathbf{62.89}^{\star}$   &$\mathbf{42.00}^{\star}$    &$\mathbf{5.23}^{\star}$    \\
			& \textbf{Range\_MCL}        &${25.14}$  &$-$  &$-$  &$-$   &$-$   &$-$   &$-$   &$-$   &$-$    &$-$  \\
			& \textbf{ERPoT(ours)}     &$\mathbf{0.59}^{\star}(5451/71972)$   &$78.01$  &$\mathbf{16.10}^{\star}$   &$\mathbf{22.47}^{\star}$   &$\mathbf{0.96}^{\star}$   &$\mathbf{0.34}^{\star}$   &$\mathbf{0.40}^{\star}$   &$67.14$   &$45.39$   &$5.33$   \\
			\hline
	\end{tabular}}}
	\label{table_4}
\end{table*}

Furthermore, pose tracking is conducted using the prior map, and the effect for KITTI00 is shown in Fig. \ref{Fig_16}(c). More importantly, Tab. \ref{table_4} shows the quantitative comparative results on KITTI dataset, with optimal results in bold and failed cases indicated by `$-$'. During the pose tracking experiments for each test sequence, success is confirmed when the final pose error remains within the specified limits: translational error within $200 \mathrm{cm}$ and rotational error within $20^\circ$. Additionally, the pose error during the process must be less than $500\mathrm{cm}$ and $30^\circ$. This criterion ensures that only the sequences meeting these standards are subjected to further evaluation.

As shown in Tab. \ref{table_4}, ERPoT successfully achieved pose tracking on nine test sequence, which uses a novel polygon-based prior map for compact and lightweight storage. Across all five test environments, the map size is consistently under $1$ MB, demonstrating the efficiency and practicality of our method, and the number of polygons/vertices is displayed after the map size. Furthermore, regarding translational and rotational errors, the proposed approach has also demonstrated optimal performance. The mean values for translational and rotational errors are both below $20\mathrm{cm}$ and $0.5^\circ$, respectively, across all test scenarios, underscoring the accuracy and precision of our method. In addition, ERPoT demonstrates exceptional runtime performance, reliably achieving a consistent processing time of approximately $40\mathrm{ms}$ with high stability.

\begin{figure}[!htb]
	\centering
	{\includegraphics[width=4cm]{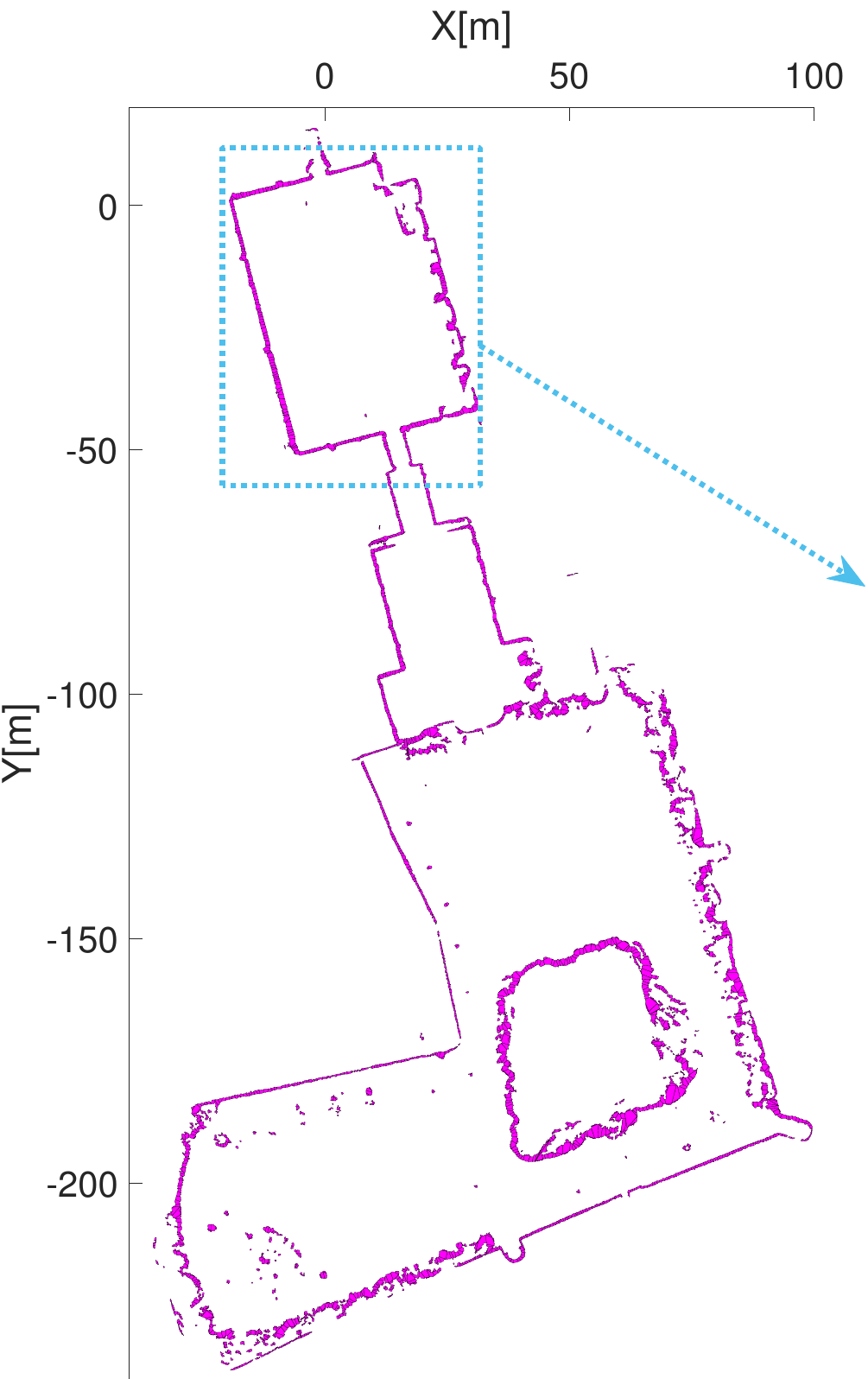}}
	{\includegraphics[width=3.5cm]{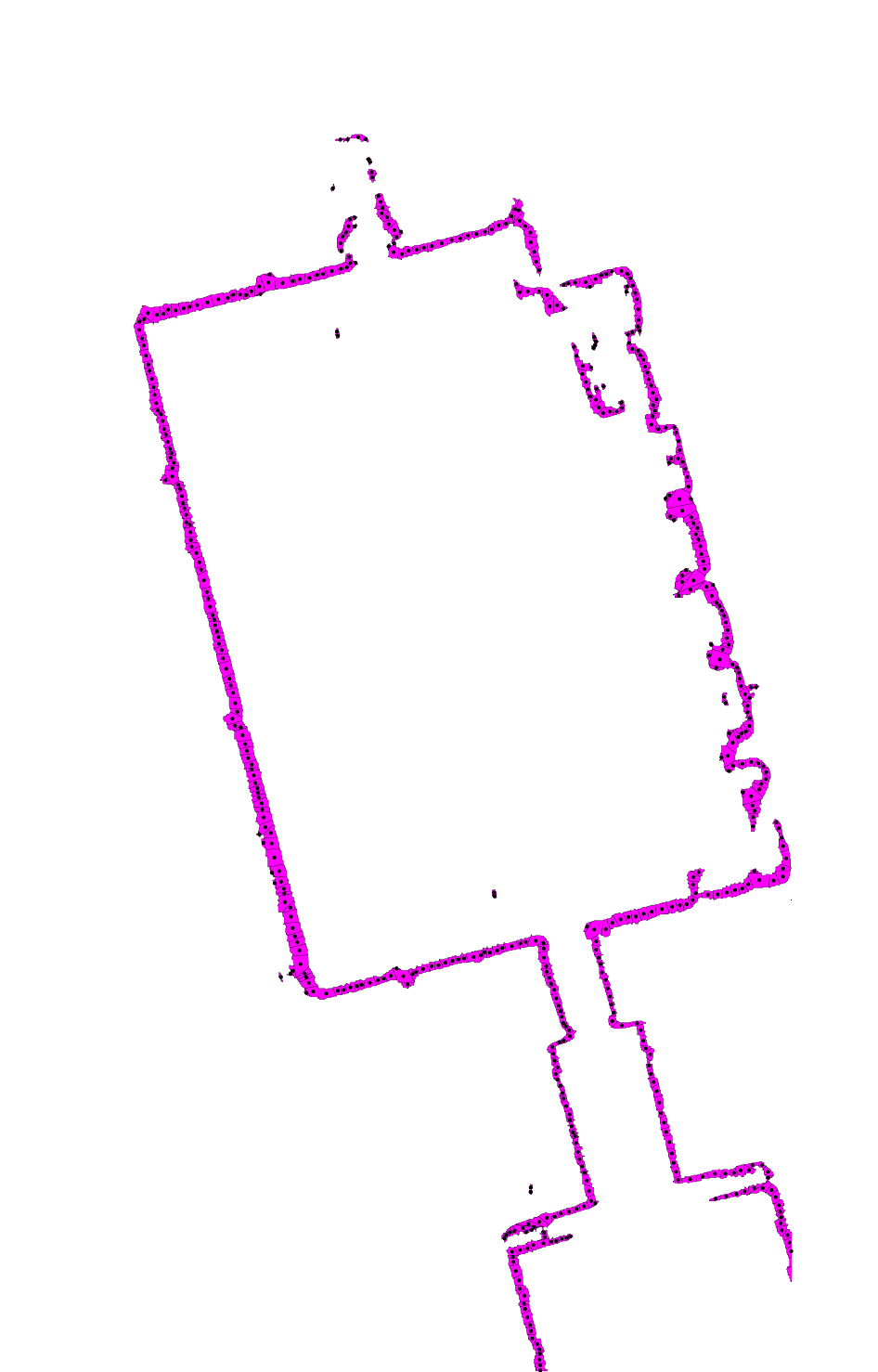}}\\
	\caption{Polygon map for the Newer College dataset.}
	\label{Fig_18}
\end{figure}

\begin{table*}[!htb]\small
	\renewcommand\arraystretch{0.7}
	\centering
	\caption{Experimental results of different approaches using the Newer College dataset}
	\setlength{\tabcolsep}{1mm}{
		{\fontsize{6pt}{6pt}
		\begin{tabular}{|l|c|c|c|c|c|c|c|c|c|c|c|}
			\hline
			\multirow{2}{*}{\textbf{Dataset No.}} &\textbf{Approach} &\textbf{Prior map} &\multicolumn{3}{c|}{\textbf{Translational error (cm)}} &\multicolumn{3}{c|}{\textbf{Rotational error (deg)}} &\multicolumn{3}{c|}{\textbf{Runtime (ms)}} \\
			\cline{4-12}
			&\textbf{name}   &\textbf{size (MB) $\downarrow$} &\textbf{Max} $\downarrow$  &\textbf{Mean} $\downarrow$  &\textbf{RMSE} $\downarrow$  &\textbf{Max} $\downarrow$ &\textbf{Mean} $\downarrow$ &\textbf{RMSE} $\downarrow$ &\textbf{Max} $\downarrow$ &\textbf{Mean} $\downarrow$ &\textbf{SD}$\downarrow$ \\
			\hline \hline
			\multirow{7}{*}{\textbf{Newer College-1}} 
			& \textbf{ALOAM\_MCL}        &\multirow{4}{*}{${62.49}$}  &$\mathbf{122.38}^{\star}$  &$36.10$  &$42.90$   &$11.31$   &$1.98$   &$2.48$   &$639.64$   &$182.60$    &$143.01$    \\
			& \textbf{FLOAM\_MCL}        &  &$-$  &$-$  &$-$   &$-$   &$-$   &$-$   &$-$   &$-$    &$-$   \\
			& \textbf{KISS\_MCL}         &  &$145.12$  &$41.44$   &$49.06$   &$\mathbf{9.24}^{\star}$   &${1.93}$   &$2.44$   &$237.92$   &$74.90$   &$23.81$  \\
			& \textbf{HDL\_localization} &  &$-$  &$-$  &$-$   &$-$   &$-$   &$-$   &$-$   &$-$    &$-$   \\
			& \textbf{SM\_MCL}           &$4397.10$  &$-$  &$-$  &$-$   &$-$   &$-$   &$-$   &$-$   &$-$    &$-$   \\
			& \textbf{Range\_MCL}        &${23.28}$   &${124.63}$  &${30.83}$   &${36.63}$   &${14.75}$   &${1.60}$   &${2.05}$   &$585.20$   &$228.98$   &$170.01$\\
			& \textbf{ERPoT(ours)}     &$\mathbf{0.32}^{\star}(2749/40341)$   &${136.00}$  &$\mathbf{18.08}^{\star}$   &$\mathbf{21.24}^{\star}$   &${19.03}$   &$\mathbf{0.88}^{\star}$   &$\mathbf{1.23}^{\star}$   &$\mathbf{66.18}^{\star}$   &$\mathbf{29.99}^{\star}$   &$\mathbf{8.68}^{\star}$  \\
			\hline
			\multirow{7}{*}{\textbf{Newer College-2}} 
			& \textbf{ALOAM\_MCL}        &\multirow{4}{*}{${62.49}$}  &$-$  &$-$  &$-$   &$-$   &$-$   &$-$   &$-$   &$-$    &$-$   \\
			& \textbf{FLOAM\_MCL}        &  &$-$  &$-$  &$-$   &$-$   &$-$   &$-$   &$-$   &$-$    &$-$   \\
			& \textbf{KISS\_MCL}         &  &$248.69$  &$65.20$   &$78.99$   &$15.17$   &$2.78$   &$3.61$   &$499.19$   &$110.53$   &$50.72$\\
			& \textbf{HDL\_localization} &  &$-$  &$-$  &$-$   &$-$   &$-$   &$-$   &$-$   &$-$    &$-$   \\
			& \textbf{SM\_MCL}           &$4397.10$  &$-$  &$-$  &$-$   &$-$   &$-$   &$-$   &$-$   &$-$    &$-$   \\
			& \textbf{Range\_MCL}        &${23.28}$   &${254.92}$  &${45.79}$   &${62.44}$   &${12.80}$   &${2.01}$   &${2.60}$   &$585.59$   &$422.85$   &$158.01$\\
			& \textbf{ERPoT(ours)}     &$\mathbf{0.32}^{\star}(2749/40341)$   &$\mathbf{210.54}^{\star}$  &$\mathbf{24.92}^{\star}$   &$\mathbf{29.47}^{\star}$   &$\mathbf{8.05}^{\star}$   &$\mathbf{1.03}^{\star}$   &$\mathbf{0.34}^{\star}$   &$\mathbf{59.30}^{\star}$    &$\mathbf{27.73}^{\star}$   &$\mathbf{7.07}^{\star}$   \\
			\hline
	\end{tabular}}}
	\label{table_5}
\end{table*}

On the other hand, comparative experiments on the Newer College dataset are conducted, with the polygon map shown in Fig. \ref{Fig_18}. The campus buildings are represented by compact polygons to create a lightweight prior map, as detailed in the right subfigure. Two test sequences (Newer College-1 and Newer College-2) are evaluated for pose tracking, with quantitative results presented in Tab. \ref{table_5}.

The Newer College dataset, collected via a handheld platform, contains noisy LiDAR data due to movement. Combined with extensive vegetation, these factors create significant challenges for pure LiDAR pose tracking. As a result, methods like {FLOAM\_MCL}, {HDL\_localization}, and {SM\_MCL} fail to achieve effective tracking in both test sequences. However, our proposed ERPoT maintains successful pose tracking capabilities, even though it exhibits slightly higher translational and rotational errors. Moreover, the proposed ERPoT also demonstrates the superior performance in terms of runtime.

\begin{figure}[!htb]
	\centering
	{\includegraphics[width=2.8cm]{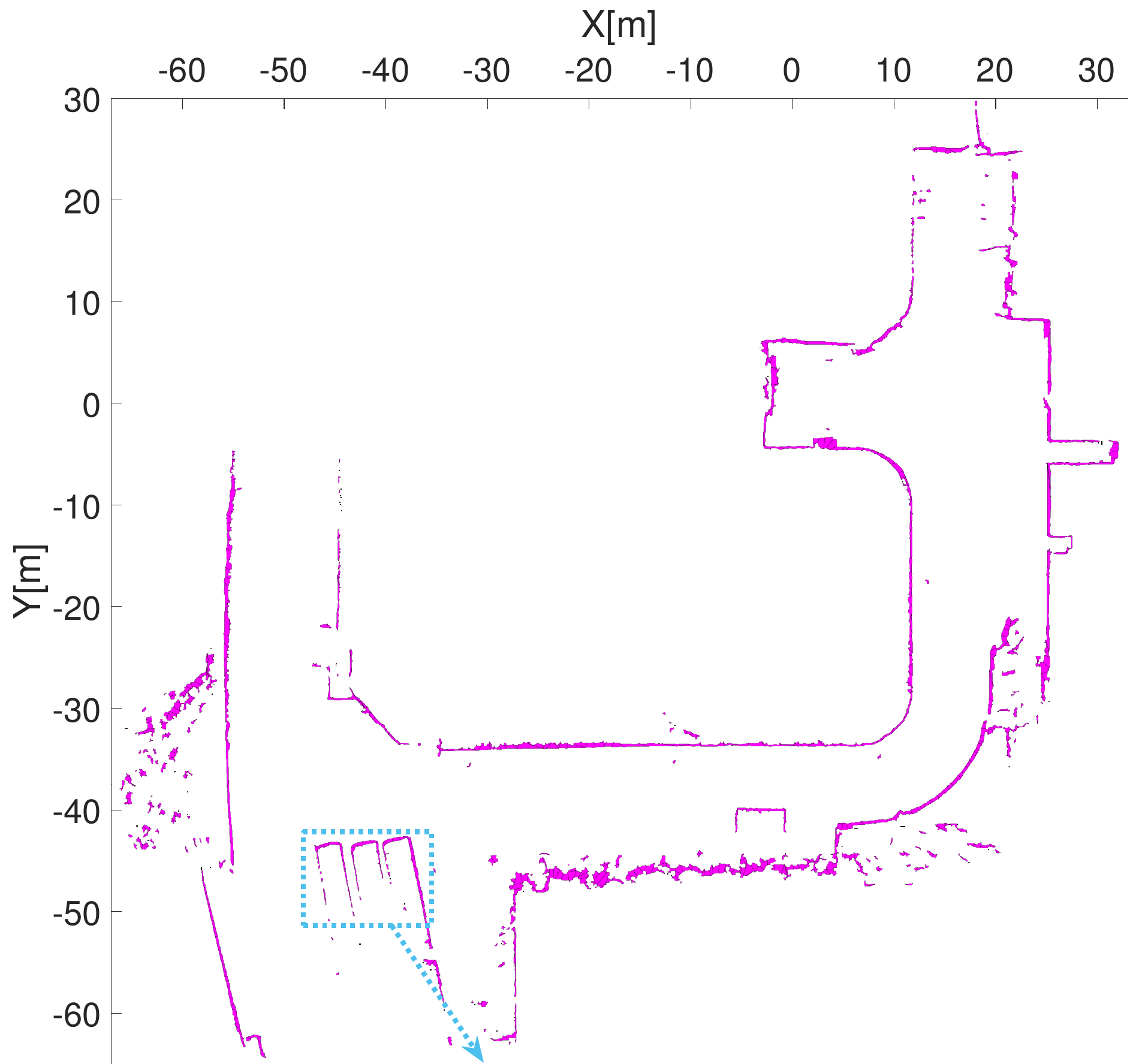}}
	{\includegraphics[width=2.92cm]{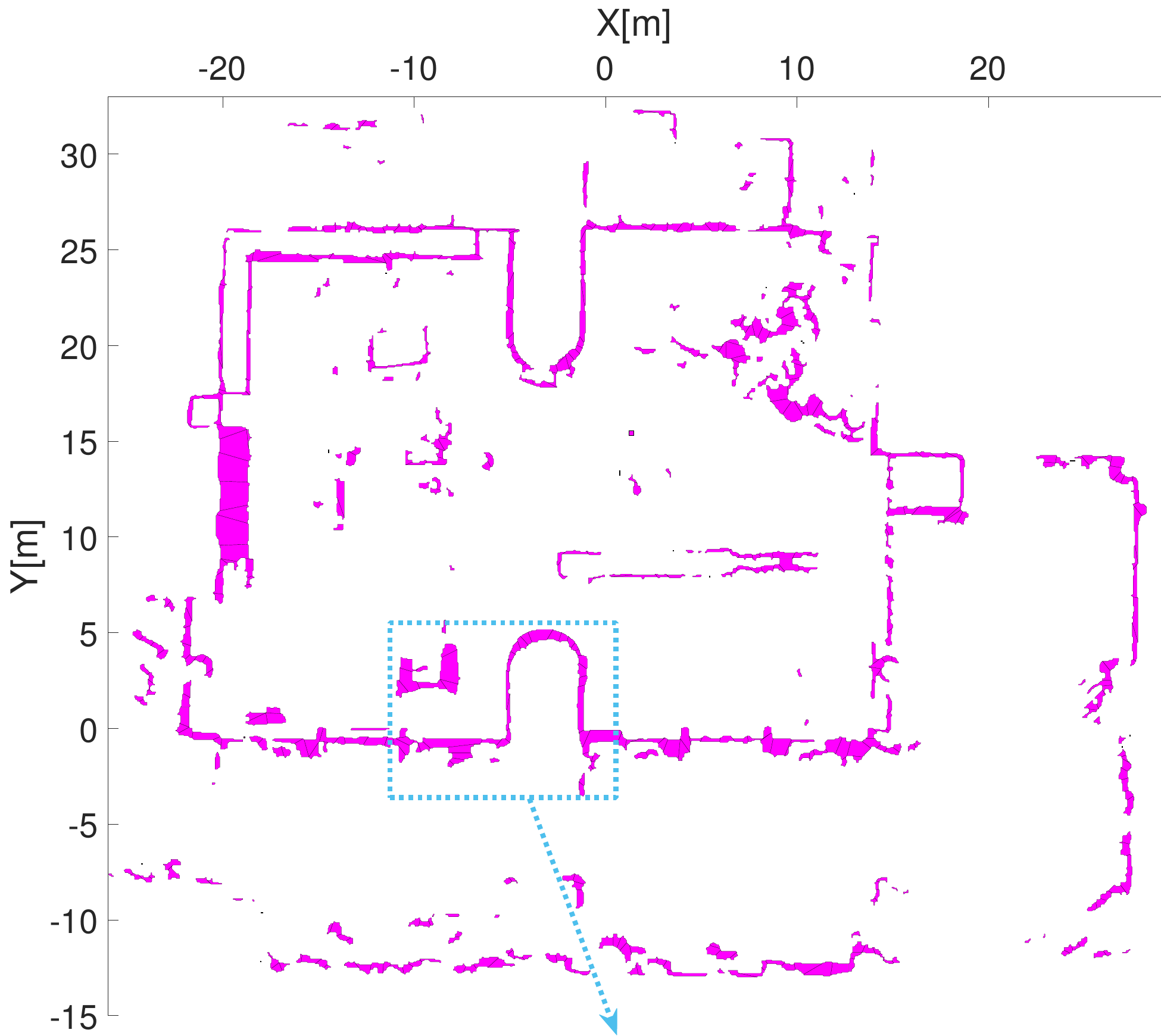}}
	{\includegraphics[width=2.92cm]{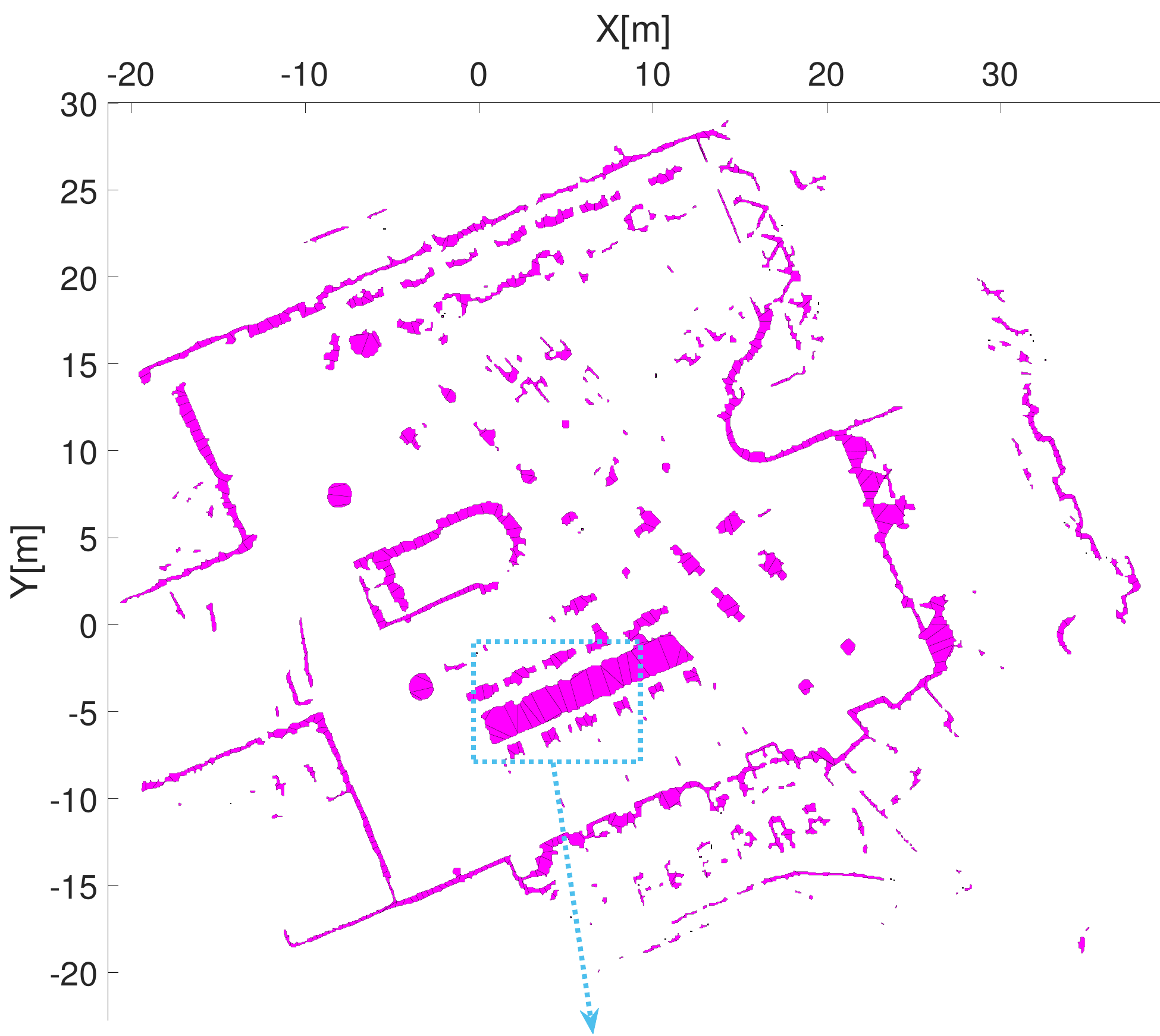}}\\
	\centering
	\subfigure[Portable-campus]{\includegraphics[width=2.85cm]{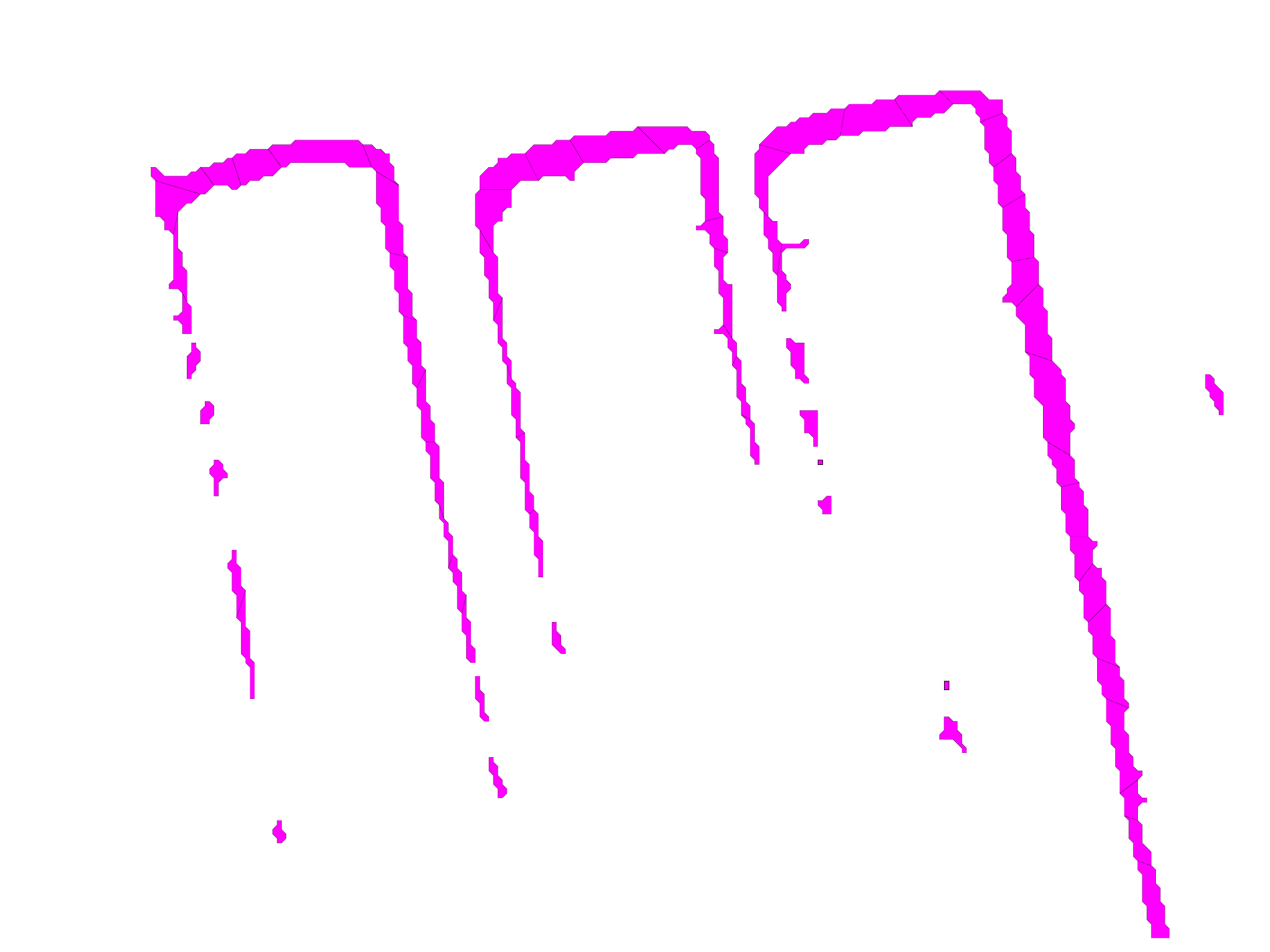}}
	\subfigure[Portable-garden]{\includegraphics[width=2.85cm]{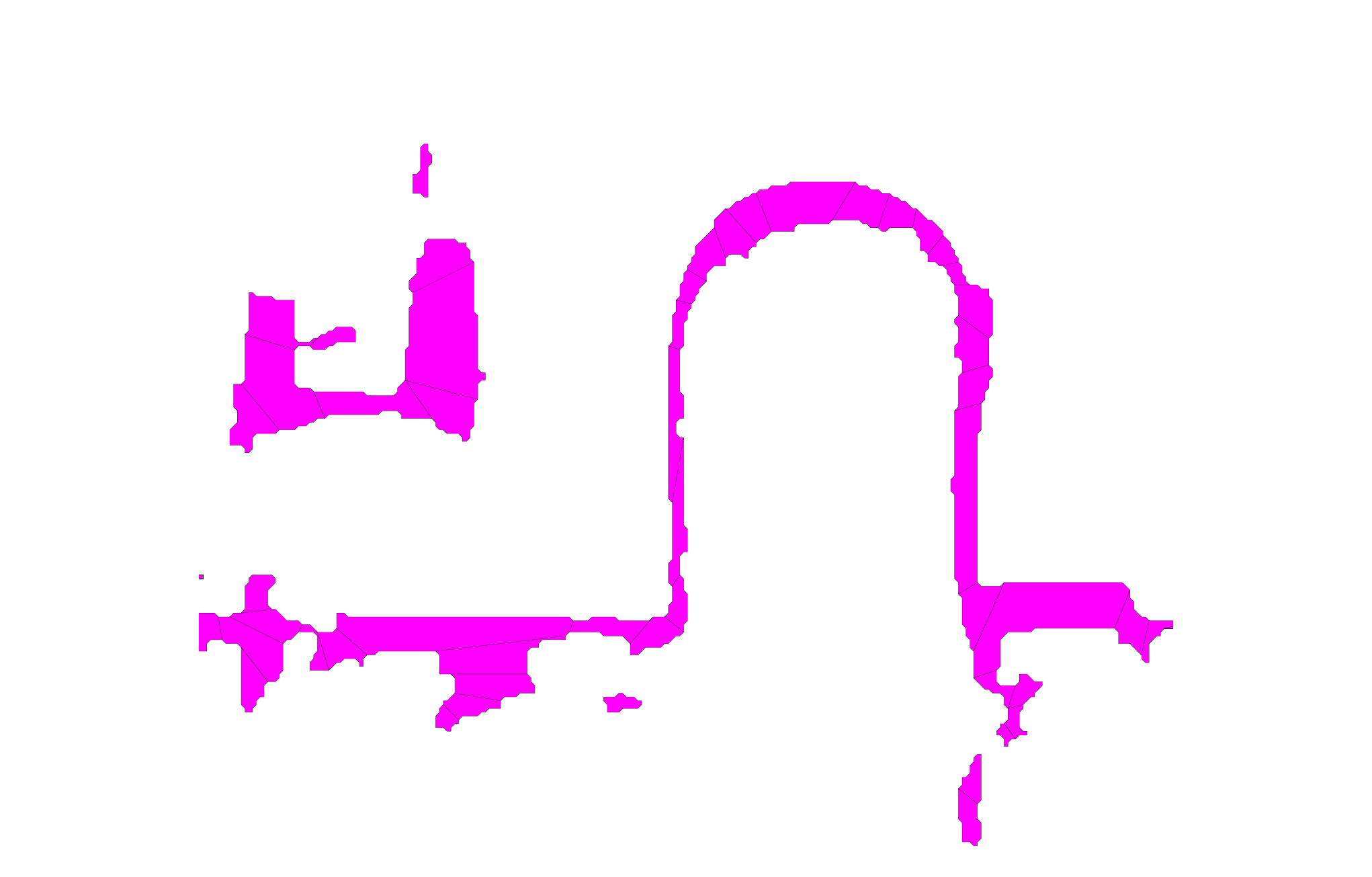}}
	\subfigure[Portable-canteen]{\includegraphics[width=2.85cm]{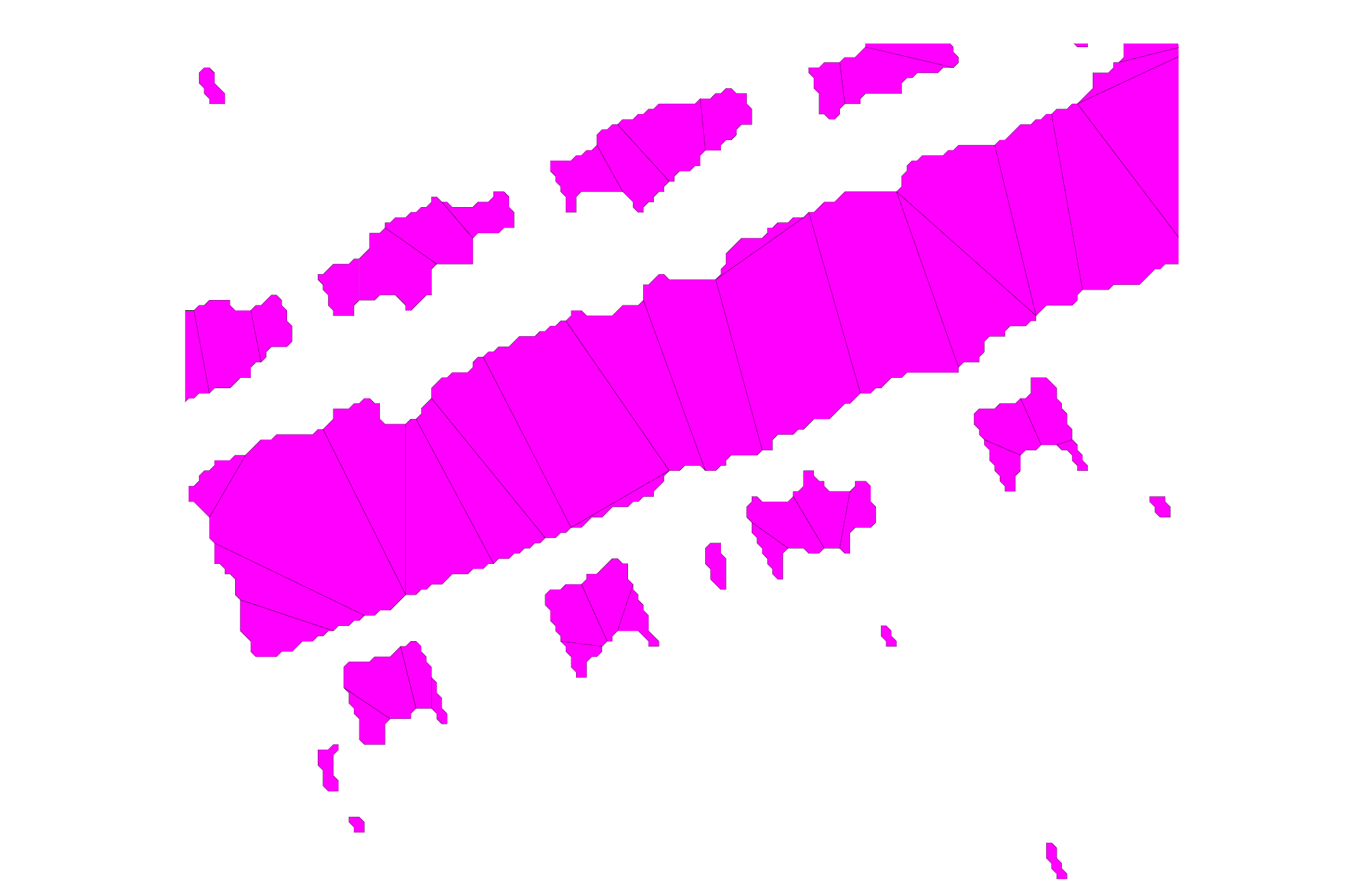}}\\
	\subfigure[Portable-building]{\includegraphics[width=8.5cm]{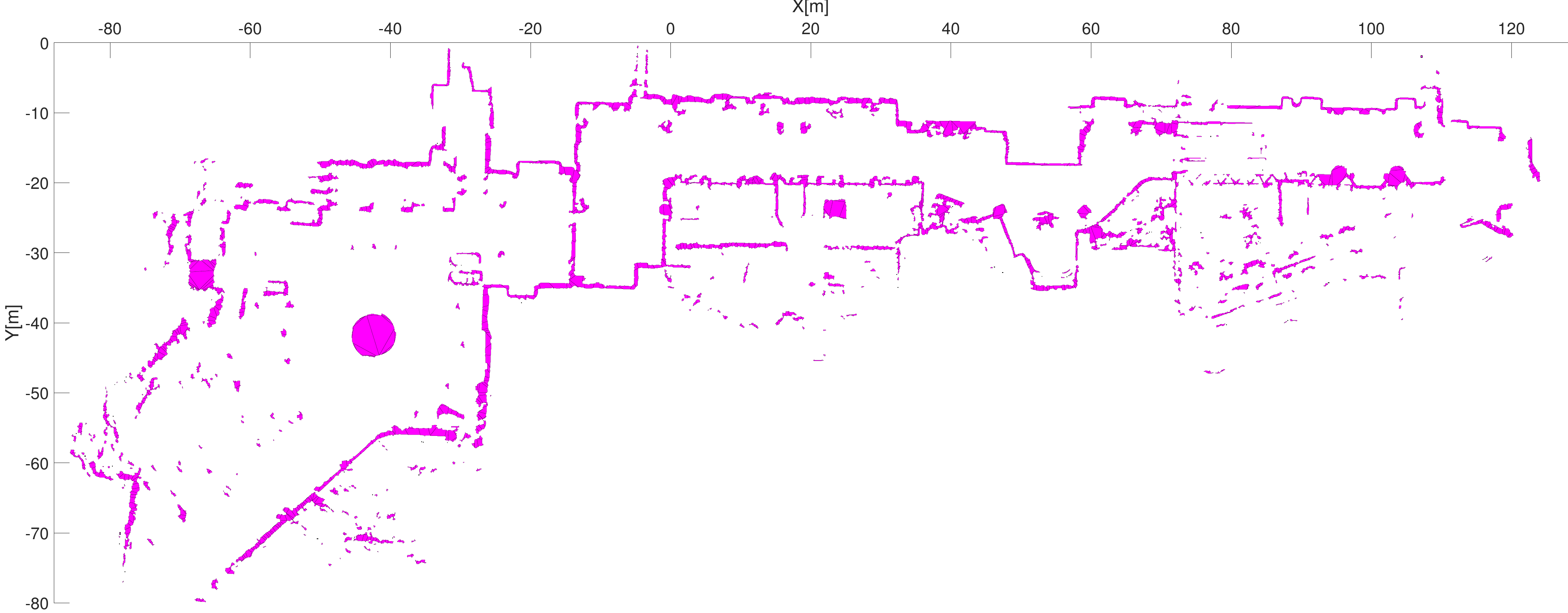}}
	\caption{Polygon maps for the FusionPortable dataset.}
	\label{Fig_19}
\end{figure}

To further validate the effectiveness and reliability of ERPoT, additional comparative experiments are conducted using the Portable-campus and Portable-building datasets. The polygon maps for these scenarios are shown in Fig. \ref{Fig_19}(a) and Fig. \ref{Fig_19}(d), with results detailed in Tab. \ref{table_6}. Notably, the prior map size ranges from $0.1$ to $0.2$ MB, demonstrating remarkable performance. ERPoT also shows superior performance in translational error, rotational error, and runtime.

\begin{table*}[!htb]\small
	\renewcommand\arraystretch{0.7}
	\centering
	\caption{Experimental results of different approaches using the FusionPortable dataset(outdoor)}
	\setlength{\tabcolsep}{1mm}{
		{\fontsize{6pt}{6pt}
					\begin{tabular}{|l|c|c|c|c|c|c|c|c|c|c|c|}
						\hline
						\multirow{2}{*}{\textbf{Dataset No.}} &\textbf{Approach} &\textbf{Prior map} &\multicolumn{3}{c|}{\textbf{Translational error (cm)}} &\multicolumn{3}{c|}{\textbf{Rotational error (deg)}} &\multicolumn{3}{c|}{\textbf{Runtime (ms)}} \\
						\cline{4-12}
						&\textbf{name}   &\textbf{size (MB) $\downarrow$} &\textbf{Max} $\downarrow$  &\textbf{Mean} $\downarrow$  &\textbf{RMSE} $\downarrow$  &\textbf{Max} $\downarrow$ &\textbf{Mean} $\downarrow$ &\textbf{RMSE} $\downarrow$ &\textbf{Max} $\downarrow$ &\textbf{Mean} $\downarrow$ &\textbf{SD}$\downarrow$ \\
						\hline \hline
						\multirow{7}{*}{\textbf{Portable-campus}} 
						& \textbf{ALOAM\_MCL}        &\multirow{4}{*}{${24.77}$}  &$206.92$  &$45.65$  &$57.48$   &$11.51$   &$1.62$   &$2.15$   &$590.09$   &$266.80$    &$134.53$   \\
						& \textbf{FLOAM\_MCL}        &  &$197.48$  &$65.63$  &$75.57$   &$16.05$   &$2.18$   &$3.14$   &$292.33$   &$144.75$    &$82.73$   \\
						& \textbf{KISS\_MCL}         &  &$132.39$  &$61.39$   &$70.56$   &$7.16$   &$1.40$   &$1.78$   &$390.19$   &$116.33$   &$49.14$\\
						& \textbf{HDL\_localization} &  &$249.66$  &$25.84$  &$31.02$   &$12.13$   &$1.12$   &$1.77$   &$564.45$   &$127.97$    &$126.70$   \\
						& \textbf{SM\_MCL}           &$2369.31$  &$256.00$  &$35.08$  &$44.02$   &$\mathbf{5.85}^{\star}$   &$\mathbf{0.96}^{\star}$   &$1.48$   &$\mathbf{71.43}^{\star}$   &$\mathbf{32.71}^{\star}$    &$\mathbf{5.48}^{\star}$   \\
						& \textbf{Range\_MCL}        &${10.68}$   &${80.66}$  &${34.82}$   &${38.30}$   &${19.59}$   &${5.57}$   &${3.59}$   &$441.12$   &$239.27$   &$151.00$\\
						& \textbf{ERPoT(ours)}     &$\mathbf{0.10}^{\star}(960/12655)$   &$\mathbf{42.38}^{\star}$  &$\mathbf{17.08}^{\star}$   &$\mathbf{19.30}^{\star}$   &${6.60}$   &${1.11}$   &$\mathbf{1.47}^{\star}$   &${101.89}$    &${58.76}$   &${10.51}$   \\
						\hline \hline
						\multirow{7}{*}{\textbf{Portable-building-1}} 
						& \textbf{ALOAM\_MCL}        &\multirow{4}{*}{${42.33}$}  &$211.11$  &$55.91$  &$80.42$   &$17.46$   &$2.54$   &$3.56$   &$564.36$   &$257.12$    &$129.66$   \\
						& \textbf{FLOAM\_MCL}        &  &$82.48$  &$26.82$  &$32.35$   &$6.91$   &$1.56$   &$2.00$   &$394.81$   &$123.75$    &$25.10$   \\
						& \textbf{KISS\_MCL}         &  &$100.37$  &$31.07$   &$39.66$   &${6.88}$   &${1.74}$   &$2.23$   &$185.86$   &$63.45$   &$29.50$  \\
						& \textbf{HDL\_localization} &  &$-$  &$-$  &$-$   &$-$   &$-$   &$-$   &$-$   &$-$    &$-$   \\
						& \textbf{SM\_MCL}           &$3335.99$  &$147.95$  &$34.63$  &$42.14$   &$7.37$   &$1.22$   &$1.70$   &$\mathbf{52.08}^{\star}$   &$\mathbf{28.30}^{\star}$    &$7.29$   \\
						& \textbf{Range\_MCL}        &${10.15}$   &${103.09}$  &${41.85}$   &${50.32}$   &${12.70}$   &${1.85}$   &${2.69}$   &$100.65$   &$47.44$   &$8.61$\\
						& \textbf{ERPoT(ours)}     &$\mathbf{0.23}^{\star}(2290/30488)$   &$\mathbf{49.78}^{\star}$  &$\mathbf{12.53}^{\star}$   &$\mathbf{15.10}^{\star}$   &$\mathbf{5.33}^{\star}$   &$\mathbf{0.46}^{\star}$   &$\mathbf{0.63}^{\star}$   &${81.03}$   &${56.81}$   &$\mathbf{5.33}^{\star}$  \\
						\hline
						\multirow{7}{*}{\textbf{Portable-building-2}} 
						& \textbf{ALOAM\_MCL}        &\multirow{4}{*}{${42.33}$}  &$-$  &$-$  &$-$   &$-$   &$-$   &$-$   &$-$   &$-$    &$-$    \\
						& \textbf{FLOAM\_MCL}        &  &$78.11$  &$25.02$  &$29.36$   &$10.58$   &$2.03$   &$2.63$   &$216.99$   &$115.48$    &$44.75$   \\
						& \textbf{KISS\_MCL}         &  &$78.40$  &$24.06$   &$28.42$   &${14.48}$   &${2.14}$   &$3.01$   &$194.57$   &$69.01$   &$32.71$  \\
						& \textbf{HDL\_localization} &  &$-$  &$-$  &$-$   &$-$   &$-$   &$-$   &$-$   &$-$    &$-$   \\
						& \textbf{SM\_MCL}           &$3335.99$  &$-$  &$-$  &$-$   &$-$   &$-$   &$-$   &$-$   &$-$    &$-$   \\
						& \textbf{Range\_MCL}        &${10.15}$   &${187.95}$  &${53.31}$   &${73.61}$   &${13.07}$   &${2.51}$   &${3.40}$   &$347.70$   &$259.41$   &$130.56$\\
						& \textbf{ERPoT(ours)}     &$\mathbf{0.23}^{\star}(2290/30488)$   &$\mathbf{38.15}^{\star}$  &$\mathbf{11.92}^{\star}$   &$\mathbf{13.06}^{\star}$   &$\mathbf{3.06}^{\star}$   &$\mathbf{0.61}^{\star}$   &$\mathbf{0.79}^{\star}$   &$\mathbf{95.26}^{\star}$   &$\mathbf{59.26}^{\star}$   &$\mathbf{9.23}^{\star}$  \\
						\hline
	\end{tabular}}}
	\label{table_6}
\end{table*}

\subsection{Comparative experiments for indoor environments}

\begin{table*}[!htb]\small
	\renewcommand\arraystretch{0.7}
	\centering
	\caption{Experimental results of different approaches using the FusionPortable dataset(indoor)}
			\setlength{\tabcolsep}{1mm}{
				{\fontsize{6pt}{6pt}
				\begin{tabular}{|l|c|c|c|c|c|c|c|c|c|c|c|}
						\hline
						\multirow{2}{*}{\textbf{Dataset No.}} &\textbf{Approach} &\textbf{Prior map} &\multicolumn{3}{c|}{\textbf{Translational error (cm)}} &\multicolumn{3}{c|}{\textbf{Rotational error (deg)}} &\multicolumn{3}{c|}{\textbf{Runtime (ms)}} \\
						\cline{4-12}
						&\textbf{name}   &\textbf{size (MB) $\downarrow$} &\textbf{Max} $\downarrow$  &\textbf{Mean} $\downarrow$  &\textbf{RMSE} $\downarrow$  &\textbf{Max} $\downarrow$ &\textbf{Mean} $\downarrow$ &\textbf{RMSE} $\downarrow$ &\textbf{Max} $\downarrow$ &\textbf{Mean} $\downarrow$ &\textbf{SD}$\downarrow$ \\
						\hline \hline
						\multirow{7}{*}{\textbf{Portable-garden}} 
						& \textbf{ALOAM\_MCL}        &\multirow{4}{*}{${10.53}$}  &$-$  &$-$  &$-$   &$-$   &$-$   &$-$   &$-$   &$-$    &$-$   \\
						& \textbf{FLOAM\_MCL}        &  &$99.01$  &$20.97$  &$23.64$   &$10.76$   &$1.52$   &$2.02$   &$203.81$   &$122.94$    &$10.76$   \\
						& \textbf{KISS\_MCL}         &  &$168.52$ &$18.85$  &$26.65$   &${16.03}$   &${1.26}$   &$1.74$   &$159.38$   &$65.91$   &$17.54$  \\
						& \textbf{HDL\_localization} &  &$-$  &$-$  &$-$   &$-$   &$-$   &$-$   &$-$   &$-$    &$-$   \\
						& \textbf{SM\_MCL}           &$644.31$  &$-$  &$-$  &$-$   &$-$   &$-$   &$-$   &$-$   &$-$    &$-$   \\
						& \textbf{Range\_MCL}        &${16.11}$   &$\mathbf{46.74}^{\star}$  &${20.13}$   &${21.85}$   &${8.32}$   &${1.38}$   &${1.80}$   &$485.29$   &$398.88$   &$ 166.04$ \\
						& \textbf{ERPoT(ours)}     &$\mathbf{0.06}^{\star}(627/8135)$   &${73.03}$  &$\mathbf{15.39}^{\star}$   &$\mathbf{17.72}^{\star}$   &$\mathbf{4.31}^{\star}$   &$\mathbf{0.66}^{\star}$   &$\mathbf{0.90}^{\star}$   &$\mathbf{103.98}^{\star}$   &$\mathbf{64.34}^{\star}$   &$\mathbf{9.45}^{\star}$  \\
						\hline \hline
						\multirow{7}{*}{\textbf{Portable-canteen}} 
						& \textbf{ALOAM\_MCL}        &\multirow{4}{*}{${13.72}$}  &$-$  &$-$  &$-$   &$-$   &$-$   &$-$   &$-$   &$-$    &$-$   \\
						& \textbf{FLOAM\_MCL}        &  &$81.12$  &$19.09$  &$23.10$   &$9.05$   &$1.49$   &$1.95$   &$322.74$   &$113.17$    &$17.34$   \\
						& \textbf{KISS\_MCL}         &  &$71.18$  &$18.98$   &$24.29$   &$5.78$   &$1.46$   &$1.82$   &$154.35$   &$\mathbf{55.63}^{\star}$   &$13.18$\\
						& \textbf{HDL\_localization} &  &$-$  &$-$  &$-$   &$-$   &$-$   &$-$   &$-$   &$-$    &$-$   \\
						& \textbf{SM\_MCL}           &$885.53$  &$-$  &$-$  &$-$   &$-$   &$-$   &$-$   &$-$   &$-$    &$-$   \\
						& \textbf{Range\_MCL}        &${18.46}$   &$\mathbf{37.39}^{\star}$  &${13.32}$   &${14.93}$   &${8.19}$   &${1.31}$   &${1.73}$   &$588.19$   &$486.95$   &$82.51$\\
						& \textbf{ERPoT(ours)}     &$\mathbf{0.11}^{\star}(1115/15195)$   &${43.05}$  &$\mathbf{9.17}^{\star}$   &$\mathbf{10.22}^{\star}$   &$\mathbf{5.02}^{\star}$   &$\mathbf{0.56}^{\star}$   &$\mathbf{0.79}^{\star}$   &$\mathbf{99.08}^{\star}$    &${62.45}$   &$\mathbf{11.16}^{\star}$   \\
						\hline
				\end{tabular}}}
	\label{table_7}
\end{table*}

To further validate the effectiveness of ERPoT in indoor environments, comparative experiments are conducted using the Portable-garden and Portable-canteen datasets. These two scenarios represent typical indoor garden and canteen environments, respectively, and include dynamic pedestrian information. Detailed information is presented in Fig. \ref{Fig_11}(c).

The prior polygon maps are generated from the nighttime LiDAR sequences, as shown in Fig. \ref{Fig_19}(b) and Fig. \ref{Fig_19}(c). Notably, the prior map size of ERPoT remains exceptionally compact, even under $0.1$MB in the garden environment. Comparative results are presented in Tab. \ref{table_7}, highlighting the challenges of pose tracking in complex indoor settings, which lead to the failure of some methods like ALOAM\_MCL, HDL\_localization, and SM\_MCL. Despite these challenges, ERPoT demonstrates excellent performance, particularly in runtime, achieving stable results with a 128-line LiDAR.

\subsection{Quantitative comparative results using self datasets}
\begin{figure}[!htb]
	\centering
	\subfigure[Self dataset-01]{\includegraphics[width=7.3cm]{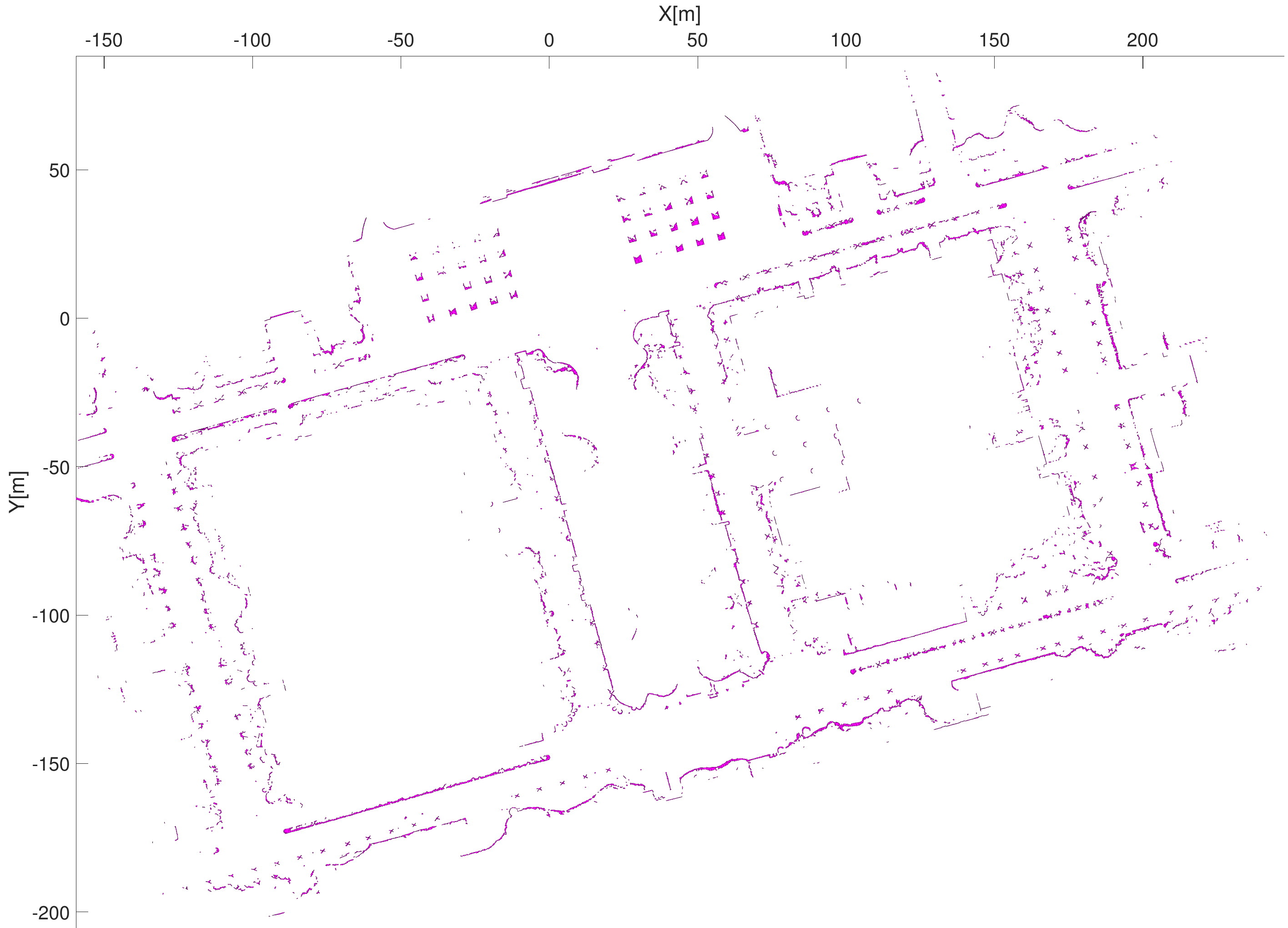}}
	\centering
	\subfigure[Self dataset-02]{
	\includegraphics[width=5cm]{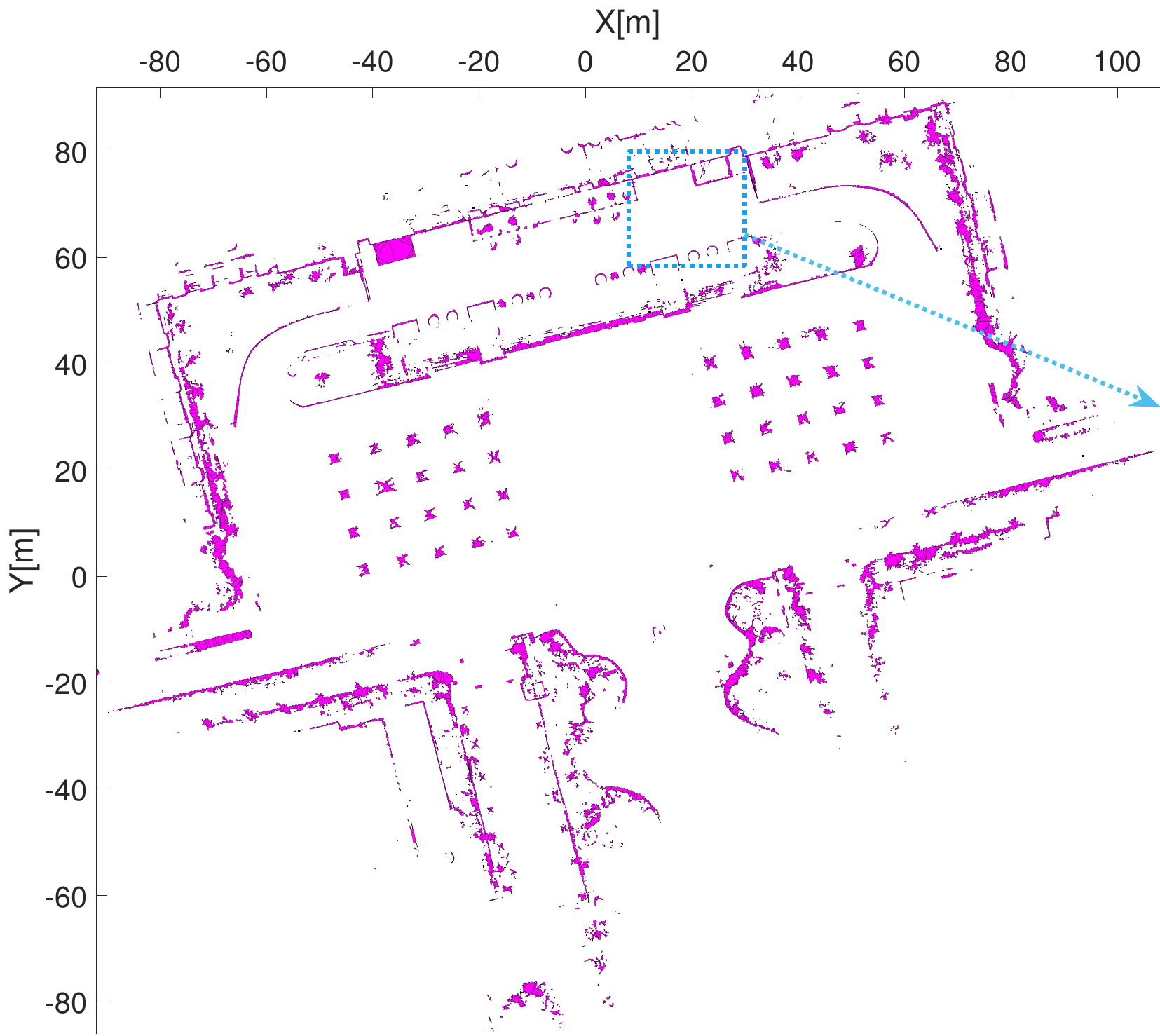}
	\includegraphics[width=3cm]{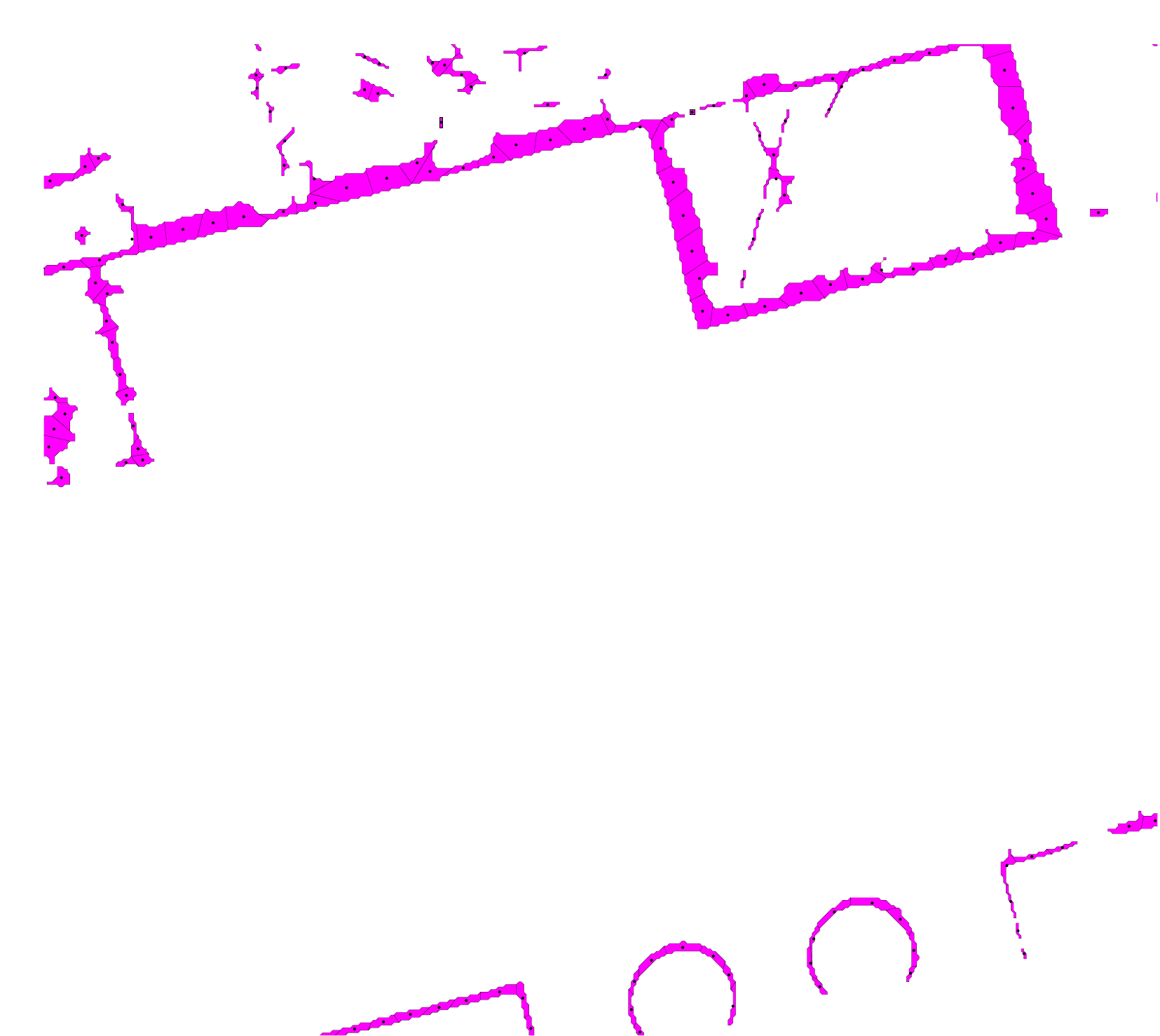}}
	\caption{Polygon maps for the Self dataset.}
	\label{Fig_20}
\end{figure}
Furthermore, to verify the effectiveness of the proposed approach in other complex operating environments, two self-recorded datasets (Self dataset-01 and Self dataset-02) are utilized for comparative evaluation of pose tracking. For Self dataset-01 and Self dataset-02, the ground truth is derived using LIO-SAM \cite{Shan2}, which integrates data from the LiDAR sensor, high-precision GPS, and an inertial measurement unit (IMU). This integration yields the ground truth for keyframes and records the relative pose of each frame with respect to its corresponding keyframe. Finally, linear interpolation is applied to calculate the pose of each frame, thereby obtaining the ground truth for the entire dataset.

\begin{figure}[!htb]
	\centering
	\includegraphics[width=8cm]{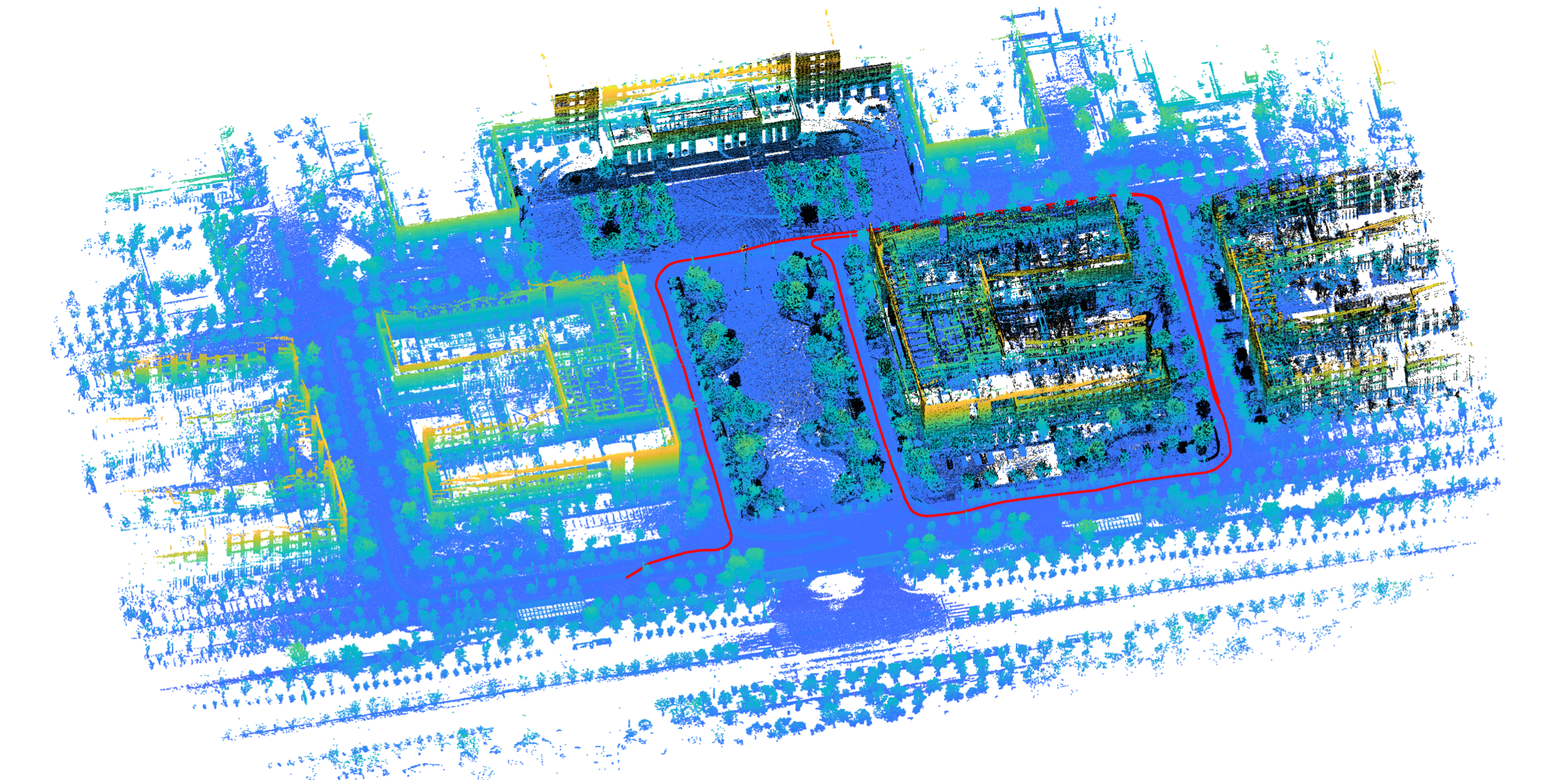}
	\caption{Diagram of point cloud changes in the testing environment for Self dataset-01. The colored point cloud represents the prior map obtained a year and a half ago, while the black point cloud represents the point cloud information of the current testing environment.}
	\label{Fig_21}
\end{figure}

For the Self dataset-01, the corresponding prior polygon map is shown in Fig. \ref{Fig_20}(a). There are three test sequences for evaluation of pose tracking, and the comparative experimental results are shown in Tab. \ref{table_8}. Note that the first two test sequences are collected at the same time as the dataset used for constructing the prior map, while the third test sequence (Self dataset-01-3) is obtained a year and a half later. As the detailed comparative experimental results in Tab. \ref{table_8}, compared to the other six approaches, the proposed ERPoT achieves the superior performance in terms of map size, translational error, and rotational error, while simultaneously maintaining high and stable algorithm running efficiency. In particular, the experimental results using Self dataset-01-3 demonstrate that the novel polygon map and pose tracking approach have long-term viability, and the long-term environmental changes in the testing environment are illustrated in Fig. \ref{Fig_21}. In contrast, {HDL\_localization} and {SM\_MCL} have encountered failures within this evaluation. Unlike {ALOAM\_MCL}, {FLOAM\_MCL}, and {KISS\_MCL}, these methods rely solely on the prior point cloud map obtained over a year and a half ago. The primary reason for these shortcomings is the significant environmental changes that hinder point cloud registration, thereby affecting the stability and accuracy of pose tracking.

\begin{table*}[!htb]\small
	\renewcommand\arraystretch{0.7}
	\centering
	\caption{Experimental results of different approaches using the Self dataset}
	\setlength{\tabcolsep}{1mm}{
		{\fontsize{6pt}{6pt}
		\begin{tabular}{|l|c|c|c|c|c|c|c|c|c|c|c|}
			\hline 
			\multirow{2}{*}{\textbf{Dataset No.}} &\textbf{Approach} &\textbf{Prior map} &\multicolumn{3}{c|}{\textbf{Translational error (cm)}} &\multicolumn{3}{c|}{\textbf{Rotational error (deg)}} &\multicolumn{3}{c|}{\textbf{Runtime (ms)}} \\
			\cline{4-12}
			&\textbf{name}   &\textbf{size (MB) $\downarrow$} &\textbf{Max} $\downarrow$  &\textbf{Mean} $\downarrow$  &\textbf{RMSE} $\downarrow$  &\textbf{Max} $\downarrow$ &\textbf{Mean} $\downarrow$ &\textbf{RMSE} $\downarrow$ &\textbf{Max} $\downarrow$ &\textbf{Mean} $\downarrow$ &\textbf{SD}$\downarrow$ \\
			\hline \hline
			\multirow{7}{*}{\textbf{Self dataset-01-1}} 
			& \textbf{ALOAM\_MCL}        &\multirow{4}{*}{${111.91}$}  &$53.49$  &$24.30$  &$26.08$   &$5.59$   &$0.91$   &$1.25$   &$508.59$   &$106.54$    &$106.63$    \\
			& \textbf{FLOAM\_MCL}        &   &$109.57$  &$33.12$   &$40.95$   &$5.16$   &$0.92$   &$1.23$   &$234.67$    &$117.58$   &$16.41$ \\
			& \textbf{KISS\_MCL}         &   &$87.40$  &$27.74$   &$32.71$   &$5.48$   &$0.90$   &$1.19$   &$271.03$    &$\mathbf{42.96}^{\star}$   &$12.83$  \\
			& \textbf{HDL\_localization} &  &$-$  &$-$  &$-$   &$-$   &$-$   &$-$   &$-$   &$-$    &$-$   \\
			& \textbf{SM\_MCL}           &$19257.15$   &$50.42$  &$20.96$   &$23.13$   &$\mathbf{1.08}^{\star}$   &$0.40$   &$0.42$   &$139.30$    &$64.38$   &$10.90$  \\
			& \textbf{Range\_MCL}        &${56.15}$   &${73.26}$  &${29.85}$   &${32.44}$   &${9.00}$   &${0.88}$   &${1.62}$   &$600.22$   &$537.70$   &$127.39$\\
			& \textbf{ERPoT(ours)}     &$\mathbf{0.98}^{\star}(8619/122066)$   &$\mathbf{22.26}^{\star}$  &$\mathbf{9.48}^{\star}$   &$\mathbf{10.24}^{\star}$   &$1.15$   &$\mathbf{0.11}^{\star}$   &$\mathbf{0.18}^{\star}$   &$\mathbf{81.16}^{\star}$    &$50.19$   &$\mathbf{5.99}^{\star}$  \\
			\hline
			\multirow{7}{*}{\textbf{Self dataset-01-2}} 
			& \textbf{ALOAM\_MCL}        &\multirow{4}{*}{${111.91}$}  &$27.28$  &$13.14$  &$14.03$   &$5.04$   &$0.60$   &$0.82$   &$291.12$   &$160.07$    &$79.91$  \\
			& \textbf{FLOAM\_MCL}        &   &$42.27$  &$14.44$   &$15.35$   &$2.66$   &$0.51$   &$0.65$   &$265.13$    &$100.98$   &$14.38$\\
			& \textbf{KISS\_MCL}         &   &$37.07$  &$22.11$   &$23.59$   &$2.91$   &$0.61$   &$0.78$   &$\mathbf{73.78}^{\star}$    &$\mathbf{37.00}^{\star}$   &$9.94$ \\
			& \textbf{HDL\_localization} &   &$\mathbf{21.03}^{\star}$  &$14.05$   &$14.33$   &$1.94$   &$1.39$   &$1.41$   &$768.93$   &$55.18$   &$25.72$ \\
			& \textbf{SM\_MCL}           &$19257.15$   &$39.41$  &$20.26$   &$21.29$   &$1.11$   &$0.62$   &$0.64$   &$92.78$    &$54.32$   &$10.26$ \\
			& \textbf{Range\_MCL}        &${56.15}$   &${73.23}$  &${29.96}$   &${30.87}$   &${2.73}$   &${0.70}$   &${0.89}$   &$610.77$   &$468.34$   &$63.05$\\
			& \textbf{ERPoT(ours)}     &$\mathbf{0.98}^{\star}(8619/122066)$   &$28.83$  &$\mathbf{6.97}^{\star}$   &$\mathbf{7.99}^{\star}$   &$\mathbf{0.65}^{\star}$   &$\mathbf{0.17}^{\star}$   &$\mathbf{0.23}^{\star}$   &$77.04$    &$51.93$   &$\mathbf{5.03}^{\star}$   \\
			\hline
			\multirow{7}{*}{\textbf{Self dataset-01-3}} 
			& \textbf{ALOAM\_MCL}        &\multirow{4}{*}{${111.91}$}  &$\mathbf{81.99}^{\star}$  &$22.77$  &$26.03$   &$3.72$   &$0.65$   &$0.84$   &$1030.19$   &$199.14$    &$200.79$  \\
			& \textbf{FLOAM\_MCL}        &  &$106.90$  &$26.90$   &$33.07$   &$11.46$   &$0.77$   &$1.13$   &$308.44$    &$147.07$   &$38.18$ \\
			& \textbf{KISS\_MCL}         &  &$-$  &$-$  &$-$   &$-$   &$-$   &$-$   &$-$   &$-$    &$-$   \\
			& \textbf{HDL\_localization} &  &$-$  &$-$  &$-$   &$-$   &$-$   &$-$   &$-$   &$-$    &$-$   \\
			& \textbf{SM\_MCL}           &$19257.15$  &$-$  &$-$  &$-$   &$-$   &$-$   &$-$   &$-$   &$-$    &$-$   \\
			& \textbf{Range\_MCL}        &${56.15}$   &${301.81}$  &${37.34}$   &${52.83}$   &${26.76}$   &${1.56}$   &${3.39}$   &$724.46$   &$421.64$   &$210.58$\\
			& \textbf{ERPoT(ours)}     &$\mathbf{0.98}^{\star}(8619/122066)$   &$139.96$  &$\mathbf{8.92}^{\star}$   &$\mathbf{10.86}^{\star}$   &$\mathbf{2.59}^{\star}$   &$\mathbf{0.14}^{\star}$   &$\mathbf{0.23}^{\star}$   &$\mathbf{108.46}^{\star}$    &$\mathbf{62.23}^{\star}$   &$\mathbf{7.53}^{\star}$   \\
			\hline \hline
			\multirow{7}{*}{\textbf{Self dataset-02-1}} 
			& \textbf{ALOAM\_MCL}        &\multirow{4}{*}{${32.92}$}  &$102.32$  &$23.09$  &$30.57$   &$3.13$    &$0.64$   &$0.83$    &$367.93$   &$139.53$    &$95.97$    \\
			& \textbf{FLOAM\_MCL}        &   &$53.06$  &$16.47$   &$20.15$   &$2.61$   &$0.57$   &$0.73$   &$139.14$    &$82.32$   &$16.74$ \\
			& \textbf{KISS\_MCL}         &   &$92.98$  &$20.29$   &$26.28$   &$3.30$   &$0.63$   &$0.83$   &$\mathbf{53.71}^{\star}$    &$\mathbf{31.62}^{\star}$   &$\mathbf{6.55}^{\star}$  \\
			& \textbf{HDL\_localization} &   &$50.67$  &$\mathbf{4.81}^{\star}$  &$\mathbf{6.18}^{\star}$   &$4.12$   &$1.38$   &$1.54$   &$205.48$   &$42.77$    &$18.76$   \\
			& \textbf{SM\_MCL}           &$5589.06$   &$30.79$  &$11.89$   &$13.17$   &$\mathbf{1.26}^{\star}$   &$0.35$   &$0.42$   &$180.10$    &$45.12$   &$11.24$  \\
			& \textbf{Range\_MCL}        &${28.83}$   &${350.34}$  &${131.15}$   &${154.27}$   &${10.39}$   &${1.56}$   &${2.41}$   &$803.37$   &$457.29$   &$209.51$\\
			& \textbf{ERPoT(ours)}     &$\mathbf{0.65}^{\star}(5855/78864)$   &$\mathbf{29.62}^{\star}$  &$6.33$   &$7.55$   &$1.40$   &$\mathbf{0.23}^{\star}$   &$\mathbf{0.33}^{\star}$   &$61.26$    &$39.07$   &$7.92$  \\
			\hline
	\end{tabular}}}
	\label{table_8}
\end{table*}

On the other hand, taking into account the diversity of real-world operating environments, the Self dataset-02 is obtained by selecting an open square that includes uphill and downhill sections. This choice ensures a comprehensive evaluation of the performance of ERPoT across diverse terrains, and simulates the challenges encountered in actual deployments. The obtained prior polygon map is shown in Fig. \ref{Fig_20}(b).

More importantly, the comparative experimental results using Self dataset-02 are shown in Tab. \ref{table_8}. In particular, the prior map is remarkably compact, with a size of merely $0.65$ MB, and is significantly more concise than the point cloud map and distance field representation. In terms of translational and rotational errors, the proposed ERPoT has achieved optimal or near-optimal performance, while maintaining high and stable operational efficiency. Experimental results demonstrate the effectiveness of ERPoT across diverse terrains.

\subsection{Ablation study}

\begin{table}[!htb]\small
	\renewcommand\arraystretch{0.68}
	\centering
	\caption{Experimental results of ablation study for ERPoT without(w/o) point-vertex constraint or point-edge constraint}
	\setlength{\tabcolsep}{1mm}{
		{\fontsize{5.6pt}{5.6pt}
					\begin{tabular}{cccccccccc}
						\hline
						\multirow{2}{*}{\textbf{Dataset}} &\multirow{2}{*}{\textbf{No.}} &\multicolumn{4}{c}{\textbf{w/o Point-vertex constrait}} &\multicolumn{4}{c}{\textbf{w/o Point-edge constrait}}\\
						\cline{3-10}
						& &\textbf{Succ.} &\multicolumn{3}{c}{\textbf{Impact on pose tracking}} &\textbf{Succ.}  &\multicolumn{3}{c}{\textbf{Impact on pose tracking}} \\
						\hline
						\multirow{18}{*}{\textbf{KITTI}}
						& \multirow{2}{*}{\textbf{00-1}} &\multirow{2}{*}{\ding{52}} &\cellcolor{gray!20}${63.22}$ &$11.52$ &$14.79$ &\multirow{2}{*}{\ding{52}} &\cellcolor{gray!20}${48.88}$ &\cellcolor{gray!20}${16.10}$ &\cellcolor{gray!20}${19.06}$ \\
						& & &\cellcolor{gray!20}${0.84}$ &$0.37$ &$0.42$ & &\cellcolor{gray!20}${0.87}$ &\cellcolor{gray!20}${0.40 }$ &\cellcolor{gray!20}${0.45}$\\
						\hhline{|~~|--------|}
						& \multirow{2}{*}{\textbf{00-2}} &\multirow{2}{*}{\ding{52}} &$20.87$ &$7.68$ &$8.77$ &\multirow{2}{*}{\ding{52}}  &\cellcolor{gray!20}${33.17}$ &\cellcolor{gray!20}${10.78}$ &\cellcolor{gray!20}${12.21}$\\
						& & &\cellcolor{gray!20}$0.83$ &$0.21$ &$0.26$ & &\cellcolor{gray!20}${1.28}$ &\cellcolor{gray!20}${0.24}$ &\cellcolor{gray!20}${0.30}$\\
						\hhline{|~~|--------|}
						& \multirow{2}{*}{\textbf{00-3}} &\multirow{2}{*}{\ding{52}} &\cellcolor{gray!20}${43.38}$ &\cellcolor{gray!20}${13.89}$ &\cellcolor{gray!20}${18.25}$ &\multirow{2}{*}{\ding{52}} &\cellcolor{gray!20}${38.19}$ &\cellcolor{gray!20}${15.95}$ &\cellcolor{gray!20}${19.37}$\\                     
						& & &\cellcolor{gray!20}${0.58}$ &\cellcolor{gray!20}${0.16}$ &\cellcolor{gray!20}${18.25}$ & &\cellcolor{gray!20}${0.55}$ &\cellcolor{gray!20}${0.16}$ &\cellcolor{gray!20}${0.20}$\\                
						\hhline{|~~|--------|}
						& \multirow{2}{*}{\textbf{02-1}} &\multirow{2}{*}{\ding{52}} &\cellcolor{gray!20}${35.54}$ &\cellcolor{gray!20}${12.30}$ &\cellcolor{gray!20}${14.11}$ &\multirow{2}{*}{\ding{52}} &${30.36}$ &\cellcolor{gray!20}${12.03}$ &\cellcolor{gray!20}${13.75}$ \\ 
						& & &${0.50}$ &\cellcolor{gray!20}${0.25}$ &\cellcolor{gray!20}${0.28}$ & &\cellcolor{gray!20}${0.70 }$ &${0.23}$ &${0.26 }$\\            
						\hhline{|~~|--------|}
						& \multirow{2}{*}{\textbf{02-2}} &\multirow{2}{*}{\ding{52}} &\cellcolor{gray!20}${86.79}$ &\cellcolor{gray!20}${19.31}$ &\cellcolor{gray!20}${24.22}$ &\multirow{2}{*}{\ding{52}} &${77.25 }$ &\cellcolor{gray!20}${19.15}$ &\cellcolor{gray!20}${22.54}$\\
						& & &\cellcolor{gray!20}${0.31}$ &\cellcolor{gray!20}${0.39}$ &\cellcolor{gray!20}${24.22}$ & &\cellcolor{gray!20}${0.86}$ &${0.29}$ &${0.35}$\\                 
						\hhline{|~~|--------|}
						& \multirow{2}{*}{\textbf{05-1}} &\multirow{2}{*}{\ding{52}} &\cellcolor{gray!20}${29.83}$ &${7.30}$ &${8.67}$ &\multirow{2}{*}{\ding{52}} &\cellcolor{gray!20}${28.38}$ &\cellcolor{gray!20}${11.07}$ &\cellcolor{gray!20}${12.20}$ \\
						& & &${1.46}$ &${8.67}$ &${0.31}$ & &${1.58}$ &\cellcolor{gray!20}${0.26}$ &\cellcolor{gray!20}${0.36 }$\\
						\hhline{|~~|--------|}
						& \multirow{2}{*}{\textbf{05-2}} &\multirow{2}{*}{\ding{52}} &\cellcolor{gray!20}${42.80}$ &${8.54}$ &${10.11}$ &\multirow{2}{*}{\ding{52}} &\cellcolor{gray!20}${36.80}$ &\cellcolor{gray!20}${14.33}$ &\cellcolor{gray!20}${15.76}$ \\
						& & &\cellcolor{gray!20}${0.76}$ &${0.25}$ &${0.29}$ & &\cellcolor{gray!20}${0.77}$ &\cellcolor{gray!20}${0.28}$ &\cellcolor{gray!20}${0.32}$\\ 
						\hhline{|~~|--------|}
						& \multirow{2}{*}{\textbf{06-1}} &\multirow{2}{*}{\ding{52}} &\cellcolor{gray!20}${44.95}$ &${7.20}$ &\cellcolor{gray!20}${8.39}$ &\multirow{2}{*}{\ding{52}} &${26.31}$ &\cellcolor{gray!20}${9.06}$  &\cellcolor{gray!20}${9.97}$ \\
						& & &\cellcolor{gray!20}${0.57}$ &\cellcolor{gray!20}${0.24}$ &\cellcolor{gray!20}${0.26}$ & &\cellcolor{gray!20}${0.52}$ &${0.23}$ &\cellcolor{gray!20}${0.26}$\\  
						\hhline{|~~|--------|}
						& \multirow{2}{*}{\textbf{08-1}} &\multirow{2}{*}{\ding{56}} & & & &\multirow{2}{*}{\ding{52}} &${72.36}$ &\cellcolor{gray!20}${17.88}$  &\cellcolor{gray!20}${23.57}$\\
						& & & & & &  &\cellcolor{gray!20}${1.29}$  &${0.33}$ &${0.40}$ \\                     
						\hline
						\multirow{4}{*}{\textbf{Newer College}}
						& \multirow{2}{*}{\textbf{00-1}} &\multirow{2}{*}{\ding{56}} & & & &\multirow{2}{*}{\ding{52}} &\cellcolor{gray!20}${146.41}$ &\cellcolor{gray!20}${19.56}$ &\cellcolor{gray!20}${22.70}$ \\
						& & & & & & &$18.20$ &$0.87$ &$1.19$ \\     
						\hhline{|~~|--------|}
						& \multirow{2}{*}{\textbf{00-2}} &\multirow{2}{*}{\ding{56}} & & & &\multirow{2}{*}{\ding{56}} & & & \\
						& & & & & & & & &\\
						\hline
						\multirow{10}{*}{\textbf{FusionPortable}}
						& \multirow{2}{*}{\textbf{campus}} &\multirow{2}{*}{\ding{56}} & & & &\multirow{2}{*}{\ding{52}} &\cellcolor{gray!20}${46.23}$ &\cellcolor{gray!20}${46.23}$ &\cellcolor{gray!20}${20.47}$ \\
						& & & & & & &$6.58$ &$1.01$ &$1.39$\\
						\hhline{|~~|--------|}
						& \multirow{2}{*}{\textbf{building-1}} &\multirow{2}{*}{\ding{56}} & & & &\multirow{2}{*}{\ding{52}} &\cellcolor{gray!20}${51.66}$ &\cellcolor{gray!20}${14.41}$ &\cellcolor{gray!20}${16.90}$ \\
						& & & & & & &\cellcolor{gray!20}${5.59}$ &${0.46}$ &\cellcolor{gray!20}${0.64}$\\
						\hhline{|~~|--------|}
						& \multirow{2}{*}{\textbf{building-2}} &\multirow{2}{*}{\ding{56}} & & & &\multirow{2}{*}{\ding{52}} &\cellcolor{gray!20}${46.27}$ &\cellcolor{gray!20}${13.69}$ &\cellcolor{gray!20}${14.93}$ \\
						& & & & & & &\cellcolor{gray!20}${3.15}$ &\cellcolor{gray!20}${0.63}$ &\cellcolor{gray!20}${0.81}$ \\            
						\hhline{|~~|--------|}
						& \multirow{2}{*}{\textbf{garden}} &\multirow{2}{*}{\ding{56}} & & & &\multirow{2}{*}{\ding{52}} &\cellcolor{gray!20}${266.39}$ &\cellcolor{gray!20}${17.03}$ &\cellcolor{gray!20}${27.72}$ \\
						& & & & & & &\cellcolor{gray!20}${21.90}$ &\cellcolor{gray!20}${0.74}$ &\cellcolor{gray!20}${1.48}$ \\
						\hhline{|~~|--------|}
						& \multirow{2}{*}{\textbf{canteen}} &\multirow{2}{*}{\ding{56}} & & & &\multirow{2}{*}{\ding{52}} &\cellcolor{gray!20}${47.00}$ &\cellcolor{gray!20}${11.32}$ &\cellcolor{gray!20}${12.53}$ \\
						& & & & & & &\cellcolor{gray!20}${5.26}$ &\cellcolor{gray!20}${0.61}$ &\cellcolor{gray!20}${0.85}$ \\
						\hline
						\multirow{8}{*}{\textbf{Self dataset}}
						& \multirow{2}{*}{\textbf{01-1}} &\multirow{2}{*}{\ding{52}} &$20.29$ &$7.36$ &$7.86$ &\multirow{2}{*}{\ding{52}} &\cellcolor{gray!20}${24.70}$ &\cellcolor{gray!20}${10.95}$ &\cellcolor{gray!20}${11.89}$ \\
						& & &\cellcolor{gray!20}${1.51}$ &$0.10$ &$0.18$ & &$1.13$ &\cellcolor{gray!20}${0.12}$ &\cellcolor{gray!20}${0.19}$\\
						\hhline{|~~|--------|}
						& \multirow{2}{*}{\textbf{01-2}} &\multirow{2}{*}{\ding{52}} &\cellcolor{gray!20}${159.53}$ &${6.17}$ &\cellcolor{gray!20}${14.24}$ &\multirow{2}{*}{\ding{52}} &\cellcolor{gray!20}${71.73}$ &\cellcolor{gray!20}${9.42}$ &\cellcolor{gray!20}${11.08}$ \\
						& & &\cellcolor{gray!20}${0.85}$ &${0.17}$ &${0.22}$ & &\cellcolor{gray!20}${0.68}$ &\cellcolor{gray!20}${0.18}$ &\cellcolor{gray!20}${0.24}$\\
						\hhline{|~~|--------|}
						& \multirow{2}{*}{\textbf{01-3}} &\multirow{2}{*}{\ding{56}} & & & &\multirow{2}{*}{\ding{52}} &${106.83}$ &\cellcolor{gray!20}${10.16}$ &\cellcolor{gray!20}${12.20}$ \\
						& & & & & & &${2.55}$ &\cellcolor{gray!20}${0.15}$ &\cellcolor{gray!20}${0.24}$\\                     
						\hhline{|~~|--------|}
						& \multirow{2}{*}{\textbf{02-1}} &\multirow{2}{*}{\ding{56}} & & & &\multirow{2}{*}{\ding{52}} &\cellcolor{gray!20}${31.39}$ &$7.44$ &\cellcolor{gray!20}${8.59}$ \\
						& & & & & & &$1.13$ &$0.23$ &$0.33$\\
						\hline
	\end{tabular}}}
	\label{table_9}
\end{table}

To systematically evaluate the contributions of various components in our proposed ERPoT, an ablation study is conducted. This study aims to provide insights into how each component affects the overall performance of the system, particularly focusing on the point-vertex and point-edge constraints. Importantly, we evaluate the performance of ERPoT both with and without these constraints across all datasets.

{The results of the ablation study are presented in Tab. \ref{table_9}. The top three columns represent translational errors, while the bottom three columns show rotational errors. Results that are worse than those of the complete ERPoT are highlighted with gray shading. It is clear that the point-edge constraint plays a significant role in successful pose tracking, and incorporating the point-edge constraint can further enhance the accuracy.

\section{Discussion}

\subsection{Limitations and future work}
While ERPoT achieves excellent performance in map size, translational error, rotational error, and runtime, it has limitations. Specifically, ERPoT relies on ground segmentation to transform dense 3D LiDAR point clouds into sparse 2D scans. This dependency can cause issues in environments where reliable ground information cannot be extracted, such as in scenarios with uneven terrain or cluttered scenes with low obstacles. In these cases, the ground segmentation process may fail, leading to degraded performance.

Moreover, the dependency of ERPoT on ground segmentation makes it less versatile compared to other pose-tracking methods that do not require such preprocessing steps. To address these limitations, future work could explore alternative preprocessing techniques that are more robust to environmental variations. On the other hand, it is worth noting that there are alternative approaches (\emph{e.g.}, ROS packages like pointcloud\_to\_laserscan) that can extract a 2D scan from a 3D LiDAR without relying on ground detection. Additionally, the lightweight and compact polygon map is ideal for multi-robot collaborative tasks, enabling efficient map sharing, faster communication, and enhanced system performance, while its concise representation supports rapid scene recognition, crucial for real-time applications like search and rescue or dynamic industrial settings.

\subsection{Probabilistic model extension}
In the field of robot pose tracking, Monte Carlo localization-based approaches play the crucial role, which makes use of a probabilistic estimation technique to properly handle uncertainties about initial, motion process, and sensor observation. In this section, we explore the integration of our proposed approach as an measurement model within the Monte Carlo Localization (MCL) framework. Our proposed ERPoT is a approach which relies on a lightweight and compact polygon map for pose tracking. Specifically, it converts dense 3D LiDAR point clouds into sparse 2D scans and employs a novel cost function for pose estimation through point-polygon matching. This approach ensures efficient and accurate pose tracking, making it suitable for integration with MCL.

The complete probabilistic localization process contains initialization, motion model, measurement model, resampling and pose estimation. In this section, we focus on the measurement model, which makes use of the kernel contributions on the proposed novel polygon map representation and the data association between the point and the polygon. For time stamp $t$, given the generated sparse 2D scan observation $\mathbf{C}_t$ and the lightweight polygon map ${\cal M}_p$, the measurement model is represented as follows
\begin{align}
p(\mathbf{C}_t^{hit}|\tilde{\mathbf{s}}_t, {\cal M}_p) = \prod\nolimits_{i = 1}^K {p\left( {\mathbf{p}_i^{2D}} | \tilde{\mathbf{s}}_t, \mathbf{P}_{i \to j}\right)},
\end{align}where the $j$-th corresponding polygon $\mathbf{P}_j \in {\cal M}_p $, and $\mathbf{C}_t^{hit}$ is subset of $\mathbf{C}_t$ with $K$ scan points that meet the condition
\begin{align}
d^*(\mathbf{p}_i^{2D}, {\cal M}_p) = {\min _{j = 1,2,...,m}}d\left( {\mathbf{p}_i^{2D}, \mathbf{P}_j} \right) < \delta_d,
\end{align}and $\delta_d$ is the data association threshold. Accordingly,
\begin{align}
p\left( {\mathbf{p}_i^{2D}} | \tilde{\mathbf{s}}_t, \mathbf{P}_{i \to j}\right) = \mathcal{N}\left(d^*;0,\sigma_{hit}^2\right),
\end{align}where $\sigma_{hit}$ is the standard deviation.

\section{Conclusion}

In this study, an efficient and reliable pose tracking framework for mobile robots called ERPoT has been proposed, which makes use of a novel prior map constructed from multiple compact polygons. Notably, this system operates solely on LiDAR data, acquiring the sparse 2D scan points through ground removal and obstacle selection. Following this, an offline map construction procedure is executed, yielding an environment map that is both lightweight and compact in its polygon representation. Ultimately, by integrating real-time 2D scan points with the prior polygon map, our approach achieves reliable long-term pose tracking based on the newly proposed point-polygon matching. Furthermore, the quantitative evaluation shows the superior performance on prior map size, pose estimation error, and runtime for pose tracking compared with the other six approaches.

\end{document}